# BHASA:
## A Holistic Southeast Asian Linguistic and Cultural Evaluation Suite for Large Language Models

**Wei Qi Leong**[*]
AI Singapore
National University of Singapore
weiqi@aisingapore.org

**Jian Gang Ngui**[*]
AI Singapore
National University of Singapore
Department of Linguistics,
University of Arizona
jiangangngui@aisingapore.org

**Yosephine Susanto**[†]
AI Singapore
National University of Singapore
yosephine@aisingapore.org

**Hamsawardhini Rengarajan**[†]
AI Singapore
National University of Singapore
hamsa@aisingapore.org

**Kengatharaiyer Sarveswaran**
Department of Computer Science,
University of Jaffna
sarves@univ.jfn.ac.lk

**William Chandra Tjhi**
AI Singapore
National University of Singapore
wtjhi@aisingapore.org

## Abstract

The rapid development of Large Language Models (LLMs) and the emergence of novel abilities with scale have necessitated the construction of holistic, diverse and challenging benchmarks such as HELM and BIG-bench. However, at the moment, most of these benchmarks focus only on performance in English and evaluations that include Southeast Asian (SEA) languages are few in number. We therefore propose **BHASA**, a holistic linguistic and cultural evaluation suite for LLMs in SEA languages. It comprises three components: (1) a NLP benchmark covering eight tasks across Natural Language Understanding (NLU), Generation (NLG) and Reasoning (NLR) tasks, (2) **LINDSEA**, a linguistic diagnostic toolkit that spans the gamut of linguistic phenomena including syntax, semantics and pragmatics, and (3) a cultural diagnostics dataset that probes for both cultural representation and sensitivity. For this preliminary effort, we implement the NLP benchmark only for Indonesian, Vietnamese, Thai and Tamil, and we only include Indonesian and Tamil for LINDSEA and the cultural diagnostics dataset. As GPT-4 is purportedly one of the best-performing multilingual LLMs at the moment, we use it as a yardstick to gauge the capabilities of LLMs in the context of SEA languages. Our initial experiments on GPT-4 with BHASA find it lacking in various aspects of linguistic capabilities, cultural representation and sensitivity in the targeted SEA languages. BHASA is a work in progress and will continue to be improved and expanded in the future. The repository for this paper can be found at: https://github.com/aisingapore/BHASA.

---

[*]Co-first authors; corresponding authors
[†]Major contributors; full contributions in Appendix A.

# Contents

# 1 Introduction

Large Language Models (LLMs) have recently been catapulted into the spotlight for their immense potential, demonstrating strong zero-shot or few-shot capabilities across a wide spectrum of Natural Language Processing (NLP) tasks [1, 2, 3, 4, 5]. The rapid development of these LLMs and the emergence of novel abilities with scale [6] have necessitated the construction of more difficult and diverse benchmarks such as BIG-bench [7] as well as holistic benchmarks such as HELM [8]. However, most of these benchmarks focus only on performance in English at the moment. Existing works benchmarking LLMs on non-English languages are few in number [9, 10, 11, 12, 13] and are not as holistic and diverse as HELM or BIG-bench.

As such, we present **BHASA**, a holistic linguistic and cultural evaluation suite for LLMs in SEA languages that we are continually improving and expanding. BHASA comprises three components: (1) a NLP benchmark covering eight tasks across Natural Language Understanding (NLU), Generation (NLG) and Reasoning (NLR) tasks, (2) **LINDSEA**, a linguistic diagnostic toolkit that spans the gamut of linguistic phenomena including syntax, semantics and pragmatics, and (3) a cultural diagnostics dataset that probes for both cultural representation and sensitivity.

The NLP benchmark covers Named Entity Recognition (NER), Question Answering, Sentiment Analysis, Toxicity Detection, Machine Translation, Abstractive Summarization, Natural Language Inference (NLI) and Causal Reasoning. In this first version of the benchmark, we include only Indonesian, Vietnamese, Thai and Tamil as target languages in the NLP benchmark due to resource constraints.

Furthermore, given that LLMs are now touted as being able to interact with users purely through natural language, the systematic evaluation of natural language proficiency becomes even more crucial. We therefore include LINDSEA as the second component in BHASA. LINDSEA is a handcrafted linguistic diagnostic benchmark designed by linguists to holistically and quantitatively evaluate a model's understanding of various linguistic phenomena specific to SEA languages. LINDSEA is designed to have a broad and deep coverage of linguistic features including syntax, semantics and pragmatics to allow for fine-grained analyses and precise pinpointing of a model's linguistic weaknesses. In this first version of the benchmark, we release both an Indonesian dataset as well as an early version of the Tamil dataset that will be expanded in a future update of this paper. Additionally, given the importance of cultural representation and sensitivity in the use of LLMs, we also conduct preliminary experiments on the matter with a set of Indonesian and Tamil culture-specific questions as the third component in BHASA.

At the time of this writing, GPT-4 is arguably among the best-performing LLMs available [14] that is touted as explicitly supporting multilingual capabilities, which importantly includes some SEA languages [15]. As such, it could be a good candidate for getting a sense of how well LLMs can currently perform in the context of SEA languages. We therefore evaluate GPT-4 on BHASA to investigate the current state-of-the-art performance on SEA languages, including also results on its immediate predecessor GPT-3.5-Turbo as a reference and to enable insights into the growth trajectory of this family of models. We reiterate that BHASA is designed to be model-agnostic, even if only GPT-3.5-Turbo and GPT-4 are evaluated in this paper.

Our initial findings, based on GPT-4's evaluation for the time being as a proxy representation of the current state-of-the-art multilingual performance in LLMs, show that:

1. In general, there have been significant improvements in zero-shot SEA language capabilities from the time of GPT-3.5-Turbo to that of GPT-4, with GPT-4 being able to better follow instructions in native prompts, sometimes even benefiting from them.

2. Among our target SEA languages, LLM performance decreases in the order of Indonesian, Vietnamese, Thai and Tamil.

3. Performance in Question Answering, Sentiment Analysis, Abstractive Summarization and Causal Reasoning in our target languages is quite good, but there is still much room for improvement in Named Entity Recognition (NER), Toxicity Detection, Machine Translation and Natural Language Inference (NLI).

4. Idiosyncratic linguistic features of Indonesian and Tamil are not well-represented in LLMs.

5. There is a conspicuous lack of cultural representation and sensitivity with regard to Indonesian and Tamil culture in LLMs.

## 2  Related work

### 2.1  Holistic NLP benchmarks

With the advent of transformer-based pre-trained models [16] starting with BERT [17], GLUE [18] and SuperGLUE [19] have become mainstays in the multi-task evaluation of NLP models on Natural Language Understanding (NLU) tasks. GLUE also contains a diagnostic dataset covering syntax, semantics, reasoning skills and knowledge that allows for a targeted diagnosis of a model's strengths and weaknesses. However, these benchmarks were quickly saturated, with models outperforming humans, often by large margins [20]. HELM [8] was an effort to holistically evaluate LLMs given the extent of the gamut of tasks that they are capable of handling. It introduced the idea that accuracy is no longer sufficient in the current age, and that other metrics such as robustness, fairness, bias, toxicity, calibration and efficiency are equally important to be measured. It also included targeted evaluations of language, reasoning, knowledge, memorization, disinformation, bias and toxicity. While it was a very ambitious study that evaluated many models on many scenarios and metrics, it predated GPT-3.5-Turbo and GPT-4 and only focused on English given the massive scope that was covered. BIG-Bench [7] was another effort in response to the emergent abilities of LLMs [6] to crowdsource innovative tasks to explore yet undiscovered capabilities of these models. It contains a few tasks for evaluating certain aspects of natural language use, but they are much smaller in scale and diversity compared to BLiMP [21] which was included in HELM. OpenAI also open-sourced their own crowdsourced benchmark, OpenAI Evals,[1] for users to report shortcomings in their models.[2] They cover a wide range of tasks across many languages, but only have two tasks related to SEA languages at the point of writing this paper – Indonesian written number to Arabic numeral conversion and Vietnamese examination multiple-choice questions. We mainly draw inspiration from HELM in aspiring to be holistic but we also go beyond HELM and BIG-Bench or OpenAI Evals in designing a more extensive linguistic diagnostic toolkit for SEA languages as well as cultural diagnostics.

### 2.2  Individual evaluations

Reasoning is extremely important in decision-making, making it a key skill to have for LLMs especially with the dawn of the new paradigm of using LLMs as task-planning agents [22]. [23] conducted a comprehensive study comparing ChatGPT-3.5 and ChatGPT-4's zero-shot reasoning capabilities, although they do not include datasets that explicitly target pragmatic reasoning. LogiEval [24] tested ChatGPT, ChatGPT-4, and GPT-4 on English and Chinese reasoning tasks, but did not target any SEA language and also did not target pragmatic reasoning. Studies that did include pragmatic reasoning did not cover SEA languages and were not tested on GPT-3.5-Turbo or GPT-4. For example, multiple works have introduced datasets to test for scalar implicatures [25, 26, 27, 28], while others have introduced datasets for presuppositions, such as IMPPRES [25] and NOPE [29]. We draw inspiration from their datasets in crafting our own dataset for pragmatic reasoning in SEA languages.

Natural Language Generation (NLG) evaluations on LLMs often involve machine translation and summarization. [30] comprehensively evaluated the machine translation capabilities of GPT models (ChatGPT, text-davinci-003, and text-davinci-002) on 18 translation directions with both high and low-resource languages, using overlap metrics (BLEU[3] and ChrF [31]), neural network-based metrics, both with references (COMET-22 [32]) and without references (COMETkiwi [33]), as well as human evaluation. However, they did not include SEA languages and the study was published before the release of GPT-4. In addition, [34] point out that translating texts containing idioms is challenging for Transformer-based systems. They proposed IdiomKB, a multilingual knowledge base for English, Mandarin and Japanese idiomatic expressions and evaluated ChatGPT-3.5 on translating these idioms using BLEU, COMETkiwi and GPT-4. However, they do not include SEA languages. As for summarization, [35] evaluated InstructGPT [36] and other LLMs on English abstractive news summarization using human evaluation. Annotators were asked to rate summaries on their faithfulness, coherence and relevance, in accordance with the guidelines in [37]. However, the study again does not cover SEA languages and did not evaluate GPT-3.5-Turbo or GPT-4.

Given that linguistic theory can inform us on models' linguistic representations [38], many works introducing linguistic diagnostics for language models have been published, but they mainly focus on English. Minimal pairs have been used in various studies to diagnose model weaknesses in syntax. For example, the SyntaxGym [39, 40] platform was designed to help centralize the process of targeted syntactic evaluation using minimal pairs. BLiMP [21] and its descendants, CLiMP [41] and JBLiMP [42] also contain minimal pairs covering various syntactic phenomena. However, they only cover English, Chinese and Japanese respectively and do not cover SEA languages. Furthermore, the grammatically acceptable sentences in the minimal pairs can sometimes sound unnatural and odd, especially for BLiMP and CLiMP which were semi-automatically generated with templates and a lexicon.

---

[1] https://github.com/openai/evals/
[2] https://openai.com/research/gpt-4
[3] https://github.com/mjpost/sacrebleu

Finally, the issue of cultural representation and sensitivity has become increasingly important for the field of NLP with the introduction of LLMs. [43] discuss the challenges in preventing cultural hegemony in NLP systems and propose that the NLP community should aim for cross-cultural representation and preservation of cultural values. In the same line of thinking, [44] surveyed the risks of disproportionate cultural representation and sensitivity, raising issues such as the imposition of norms and values from a single originating culture and amplification of marginalization. They also point out the difficulties surrounding evaluation of such issues owing to the fact that there is no singular standard for cultural values and sensitivity. Nevertheless, attempts have been made at evaluating models for cultural biases. For example, Anthropic released GlobalOpinionQA [45], a large dataset containing 2,556 multiple-choice questions and responses on global issues across different countries. Their study found that their LLM responses tended to be more similar to the opinions of certain populations from the USA and some European or South American countries. Overall, to the best of our knowledge, there are no studies specifically targeting SEA languages for cultural representation. Given that GPT-4 is arguably one of the best-performing multilingual models at the time of writing, we conduct a preliminary study to evaluate the adequacy of GPT-4's Indonesian and Tamil cultural representation and sensitivity, which gives us a tentative yardstick to see how such representation and sensitivity can be further improved. In addition, we intend to design a model-agnostic systematic study of cultural sensitivity in the future.

## 2.3 Evaluation of LLMs in SEA Languages

Although much work has sprung up in recent times evaluating the performance of LLMs in English, only a handful of studies include SEA languages in their benchmarks. One of the first studies on ChatGPT that did so evaluated its zero-shot performance on Indonesian sentiment analysis and machine translation between English and Indonesian, Javanese and Sundanese [11]. The study used human evaluation for translation but was unclear on the protocol, using only a vague binary choice of correct or incorrect for each sentence. Furthermore, only small samples of between 30 to 200 sentences or documents were used for each task, which may be insufficient to draw statistically significant conclusions. Following that study, OpenAI published the GPT-4 Technical Report [15] which reported strong multilingual 3-shot performance of GPT-4 on a machine-translated version of MMLU [46], including Indonesian and Thai. However, essential details of the evaluation were not revealed and it remained unclear whether GPT-4 would be able to perform equally well on non-translated test data. [12] evaluated ChatGPT with both English and machine translated prompts in a zero-shot setting on multiple tasks in multiple languages, including part-of-speech (POS) tagging, NLI, question answering, commonsense reasoning, and abstractive summarization in Vietnamese, Thai and Tamil.

MEGA [9] followed soon after, aspiring to perform comprehensive multilingual benchmarking in the style of HELM. They covered a wide range of tasks, languages and LLMs in their work, even including various prompting strategies. Their scope of tasks is the most similar to the NLP benchmark component of BHASA, covering NER, question answering, NLI, causal reasoning and abstractive summarization in our target languages. However, they do not cover machine translation at all, and the toxicity classification and sentiment analysis tasks do not cover the languages that we focus on. Furthermore, although they do evaluate GPT-3.5-Turbo and GPT-4, they only evaluated GPT-4 on NLI and question answering. A key finding from the study was that machine-translated prompts often contain translation errors which influence task performance negatively, and therefore it was recommended that machine-translated prompts be verified by native speakers. We took this into consideration and asked native speakers to manually translate all our prompts. BUFFET [10] was also established to evaluate language models across a broad range of tasks and languages, but was more targeted at evaluating few-shot cross-lingual transfer. Unlike MEGA and BUFFET, BHASA focuses on zero-shot instead of few-shot evaluations.

Finally, M3Exam [13] was released as a multilingual, multimodal and multilevel benchmark for LLMs. It comprises 12,317 multiple-choice examination paper questions across 9 languages, including Vietnamese, Thai and Javanese and was tested on GPT-3.5-Turbo, GPT-4, BLOOM and Vicuna. However, it is more similar to MMLU and does not specifically target any aspect of NLU or NLG.

Overall, although there have been a few endeavors to measure multilingual performance of LLMs, the evaluations done for SEA languages often only cover tasks for which multilingual datasets are more conveniently available, do not include linguistic diagnostics or cultural diagnostics, often use machine-translated prompts which are known to have translation errors that negatively impact models' performance, and also usually have not been tested on GPT-4.

## 3 BHASA: A NLP benchmark for SEA Languages

### 3.1 Overall Design

In order to fill in the gaps in the literature on LLM performance in SEA languages, we designed BHASA (**B**enchmark for **H**olistic Evaluation of Generative **A**I in **S**outheast **A**sian Languages),[4] a benchmark that aims to provide a holistic assessment of LLM abilities in SEA languages. BHASA comprises 3 major components (see Figure 1), (1) a NLP benchmark, (2) LINDSEA, a linguistic diagnostic dataset, and (3) a cultural diagnostic dataset. In this section (Section 3), we will look at the first component, the NLP benchmark.

For our NLP benchmark, we take inspiration from HELM [8] in developing a taxonomy of tasks before proceeding with the selection of tasks and datasets. We define key competencies of a language model as a trinity – Natural Language Understanding (NLU), Natural Language Generation (NLG) and Natural Language Reasoning (NLR). Although NLU and NLG have already been taxonomized quite well in other benchmarks, NLR is often not included as a category. We therefore created a NLR taxonomy based on existing works [23, 47, 48, 49, 50]. As can be seen from Figure 2, NLR tasks can be classified into two main categories – logical reasoning and higher-level reasoning. Logical reasoning can be further divided into deductive and non-deductive reasoning (which includes inductive, abductive and analogical reasoning). On the other hand, higher-level reasoning tasks do not target a single type of reasoning skill, but rather require a combination of them, sometimes requiring certain types of knowledge such as in commonsense reasoning. Natural Language Inference (NLI) and Question Answering are usually classified as NLU tasks, but they actually refer more to the task format than the skill tested [50]. As such, certain NLI or Question Answering tasks may fall under the umbrella of NLR if they target certain reasoning skills.

Unlike HELM [8], KoLA [51], BIG-Bench [7] and some other benchmarks, which explicitly test for the knowledge of language models, we were unable to include such tests in BHASA as datasets are scarce. For the time being, we have created a small dataset for testing cultural knowledge (see Section 5), but we intend to look into knowledge tasks for SEA languages in future iterations of BHASA.

We design our NLP benchmark to evaluate zero-shot performance of LLMs using both English and native prompts. As we want the benchmark to be of high quality, we ensure that prompts are translated manually by native speakers and that datasets chosen are sourced natively or at least professionally translated. The following sections will go further into detail about how we designed the benchmark and the rationale behind our decisions.

### 3.2 Language selection

While we would ideally have liked to include all official ASEAN languages in this benchmark evaluation, we had to accept the reality that many languages in SEA are low-resource languages and therefore did not have sufficient datasets for all languages. In the end, we decided to include Indonesian, Vietnamese, Thai and Tamil after considering both resource availability and typological diversity. We will extend the evaluation to other SEA languages gradually. Table 1 below summarizes the features of the four target languages.

### 3.3 Task and dataset curation

We selected the tasks for the benchmark based on whether there was a dataset of reasonable quality available for at least 3 of the 4 languages targeted. We selected datasets that contained data originally in the target language as far as possible and where translated datasets were used, we ensured that they were translated by professional translators or that elaborate effort has been shown to ensure that the translation was done accurately. This is important because translated datasets often contain translationese [53] which can differ significantly from natively written text [54, 55, 56, 57].

---

[4]The name of the benchmark is a play on the Sanskrit word भाषा ($bh\bar{a}\d{s}\bar{a}$), which means language, as a nod to the influence Sanskrit has had on many of the languages in Southeast Asia.

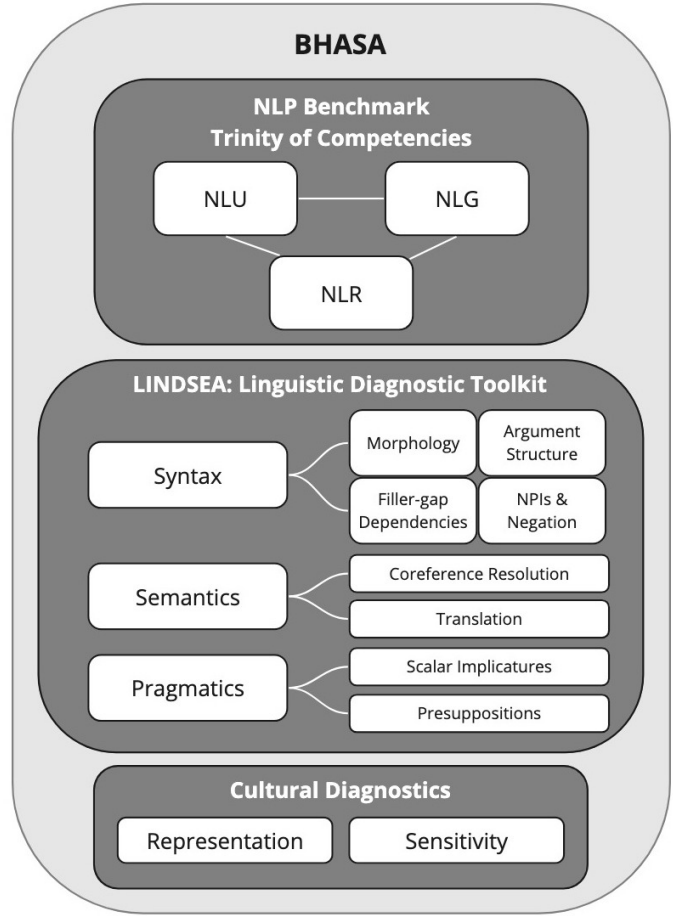

Figure 1: The BHASA benchmark and its three components

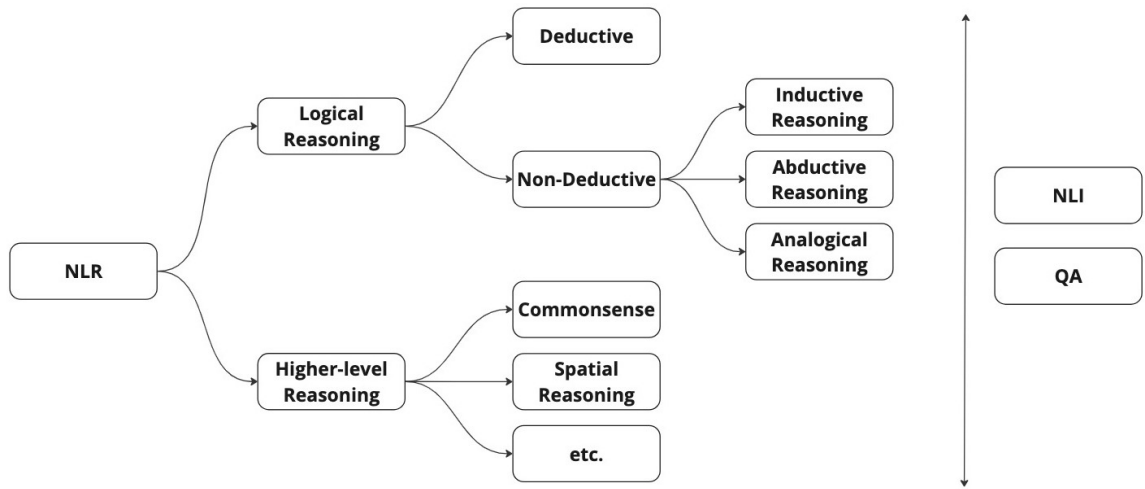

Figure 2: Taxonomy of Natural Language Reasoning tasks

At the end of the data collection and selection process, we finalized a set of 8 tasks with datasets available for all 4 languages with the exception of Toxicity Detection for which we were unable to find a Tamil dataset of reasonable quality. Table 2 below summarizes the tasks and the details of the respective datasets.

| | Code | Family | Script | | Word Order | Morphological Typology | Resource Class |
| | | | Name | Type | | | |
|---|---|---|---|---|---|---|---|
| Indonesian | id | Austronesian | Latin | Alphabet | SVO | Agglutinative | 3 |
| Vietnamese | vi | Austroasiatic | Modified Latin | Alphabet | SVO | Analytic | 4 |
| Thai | th | Tai-Kadai | Thai | Abugida | SVO | Analytic | 3 |
| Tamil | ta | Dravidian | Tamil | Abugida | SOV | Agglutinative | 3 |

Table 1: Linguistic details of the 4 target languages of BHASA. Resource classes are obtained from [52], with 5 being languages with the most resources (e.g. English) and 0 being languages with virtually no resources (e.g. Warlpiri)

| Competency | Task | Dataset | Language | Output | Metrics | Domain | Translation |
|---|---|---|---|---|---|---|---|
| NLU | NER | NERGrit | id | 3 classes | F1 | Wikipedia | Native |
| | | VLSP2018 | vi | 4 classes | F1 | News | Native |
| | | ThaiNERv2 | th | 13 classes | F1 | News | Native |
| | | Naamapadam | ta | 3 classes | F1 | General | Native |
| | QA | TyDiQA | id | span | F1 | Wikipedia | Native |
| | | XQuAD | vi, th | span | F1 | Wikipedia | PT |
| | | IndicQA | ta | span | F1 | Wikipedia | Native |
| | Sentiment Analysis | NusaX | id | 3 classes | F1 | Social Media | Native |
| | | UIT-VSFC | vi | 3 classes | F1 | Reviews | Native |
| | | Wisesight | th | 4 classes | F1 | Social Media | Native |
| | | IndicSentiment | ta | 2 classes | Accuracy | Reviews | PT |
| | Toxicity Detection | MLHSD | id | 3 classes | F1 | Twitter | Native |
| | | ViHSD | vi | 3 classes | F1 | Social Media | Native |
| | | Thai Toxicity Tweet | th | 2 classes | F1 | Twitter | Native |
| NLG | Machine Translation | FLORES | id, vi, th, ta | translation | ChrF++, COMET22, Human | News | PT |
| | Abstractive Summarization | XLSum | id, vi, th, ta | summary | ROUGE-L, Human | News | Native |
| NLR | NLI | IndoNLI | id | 3 classes | Accuracy | Wikipedia, News | Native |
| | | XNLI | vi, th | 3 classes | Accuracy | General | PT |
| | | IndicXNLI | ta | 3 classes | Accuracy | General | AT |
| | Causal Reasoning | XCOPA | id, vi, th, ta | 2 choices | Accuracy | General | PT |

Table 2: Details of the 8 tasks and their individual datasets in the BHASA benchmark

In the following subsections, we will describe each of the 8 benchmark tasks. The English prompt used for each task will be shown as well, but due to space constraints, all translated prompts have been placed in Appendix B (Table 46).

### 3.3.1 Named entity recognition (NER)

Named Entity Recognition (NER) is a sequence labeling task which involves detecting and classifying named entities in a text. The only multilingual dataset containing Southeast Asian languages that we know of is WikiAnn [58]. Despite the widespread use of WikiAnn in multilingual benchmarks such as XTREME [59] and, more recently, BUFFET [10], its quality as a benchmark dataset has been called into question [60], given that the data is automatically derived and was even termed as a "silver standard" by the original authors. In view of this, we opt for individual monolingual NER benchmark datasets used by the NLP communities of the respective languages.

For Indonesian, we included NER-Grit which was introduced in the IndoNLU benchmark [61], and for Vietnamese, we chose the VLSP 2018 NER Shared Task [62]. For Thai, we included ThaiNER (version 2.0) [63] which was also used in the evaluation of WangchanBERTa [64]. Finally, we extracted the Tamil subset of the Naamapadam multilingual NER dataset [65] for Tamil.

Although we endeavored to measure the zero-shot performance of the models without in-context learning (as stated in Section 3.1), we could not avoid providing a single example for the NER task in order to specify the required format of the model output. We modify GPT-NER's [66] prompt slightly to include the entity type in the output.

---

Prompt:

Perform Named Entity Recognition (NER) on the following text using only the following labels: {LABELS}.
Please follow the format given in the following example:
Example Text: Asia is a continent.
Example Answer: @@PLACE@@Asia## is a continent.
Text: {TEXT}
Answer:

---

Table 3: Prompt template for the NER task

### 3.3.2 Question answering (QA)

Question Answering (QA) is an important NLP task associated with many user-facing applications, such as chatbots and web search. However, despite its importance, the task is actually not well-defined and covers a wide range of tasks. In fact, any NLP task could arguably be reformulated as a QA task [67]. For the purposes of this paper, we define QA to refer strictly to the task of machine reading comprehension (MRC), a subset of textual QA, which requires a system to answer questions using a text as context. As reasoning skills will be tested under the NLR section of the benchmark, we did not endeavor to source for QA datasets that tested for reasoning skills (unlike HELM).

We follow existing work [9, 10, 12] in choosing the TyDi QA dataset [68] for Indonesian, XQuAD [69] for Vietnamese and Thai, as well as IndicQA [70] for Tamil. In the case of TyDi QA, we use the minimal answer span task instead of the passage selection task to align the task with XQuAD and IndicQA which are both minimal span extraction tasks as well. IndicQA also contains unanswerable questions for which the answer cannot be found within the provided text.

---

Prompt:

You will be given a paragraph and a question.
Answer the question by extracting the answer from the paragraph.
Paragraph: {TEXT}
Question: {QUESTION}
Answer:

---

Table 4: Prompt template for the QA task

### 3.3.3 Sentiment analysis

Sentiment Analysis is a NLU task that involves the identification of the sentiment of a text. It is highly relevant to many industrial applications such as for understanding market sentiment in financial settings or for understanding consumer sentiment towards products. Most datasets define sentiment labels as 'positive', 'negative' or 'neutral'. While there is also a more fine-grained sentiment analysis task known as aspect-based sentiment analysis which involves classifying the sentiment with respect to one or more aspects in a text, we decided to leave it for future work due to the lack of datasets and the cost of running additional evaluations.

We use the NusaX dataset [71] for Indonesian, the UIT-VSFC dataset [72] for Vietnamese, the Wisesight Sentiment dataset [73] for Thai and IndicSentiment dataset [70]for Tamil.

### 3.3.4 Toxicity detection

Toxicity Detection involves the identification of toxic content such as hate speech and abusive language in text. It is an increasingly important task in NLP due to the need for content moderation at scale, for example on social media platforms [8]. However, most studies have only focused on the Western world and there has hitherto been no other effort to evaluate LLMs on this task for SEA languages. We therefore seek

---

Prompt:

What is the sentiment of the following sentence?
{TEXT}
Answer only with a single word: Positive/Negative/Neutral

---

Table 5: Prompt template for the Sentiment Analysis task

to understand whether LLMs are able to perform toxicity detection in SEA languages given that toxicity triggers can differ between Western and Eastern contexts [74].

We use the Indonesian Twitter Multi-label Hate Speech and Abusive Language Detection Dataset (henceforth the MLHSD dataset) [75] for Indonesian, the ViHSD Dataset [76] for Vietnamese, and Thai Toxicity Detection Dataset [77] for Thai. As the MLHSD dataset was designed as a multi-label classification task as opposed to the multi-class classification task in Vietnamese and Thai, we opted to map the labels for the MLHSD dataset such that every text can only be one of three labels, 'Clean', 'Abusive' or 'Hate', in order to better align the task across the three languages.

As the definition of toxicity can differ across datasets, we consulted the paper accompanying each dataset to ensure that we define each label in the prompt according to the definitions stipulated by the paper authors.

---

Prompt:

You are a hate speech detector. The label definitions are as follows:
Clean: No harassment.
Abusive: Involves harassment and even profanity, but does not attack any specific object.
Hate: Directly harasses or abuses a specific object.
When given a sentence, classify it with a single word: Clean/Abusive/Hate.
Sentence: {TEXT}

---

Table 6: Prompt template for the Toxicity Detection task

### 3.3.5 Machine translation

LLMs have demonstrated great potential in machine translation [30], but mainly only for high-resource languages, while the performance for low-resource languages still lags far behind supervised baselines [11, 30, 78]. Given OpenAI's claim that GPT-4 has improved multilingual performance compared to GPT-3.5 [15], it would be important for us to verify its machine translation abilities, especially on SEA languages which are generally considered to be relatively low-resource languages. We use the FLORES-200 dataset [79] for this task as it contains all four of our target languages and is a parallel dataset, allowing for fairer comparisons across languages. Stringent processes were put in place for quality control and for the selection of translators and reviewers to ensure the quality of the dataset, hence our decision to use it.

---

Prompt:

Translate the following text into {LANGUAGE}.
Text: {TEXT}
Translation:

---

Table 7: Prompt template for the Machine Translation task

#### 3.3.6 Abstractive summarization

Abstractive Summarization requires a system to read a document, identify the key points within, and summarize them into a coherent and fluent text while paraphrasing the document. It is a much more challenging task compared to Extractive Summarization which involves extracting the most salient span(s) of text in a document verbatim, and therefore will require much stronger NLG abilities. For this task, we chose the XLSum dataset [80] which covers all 4 of our target languages.

---

Prompt:

Article: {TEXT}
Summarize this {LANGUAGE} article in 1 or 2 sentences. The response should be in {LANGUAGE}.
Summary:

---

Table 8: Prompt template for the Abstractive Summarization task

#### 3.3.7 Natural language inference (NLI)

NLI is a task that involves determining whether a given premise entails or contradicts a hypothesis, typically using a 3-way classification scheme with the labels 'entailment', 'contradiction' and 'neutral'. It is a rather vague task that is defined more by its format than its content. For example, the SNLI dataset has been analyzed to contain three major categories of premise-hypothesis entailment problems, namely paraphrasing, compound semantics understanding, and reasoning with implicit premises [50]. This has caused NLI to be regarded both as a NLU and as a NLR task. Critics have commented that the use of the label 'entailment' is inappropriate given that its technical definition in linguistics and logic requires that the hypothesis follow necessarily from the premise in all situations [81]. This is not the case in many NLI datasets since they often involve presuppositions, implicatures, and world knowledge in addition to logical entailment [82]. Nevertheless, we categorize NLI as a higher-level reasoning task under NLR, acknowledging that multiple skills are being tested and carry out more pinpointed evaluations of reasoning skills subsequently using our own handcrafted datasets (see Section 4).

For the NLI task, we choose the IndoNLI lay-authored dataset [83] for Indonesian, XNLI [84] for Vietnamese and Thai, and IndicXNLI [85] for Tamil. While translating the prompt template for the task into our target languages, we found that the terms *premise*, *hypothesis*, *entailment*, *contradiction* and *neutral* are not easily translatable into our target languages and could cause confusion for the models. As such, we decided to design the prompt template in a way that avoids using those words and that uses simple words in replacement.

---

Prompt:

You will be given two sentences, X and Y.
X: {PREMISE}
Y: {HYPOTHESIS}
Determine which of the following statements applies to sentences X and Y the best.
A: If X is true, Y must be true.
B: X contradicts Y.
C: When X is true, Y may or may not be true.
Answer strictly with a single letter A, B or C.

---

Table 9: Prompt template for the NLI task

### 3.3.8 Causal reasoning

Causal Reasoning is a subset of reasoning that involves identifying and understanding the causes and effects of events or phenomena [47]. It is often considered to require commonsense knowledge [86, 87, 50]. We therefore classify it as higher-level reasoning under the NLR category.

We test the models on both forward and backward causal reasoning (effect and cause respectively) using the multilingual XCOPA dataset [87] for all target languages.

---

Prompt:

Situation: {PREMISE}
Given this situation, which of the following choices is most likely to be its {QUESTION}?
A: {CHOICE 1}
B: {CHOICE 2}
Respond strictly with the letters A or B only.

{QUESTION} = cause/effect

---

Table 10: Prompt template for the Causal Reasoning task

## 3.4 Experimental setup

### 3.4.1 Models

In this paper, we focus on evaluating GPT-4 as it has been reported to achieve strong 3-shot performances on a machine-translated MMLU benchmark given its multilingual capabilities [15]. We also include GPT-3.5-Turbo to observe the purported improvements in performance from GPT-3.5-Turbo to GPT-4.

The GPT-3.5-Turbo model used was gpt-3.5-turbo-0301 and the GPT-4 model used was gpt-4-0314. The two models have a context length of 4096 and 8192 tokens respectively. Disregarding this difference in context length, we evaluate both models fairly by truncating the input where necessary as if both models only had a context length of 4096.

We set the temperature to be 0 for all NLU and NLR tasks to maximize the stability and reproducibility of the generations. As for NLG tasks, we set the temperature to be 0 and 0.3 for Machine Translation and Abstractive Summarization respectively in line with findings from previous works [88, 89, 35]. We use a top_p of 1, frequency_penalty of 0 and presence_penalty of 0 as per OpenAI's defaults.

### 3.4.2 Prompts

The sensitivity of LLMs to prompts has been demonstrated in multiple studies, leading to the birth of a new research direction known as prompt engineering. Prompt engineering papers have proposed various improvements to prompts, such as the use of few-shot examples [1], chain-of-thought prompting [90, 6, 91, 92], and even automatic generation of prompts [93].

In our paper, we avoid the use of few-shot examples as we want to test the zero-shot capabilities of LLMs without any in-context learning. Furthermore, the use of few-shot examples introduces a plethora of variables which can have significant impact on model performance, such as the number of examples [94, 95], the choice of examples [94, 96], the language of the examples [9], and the order of the examples [97, 98]. We therefore leave the exploration of these options to future work.

Instead, we focus our efforts on the translation of prompts and test the models with both English and target language prompts in zero-shot settings. Existing benchmarking efforts often use machine-translated prompts [9, 10, 12] but this has been reported to elicit drops in model performance due to errors in translation [9]. As such, we worked with native speakers to translate the English prompts into their respective languages in order to ensure the accuracy of the prompts and to maintain, as far as possible, the same meaning across all languages. However, we found that the translation exercise was more challenging than expected because there were many terms that did not translate well into our target languages and there were certain idiosyncrasies of these languages that had to be taken into consideration before deciding on a particular translation.

Firstly, certain task-related keywords do not have exact equivalents in our target languages. A case in point would be the words *premise*, *hypothesis*, *entailment* and *neutral* in the NLI task. These words also have a slightly different meaning in the context of NLI, which would make a direct translation of these terms inappropriate. We deal with this issue by doing away with these opaque terms, opting instead to paraphrase them using layman terms for transparency. For example, entailment is paraphrased as "If X is true, Y must be true".

Secondly, certain words can have multiple translations with slightly different nuances and the choice of translation could impact the way the model responds. For example, there are two ways to translate the word summary in both Indonesian and Tamil. *Ringkasan* and சுருக்கம் (*curukkam*) refer to a shortened text while *rangkuman* and பொழிப்பு (*polippu*) refer to a text that is shortened in a systematic way to retain its essence and are the words usually used in the context of a language examination.

Finally, we found that even the way the user addresses the model can affect the way it responds. For example, when asking the model to do anything in Tamil, which is morphologically rich and exhibits subject-verb agreement, the user needs to decide whether to address the model more respectfully or more informally as that will affect the inflection of the verb used.[5]

These issues apply not only to our target languages but also to any other language, especially those that are typologically or culturally distant from English. We therefore encourage researchers and practitioners to take these into consideration when prompting language models natively in non-English languages.

A full list of prompts in every language used can be found in Appendix B (Table 46).

### 3.4.3 Data cleaning and sampling

**Data cleaning**  Before using the datasets for the benchmark, we did exploratory data analysis for each dataset to understand the quality and distribution of the data. We had to drop certain data points due to missing labels or text. This included 1 sentence in the IndicSentiment dataset which had no label (reducing the dataset size from 1000 sentences to 999 sentences), and 1140 tweets in the Thai Toxicity Detection dataset which either had no labels or had "TWEET_NOT_FOUND" as the text (reducing the dataset size from 3300 tweets to 2160 tweets).

Furthermore, we discovered an alarming number of mislabelled sentences in the XCOPA dataset. While there is generally an equal distribution of questions across languages, with 250 cause and 250 effect questions in each test set, the Indonesian and Thai datasets had 254 and all 500 out of 500 instances asking for the effect respectively. Upon further inspection, we determined that the 4 questions in the Indonesian dataset were indeed mislabelled and proceeded to relabel them as 'cause' in line with the other datasets. The same was done for the Thai dataset to get 250 cause and 250 effect questions (see Table 48 in Appendix C for the 4 Indonesian instances as well as a sample of 4 mislabelled Thai instances). This was a surprising finding as the XCOPA dataset was professionally translated and had near perfect inter-annotator agreement with efforts to ensure the quality of the dataset [87]. We investigated all other test datasets in XCOPA and noted that, in total, 5 out of 11 of the datasets had the same issue to differing degrees (see Table 11). Even more concerning was the fact that there has been no study that we know of that reported these errors, making it unclear whether studies using XCOPA are reporting metrics on the original erroneous dataset or if they also cleaned the dataset before testing. This highlights the importance of scrutinizing datasets used for benchmarking, even if they have already been used widely.

**Data sampling**  Due to limited resources, we sample all datasets to a maximum of 1000 instances (documents or sentences) for all tasks except for Machine Translation, for which we use the full test dataset of 1012 sentences, as well as for Question Answering and Abstractive Summarization, for which we use a sample of 100 documents given their length (see Table 12). Sampling was done randomly but checks were done to ensure that the distributions of labels and document lengths were preserved in the samples. Furthermore, the document lengths for IndicQA and XLSum (Thai and Tamil datasets) had to be clipped to 3500 and 3300 tokens[6] respectively due to the context length of gpt-3.5-turbo-0301 being capped at 4096.

---

[5]For example, the verb *answer* in the phrase "please answer with a single letter A or B" can be translated as பதிலளி (*patilaḷi*) or பதிலளியுங்கள் (*patilaḷiyunkaḷ*) depending on whether we are treating the model with less or more formality respectively. Furthermore, பதிலளிக்கவும் (*patilaḷikkavum*) and பதிலளிக்குக (*patilaḷikkuka*) can also be used as an imperative that is more obligatory and is often used in instructions.

[6]Tokenization was done using OpenAI's cl100k_base tokenizer (which is used in GPT-4 and GPT-3.5-Turbo).

|  | Cause | Effect |
|---|---|---|
| Chinese | 250 | 250 |
| Estonian | 250 | 250 |
| Haitian Creole | 250 | 250 |
| **Indonesian** | **246** | **254** |
| **Italian** | **246** | **254** |
| Quechua | 250 | 250 |
| **Swedish** | **213** | **287** |
| Tamil | 250 | 250 |
| **Thai** | **0** | **500** |
| **Turkish** | **199** | **301** |
| Vietnamese | 250 | 250 |

Table 11: Number of questions per language asking for cause and effect in XCOPA datasets. Every language should have 250 questions each for cause and effect.

| Competency | Task | Dataset | Language | Set | Total | Sample | Trimming |
|---|---|---|---|---|---|---|---|
| NLU | NER | NERgrit | id | Test | 209 | 209 | - |
| | | VLSP2018 | vi | Test | 8454 | 1000 | - |
| | | ThaiNERv2 | th | Test | 1472 | 1000 | - |
| | | Naamapadam | ta | Test | 758 | 758 | - |
| | QA | TyDiQA | id | Dev | 565 | 100 | - |
| | | XQuAD | vi, th | Test | 1190 | 100 | - |
| | | IndicQA | ta | Test | 1804 | 100 | 3500 |
| | Sentiment Analysis | NusaX | id | Test | 400 | 400 | - |
| | | UIT-VSFC | vi | Test | 3166 | 1000 | - |
| | | Wisesight | th | Test | 2671 | 1000 | - |
| | | IndicSentiment | ta | Test | 999 | 999 | - |
| | Toxicity Detection | MLHSD | id | Test | 13169 | 1000 | - |
| | | ViHSD | vi | Test | 6680 | 1000 | - |
| | | Thai Toxicity Tweet | th | Test | 2160 | 1000 | - |
| NLG | Machine Translation | FLORES | id, vi, th, ta | Devtest | 1012 | 1012 | - |
| | Abstractive Summarization | XLSum | id | Test | 4780 | 100 | - |
| | | | vi | Test | 4013 | 100 | - |
| | | | th | Test | 826 | 100 | 3300 |
| | | | ta | Test | 2027 | 100 | 3300 |
| NLR | NLI | IndoNLI | id | Test | 2201 | 1000 | - |
| | | XNLI | vi, th | Test | 5010 | 1000 | - |
| | | IndicXNLI | ta | Test | 5010 | 1000 | - |
| | Causal Reasoning | XCOPA | id, vi, th, ta | Test | 500 | 500 | - |

Table 12: Test dataset details for BHASA's NLP benchmark. Trimming refers to the number of tokens that input documents were truncated to (using OpenAI's cl100k_base tokenizer).

## 3.5 Evaluation

### 3.5.1 Metrics

In terms of metrics, we believe that the method adopted in HELM [8] to measure 7 categories of metrics (Accuracy, Calibration, Robustness, Fairness, Bias, Toxicity and Efficiency) would be ideal. However, given constraints in resources, we opted to focus solely on accuracy metrics for this study and leave the rest for future works. Accuracy metrics for each task were selected based on convention, with F1 score or accuracy being used for most tasks. However, as automated metrics for NLG may not correlate well with human judgments, be uninterpretable, have inherent biases and be unable to capture certain task-specific nuances [99], we include human evaluation on a small sample size to provide additional qualitative insights on the models' NLG performance.

This is especially important for Abstractive Summarization which conventionally uses the ROUGE metric [100] because the metric was recently reported to be challenging to compare across studies due to code errors in evaluation packages and differences in evaluation protocol which are not clearly explained [101]. We still include the metric in line with convention but we make our evaluation parameters transparent. We chose to use XLSum's repository's multilingual ROUGE scoring package[7] as it is used to evaluate models tested on XLSum. We used all defaults and also specified the use of PyThaiNLP's *newmm* tokenizer for the evaluation of Thai summarization. Stemming and stopword removal was not performed. The $\beta$ for the $F_\beta$ score was set at 1 as per the default. As for the bootstrap aggregation, we set 4156 as the random *numpy* seed for reproducibility and used the 'mid' result (i.e. mean) as the final score.

For the task of Machine Translation, we follow the recommendation of the WMT22 Metrics Shared Task to employ neural network-based metrics as they are more aligned with human evaluation and robust to different domains [102]. We adopted COMET as our neural network-based metric and ChrF++ [103] as our overlap metric as they were both used as the official metrics for the WMT Shared Task 2022 [104]. However, instead of using COMET-20 [105] which was used in the WMT Shared Task 2022, we opted for the improved COMET-22 metric [32] as it is one of the top performing metrics in the WMT22 Metrics Shared Task [102], surpassing the performance of COMET-20.

### 3.5.2 Human evaluation

As mentioned in Section 3.5.1, we perform human evaluation as a sanity check for the chosen automated metrics as well as to provide qualitative insights. For each target language, 3 lay annotators who are native speakers of the respective target languages and have a high proficiency in English were recruited. Data was sampled randomly to obtain 20 sentences for each translation direction as well as 10 articles for summarization. Annotators were presented with 5 translations for each sentence and 5 summaries for each article (4 responses from GPT-3.5-Turbo and GPT-4 using both English and translated prompts, and one reference) and were asked to evaluate them according to our guidelines.

The guidelines for Abstractive Summarization were adapted from various works [106, 37, 35] and require the annotators to score summaries based on relevance, faithfulness of content, fluency, and coherence (see Appendix D for more details). Relevance, fluency and coherence are scored on a Likert scale of 1 to 5, while faithfulness is scored binarily as 0 (not faithful) or 1 (faithful). On the other hand, the guidelines for the Machine Translation task were modified from the Multidimensional Quality Metrics (MQM) framework [107, 108]. Annotators are asked to look out for errors in accuracy, terminology, linguistic conventions, style and locale conventions (see Appendix D for more details), and to mark translations as incorrect if any of those errors exists. Annotators have to note down the type of errors present and are also encouraged to add in comments to explain the error. While the use of MQM in [108] involves highlighting the relevant span of text and tagging it with the error type and severity, we decided to forego these steps and simplify the task to reduce the task complexity for the annotators, especially since they are not professional translators.

### 3.5.3 LLMs as NLG evaluators?

Despite the rise in popularity of LLM-based evaluations [109, 110, 111, 112, 113, 114, 115, 116, 117] as a way to scale up evaluations at a lower cost, we decided not to adopt this approach for a number of reasons.

Firstly, given the fact that the reliability or effectiveness of LLMs as evaluators has not been demonstrated for SEA languages, we might not be able to trust the results of the evaluation. Current studies mostly focus on evaluating English outputs, and the effectiveness of LLMs in evaluating machine translation has only been explored for Chinese, German and Russian [113].

Secondly, it has been shown that LLMs can have bias towards both themselves and other models [109, 117], making their judgments less reliable. Furthermore, although GPT-4 has been found to be the best judge in multiple studies [113, 114, 117], it would be undesirable for us to use it since we are also evaluating GPT-4 itself. After all, *quis custodiet ipsos custodes*?[8]

---

[7] https://github.com/csebuetnlp/xl-sum/tree/master/multilingual_rouge_scoring

[8] "*Quis custodiet ipsos custodes*?" originates from the Roman satirical work, the Satires, and means "Who will watch the watchmen?"

Finally, LLMs also exhibit various other biases such as position bias (where the order of presenting options significantly alters judgment) [116] and verbosity bias (where the output length can alter judgment regardless of text quality) [117]. There is also uncertainty in how prompts should be written to ensure reliable judgments.

Because of this, we eschew the use of LLMs to evaluate NLG and rely only on automated metrics and human evaluation for this paper.

## 3.6 Results and analysis

In general, we find that the models' performances vary across languages and decrease in the order of Indonesian, Vietnamese, Thai and Tamil. Both models are also less able to follow instructions when prompted in Thai or Tamil. Furthermore, in general, we find that GPT-4 does indeed outperform GPT-3.5-Turbo significantly on our benchmarking tasks and target languages. This gap is even more distinct in the translated prompt setting where GPT-3.5-Turbo has a higher rate of not being able to follow the instructions or answering wrongly.

In the following sections we look at the performance of the models on each task separately.

### 3.6.1 Named entity recognition (NER)

|  | id | vi | th | ta |
| --- | --- | --- | --- | --- |
| Fine-tuned | 79.09 (XLM-R) | 77.52 (VNER) | 84.80 (WangchanBERTa) | 64.5 (MuRIL) |
| GPT-3.5-Turbo | 31.76 | 16.85 | 15.87 | 21.65 |
| GPT-3.5-Turbo (T) | 33.88 | 23.64 | 10.71 | 9.55 |
| GPT-4 | **62.05** | **60.96** | **38.92** | **51.84** |
| GPT-4 (T) | 60.47 | 50.00 | 29.03 | 36.78 |

Table 13: Strict NER F1 scores of GPT-3.5-Turbo and GPT-4. The performances of fine-tuned models were obtained from [61] for Indonesian, [62] for Vietnamese, [64] for Thai, and [70] for Tamil.

Both models generally did not do well for the NER task and this was generally due to the output not being in the required format. Although the prompt requires the model to label entities within the text by delimiting it with the @@ and ## symbols, these symbols were sometimes left out or used partially. The models may also remove the whitespaces between tokens, merging them and causing the number of tokens of the prediction to differ from that of the ground truth. This is understandable as the task of NER is that of sequence labeling which is not very compatible with the text generation task that LLMs are trained on [66].

Another factor contributing to the poor performance is likely related to the difficulty in specifying the definition of each class in the prompt. The model has to rely on transfer learning from whatever data it has previously been trained on to deduce what the classes in the prompt refer to, but the model is often mistaken. For example, pronouns are sometimes labeled as PERSON.

Nevertheless, GPT-4 performs significantly better than GPT-3.5-Turbo in all 4 languages. Translated prompts result in large drops in performance across the board except for GPT-3.5-Turbo when prompted in Indonesian or Vietnamese, in which case performance actually improves unexpectedly. The drop in performance is likely due to the translated labels being unfamiliar to the model as NER datasets do not usually use labels written in the target language. Interestingly, the drop in performance of GPT-4 with Indonesian prompts is much smaller compared to prompts in other languages, suggesting that its abilities in understanding Indonesian is much better.

Overall, GPT-3.5-Turbo and GPT-4 are probably not suitable for use in a zero-shot setting for NER in these 4 languages. In general, researchers and practitioners wanting to use LLMs for NER may consider adding examples and a self-verification step [66] for better performances.

|  | id | vi | th | ta |
|---|---|---|---|---|
| Human | 92.00 | - | - | - |
| Fine-tuned | 84.80 (mBERT) | 79.10 (XLM-R Large) | 74.20 (XLM-R Large) | 45.40 (IndicBERTv2) |
| GPT-3.5-Turbo | 48.45 | 39.47 | 45.34 (59.60) | 28.88 |
| GPT-3.5-Turbo (T) | 45.58 | 44.53 | 39.00 (49.88) | 13.28 |
| GPT-4 | **76.89** | **73.85** | **75.30 (83.69)** | **34.71** |
| GPT-4 (T) | 73.60 | 72.02 | 64.41 (75.19) | 34.38 |

Table 14: F1 score of GPT-3.5-Turbo and GPT-4 for Question Answering. Fine-tuned model and human performances were obtained from [68] for Indonesian, [69] for Vietnamese and Thai, and [70] for Tamil. Results on the Thai dataset in parentheses refer to results obtained after appropriate tokenization.

### 3.6.2 Question answering

GPT-4 performs significantly better than GPT-3.5-Turbo for Question Answering in terms of F1, and this is largely due to its ability to extract strictly the most relevant span for the answer, which also results in high Exact Match scores. On the other hand, GPT-3.5-Turbo tends to extract an entire sentence and does not answer to the point and therefore loses out on precision. Nevertheless, post-analyses revealed that 80% to 90% of these sentences extracted by GPT-3.5-Turbo do indeed contain the relevant answer span, so the model does understand the document and the question, but simply fails to answer the question directly by excluding the irrelevant portions.

Due to the fact that the SQuAD evaluation script tokenizes model predictions using whitespace, we had to retokenize model responses for Thai for a fairer evaluation. Since Thai is a *scriptio continua* language that does not separate words with spaces, not tokenizing the model responses would mean calculating F1 on a sentence level and over-penalizing the model. The F1 scores are much higher after retokenization using PyThaiNLP's *newmm* tokenizer and we report these results in parentheses beside the initial results. Surprisingly, GPT-4 in a zero-shot setting can outperform a fine-tuned XLM-R Large model on the Thai XQuAD dataset. In the other languages however, GPT-4 still lags slightly behind smaller fine-tuned models.

This gap in performance is especially conspicuous for the Tamil IndicQA dataset where the results seem quite dismal and we sought to investigate the reasons behind it. IndicQA is intrinsically harder than the other datasets due to it containing unanswerable questions, and this was likely partially the reason for the low F1 scores. Out of 28 unanswerable questions, GPT-3.5-Turbo managed to identify them as such for 18 and 0 questions with English and translated prompts respectively, while GPT-4 managed to do so for 20 and 13 questions with English and translated prompts respectively. Looking into individual question responses however, we find that the model actually performed better than what was suggested by the metrics. In some cases, the noun in the model's answer was inflected differently from that in the ground truth and was not considered a match. For example, one of the questions asked "who built the Ram Jhula bridge", and the ground truth read "by the Indian government" (using the instrumental case to mark the noun as an agent in a passive voice construction) while GPT-4 answered "the Indian government" (see Example 1 and 2 for the linguistic glossing). GPT-4's answer is arguably more accurate since the question was posed in the active voice and not passive voice. In fact, it is rather impressive that it was able to change the inflection of the noun into the nominative case to better answer the question. However, as the task is supposed to be extractive instead of abstractive, we accept the ground truth as the correct answer. Nevertheless, we might want to explore lemmatization as a fairer way to evaluate models on such languages which are morphologically rich.

(1) இந்திய அரசாங்கத்தால்
Intiya aracāṅkatt-āl
India government-INST
'By the Indian government'

(2) இந்திய அரசாங்கம்
    Intiya    aracāṅkam
    India    government
    'The Indian government'

We also found that some questions were not phrased correctly or their answers were inaccurate. For example, one question asked for the year that the Indian National Committee for Space Research (INCOSPAR) was founded and the ground truth was "1945". However, this fact was neither in the text given to the model nor even factually correct (INCOSPAR was founded in 1962). Another question asked for the name of one of the lakes dug during the Pallava dynasty. In the corresponding text, a long list of lakes was provided and the ground truth chose the first three lakes in the list as the answer. The question is not well-framed because any of the lakes could be the answer, and the ground truth is not even answering the question correctly since it gave three names instead of just one. GPT-4 actually did answer correctly by responding with the first lake in the list. These errors in the dataset may explain why the performance of fine-tuned IndicBERTv2 [70] was not able to attain even 50% in F1 score.

### 3.6.3 Sentiment analysis

|  | id | vi | th |
|---|---|---|---|
| Fine-tuned | 91.60 (XLM-R Large) | 87.94 (MaxEnt) | 67.05 (WangchanBERTa) |
| GPT-3.5-Turbo | 82.60 | 63.80 | **44.09** |
| GPT-3.5-Turbo (T) | 79.53 | 58.03 | 40.71 |
| GPT-4 | **84.28** | 60.02 | 40.07 |
| GPT-4 (T) | 74.14 | **65.19** | 39.15 |

Table 15: Macro-F1 scores of GPT-3.5-Turbo and GPT-4 on the Sentiment Analysis task for Indonesian, Vietnamese and Thai. Fine-tuned model performances are obtained from [71] for Indonesian, [72] for Vietnamese and [64] for Thai.

|  | ta |
|---|---|
| Fine-tuned | 93.60 |
| GPT-3.5-Turbo | 77.18 |
| GPT-3.5-Turbo (T) | 23.62 |
| GPT-4 | **83.98** |
| GPT-4 (T) | 71.97 |

Table 16: Accuracy of GPT-3.5-Turbo and GPT-4 on the IndicSentiment dataset. The fine-tuned model reported here is IndicBERTv2 [70].

The zero-shot performance of GPT-4 and GPT-3.5-Turbo varies a lot from language to language. Both models did well for the Indonesian NusaX dataset, with GPT-4 performing slightly better with a lead of 1.68% F1 score when prompted in English. It was surprising however to note that GPT-4 performed worse than GPT-3.5-Turbo when prompted in Indonesian, given that GPT-4 seems to understand Indonesian better in other tasks. It was found that a large contributor to this difference in performance came from misclassifying neutral sentences. 20 out of 96 neutral sentences (20.83%) were classified as positive by GPT-4, and they are generally advice or statements describing a situation in a neutral way. In order to do error analysis, we asked GPT-4 to explain its decisions. For sentences which are advice, GPT-4 explained that the advice is good for the recipient and therefore has a positive sentiment, whereas for statements, it tends to interpret the situation and give its own perspective on the matter (see Table 17). While this does demonstrate GPT-4's stronger Indonesian capabilities, it can be undesirable for the model to inject such judgments depending on the context. To reduce the chances of this happening, we suggest that researchers and practitioners can consider using more informative prompts to define the labels more clearly.

| Sentence | Explanation by GPT-4 |
|---|---|
| (Advice)<br><br>Jangan lupa minum air mineral 8 − 10 gelas per harinya. | Kalimat tersebut memiliki sentimen positif<br>karena memberikan saran atau nasihat yang baik untuk menjaga kesehatan,<br>yaitu dengan mengingatkan untuk minum air mineral 8-10 gelas per hari.<br>Hal ini menunjukkan perhatian terhadap kesehatan dan kesejahteraan orang lain. |
| Do not forget to drink 8 − 10 glasses of mineral water per day. | The sentence has a positive sentiment<br>because it gives a good suggestion or advice to take care of one's health,<br>that is to say by reminding one to drink 8-10 glasses of mineral water a day.<br>This shows concern for the health and well-being of others. |
| (Statement)<br><br>Kantor penghubung korsel-korut dibuka pekan ini. | Kalimat tersebut menunjukkan adanya pembukaan kantor penghubung<br>antara Korea Selatan dan Korea Utara, yang merupakan langkah maju<br>dalam hubungan antar negara tersebut. Hal ini dapat dianggap sebagai<br>tanda peningkatan kerja sama dan komunikasi antara kedua negara,<br>sehingga sentimennya positif. |
| The North Korea-South Korea Liaison Office opened this week. | The sentence indicates the opening of a liaison office between North Korea<br>and South Korea, which is a step forward in the relations between said countries.<br>This can be considered a sign of increasing cooperation and communication<br>between the two countries, so the sentiment is positive. |

Table 17: Explanations given by GPT-4 on why it labeled neutral sentences as having a positive sentiment when prompted in Indonesian.

As for the Vietnamese dataset, it seems that GPT-4 does worse than GPT-3.5-Turbo when prompted in English but does better when prompted in Vietnamese. It was found through error analysis that GPT-4 performed worse when prompted in English because it classified 55 out of 445 (12.36%) negative sentences as neutral. Many of these sentences contain modal verbs such as *nên* ('should') and *cần* ('need') or are imperative in nature. When prompted in Vietnamese, the understanding of these modals and imperatives as presupposing the utterer's dissatisfaction with the current state of affairs helps the model make the right decision. However, when prompted in English, the model prefers to view such suggestions as neutral (see Table 18).

Performance in Thai is poor across all runs and this was mainly due to the misclassification of neutral sentences. This is likely an issue of label definition as many of these neutral sentences are actually advertisements which sound positive but are classified as neutral in the ground truth as they are not product reviews. Another factor contributing to the poor performance is the fact that there are 4 classes for the Thai Wisesight Sentiment dataset, including the class 'Question' on top of the usual 'Positive', 'Negative' and 'Neutral'. GPT-4 in particular tends to label statements as questions, with 212 and 169 false positives when prompted in English and Thai respectively. It seems that the model labels rhetorical questions and echo questions as questions even if they are not meant to be interpreted literally as such. However, it has to be acknowledged that specifying the definition of the question label could help steer the model's responses towards the intended interpretation, and perhaps excluding the question label could result in a better comparison across languages and datasets. We leave these explorations for future iterations of BHASA.

Finally, the models performed reasonably well for the Tamil dataset, although that is perhaps to be expected given that only two classes, 'Positive' and 'Negative', were involved. Despite the explicit instruction to answer with only one of the two labels, both models answered "Neutral" in more than 50 instances. This could be a sign that the models had been strongly instruction-tuned in English to interpret the sentiment analysis task as one that has three classes. The huge jump in performance from GPT-3.5-Turbo to GPT-4 when prompted in Tamil can be attributed to GPT-4's ability to understand the Tamil instructions better and to answer with either label, whereas GPT-3.5-Turbo fails to do so and generates long responses.

### 3.6.4 Toxicity detection

Toxicity Detection is a more challenging task for the models as the data comes from social media and is in an informal register, thus the vocabulary can be quite different from the standard language. Furthermore, a good understanding of the cultures associated with the language is required as well to interpret potentially harmful messages accurately. There is generally an improvement in the toxicity detection abilities of the model from GPT-3.5-Turbo to GPT-4 except in the case of Thai. Through error analysis, it was found that this unexpected drop in F1 score for GPT-4 was due to an increase in false positives. As the Thai dataset is designed to be challenging by including polysemous words that may be innocuous or abusive depending on context, it is perhaps the case that GPT-4 errs on the side of caution by flagging out sentences that contain such words even if they may be harmless in those contexts.

| Sentence | Explanation by GPT-4 |
|---|---|
| | (English prompt) |
| Bài tập thầy nên sát với đề thi hơn. | The sentence "bài tập thầy nên sát với đề thi hơn" |
| | translates to "the teacher's exercises should be closer to the exam questions." |
| The teacher's exercises should be closer to the examinations. | This statement is a suggestion or recommendation, and it does not express a strong positive or negative emotion. |
| | Therefore, the sentiment can be considered neutral. |
| | |
| | (Vietnamese prompt) |
| | Câu "bài tập thầy nên sát với đề thi hơn" |
| | mang sắc thái tiêu cực vì nó ngụ ý |
| | rằng bài tập hiện tại do thầy đưa ra chưa |
| | đạt yêu cầu hoặc chưa phù hợp với đề thi. |
| | Điều này cho thấy sự không hài lòng |
| | của người nói và muốn thấy cải thiện trong việc lựa chọn bài tập. |
| | |
| | The sentence "the teacher's exercises should be closer to the examinations" |
| | has a negative sentiment because it implies that the current exercises given |
| | by the teacher are not satisfactory or unsuitable for the examinations. |
| | This shows the speaker's dissatisfaction and desire for the teacher |
| | to improve their choice of exercises. |

Table 18: Explanations given by GPT-4 on why it labeled negative sentences as having a neutral sentiment. Note that it labels the sentence as negative when prompted in Vietnamese.

| | id | vi | th |
|---|---|---|---|
| Fine-tuned | - | 62.69 (mBERT) | 78.00 (Logistic Regression) |
| GPT-3.5-Turbo | 53.99 | 41.61 | 64.36 |
| GPT-3.5-Turbo (T) | 49.25 | 40.48 | 60.92 |
| GPT-4 | 57.53 | **51.52** | 61.51 |
| GPT-4 (T) | **63.96** | 51.23 | **70.67** |

Table 19: Macro-F1 scores of GPT-3.5-Turbo and GPT-4 on Toxicity Detection. Fine-tuned model performances were obtained from [76] for Vietnamese and [77] for Thai.

An interesting observation is that there is significant improvement in the performance of GPT-4 when prompted in translated prompts for the Indonesian and Thai datasets. It is perhaps the case that having the entire prompt in the native language could help guide the model towards a more culturally appropriate interpretation of the texts instead of using an Anglophone perspective that might be associated with using English.

Overall, however, there seems to be much room for improvement in terms of understanding hate or toxic speech in Indonesian, Vietnamese and Thai.

### 3.6.5 Machine translation

| | id→en | | | vi→en | | |
|---|---|---|---|---|---|---|
| | ChrF++ | COMET-22 | Human | ChrF++ | COMET-22 | Human |
| Reference | - | - | 3.0/20 | - | - | 5.7/20 |
| GPT-3.5-Turbo | 66.05 | 89.52 | 13.3/20 | 58.85 | 87.25 | 11.7/20 |
| GPT-3.5-Turbo (T) | 66.27 | 89.70 | **14.3/20** | 58.94 | 87.37 | 9.7/20 |
| GPT-4 | 67.15 | 89.85 | 13.0/20 | 60.29 | 87.81 | 12.7/20 |
| GPT-4 (T) | **67.53** | **89.95** | 11.7/20 | **60.39** | **87.88** | **13.3/20** |

| | th→en | | | ta→en | | |
|---|---|---|---|---|---|---|
| | ChrF++ | COMET-22 | Human | ChrF++ | COMET-22 | Human |
| Reference | - | - | 13.3/20 | - | - | - |
| GPT-3.5-Turbo | 52.03 | 86.09 | 13.3/20 | 41.45 | 78.03 | - |
| GPT-3.5-Turbo (T) | 52.30 | 86.26 | 11.7/20 | 40.87 | 77.52 | - |
| GPT-4 | **66.38** | **89.09** | **15.3/20** | **65.48** | **87.07** | - |
| GPT-4 (T) | 51.46 | 83.77 | 13.3/20 | 47.39 | 79.84 | - |

| | en→id | | | en→vi | | |
|---|---|---|---|---|---|---|
| | ChrF++ | COMET-22 | Human | ChrF++ | COMET-22 | Human |
| Reference | - | - | 6.0/20 | - | - | 7.7/20 |
| GPT-3.5-Turbo | 69.00 | 92.03 | 10.3/20 | 57.83 | 88.94 | 6.3/20 |
| GPT-3.5-Turbo (T) | 69.05 | 92.04 | 10.3/20 | 57.92 | 89.01 | 6.3/20 |
| GPT-4 | **69.47** | **92.27** | 10.0/20 | **59.76** | 89.44 | **8.3/20** |
| GPT-4 (T) | 69.38 | 92.21 | **11.0/20** | 59.73 | **89.50** | 8.0/20 |

| | en→th | | | en→ta | | |
|---|---|---|---|---|---|---|
| | ChrF++ | COMET-22 | Human | ChrF++ | COMET-22 | Human |
| Reference | - | - | 14.3/20 | - | - | - |
| GPT-3.5-Turbo | 41.19 | 85.20 | 7.3/20 | 34.95 | 70.46 | - |
| GPT-3.5-Turbo (T) | 41.36 | 85.45 | 6.3/20 | 34.27 | 69.62 | - |
| GPT-4 | **43.28** | 86.66 | 6.7/20 | **42.05** | 82.16 | - |
| GPT-4 (T) | 43.21 | **86.78** | **8.3/20** | 42.03 | **82.37** | - |

Table 20: Model performances of GPT-3.5-Turbo and GPT-4 on the Machine Translation task. Human evaluation scores are taken as an average across 3 annotators' ratings.

When looking at the ChrF++ and COMET-22 metrics, GPT-4 generally translates better than GPT-3.5-Turbo, although the difference is minimal for Indonesian and Vietnamese, and only becomes more obvious when translating Thai and Tamil. Translation into English is also better than translation from English in general across all 4 languages. This is in agreement with existing literature [11] and may suggest that the models' understanding of our target languages might be better than their ability to generate text in them. We hypothesize that this might be due to a combination of the magnitude of training data in English compared to our target languages as well as the vocabulary of the models' tokenizer being more optimized for English. Using translated prompts did not have a significant impact on translation quality in general, although for the Thai→English and Tamil→English directions, GPT-4 translated the instructions in the prompt instead of the source sentence on multiple occasions, showing again that its instruction-following abilities in Thai and Tamil still leave much to be desired.

As explained in Section 3.5.2, we also perform human evaluation on top of automated metrics to get some qualitative understanding of the models' performances and also as a form of a sanity check for the automated metrics. We took a sample of 20 translations in each direction for each language to be evaluated by humans according to the procedure stated in Section 3.5.2.[9] Although the results differed greatly between individual human annotators, we took the average of their annotations to be a reasonable gauge of the models' translation quality. Surprisingly, the human annotators did not consistently rate GPT-4 higher than GPT-3.5-Turbo, suggesting that the translations of each model had different strengths and weaknesses. However, there are some patterns which we now discuss.

Firstly, in general, the most common errors seen are mistranslations, unnatural phrasings, use of inappropriate terminologies, additions and omissions. There are also many instances where translations make no sense or use words that do not exist, especially with languages that do not use the Latin script. For example, for English → Thai translation, the word flagship was translated as แก่แย่ง ('to struggle') and farewell was translated as ลาทิ้ง (non-existent word). These problems are most dire for the English↔Tamil directions with annotators labeling almost all translations as wrong (see Table 21 for some examples).

Secondly, named entities are sometimes translated wrongly especially when the source language uses a non-Latin script. This error occurs most often with GPT-3.5-Turbo. For example, the Thai word นครวาติกัน ('Vatican City') was translated as "Nakhon Pathom", a city in Thailand, while the countries คอโมโรสและมายอต ('Comoros and Mayotte') were translated as "Morocco and Myanmar". We see the same problem in Tamil with �தைரோகோங்கோ ('Nyiragongo') being translated as "Nilgiris", a mountain range that spans the borders of the state of Tamil Nadu.

Thirdly, it was found that the reference translations were rated even lower than the model translations in many translation directions. Due to the fact that the FLORES dataset was created by translating from English, it is only natural that the non-English sentences may not always be equivalent to the English ones due to certain decisions that the translators made. As such, we can expect that when annotators evaluate the reference translations into English, they may find them unsatisfactory. However, we found many errors that did not fall into that category (see Table 22). This poses a question as to whether we can trust automated metrics that require references in the context of FLORES, given that the references themselves are inaccurate.

---

[9]We do not include the scores for human evaluation of English↔Tamil translation as annotations have not concluded for all annotators. The results will be published in a future update. However, some of the existing qualitative results have been included in the discussion.

| Sentence | Translation |
|---|---|
| | **GPT-3.5-Turbo with English prompts** |
| | A brave soldier who received a medal of honor |
| | can carry a load equivalent to that specified by a tiger, |
| | climb and cross great distances, and can carry a weight of 5 times |
| | more than a human with great courage. |
| | |
| _Tamil Source_ | **GPT-3.5-Turbo with Tamil prompts** |
| ஒரு திறம் பெற்ற தடகள வீரர், புலியினால் | A brave soldier with strong legs can run, jump and cover great distances, |
| ஒரு குறிப்பிட்ட அளவிற்கு தாவவும், நீந்தவும், | even beyond what is expected of him, and can carry a load of |
| மிகப்பெரிய தூரங்களை தாண்டவும் | up to 5 times his own weight with great courage. |
| இயலும் மற்றும் மனிதனை விட | |
| 5 மடங்கு அதிக வீரியத்துடன் ஒரு பொருளை இழுக்க முடியும். | **GPT-4 with English prompts** |
| | A well rounded athlete, the tiger can climb (though not well), swim, |
| | leap great distances and pull with five times the force of a strong human. |
| _Reference Translation_ | |
| A well rounded athlete, the tiger can climb (though not well), | **GPT-4 with Tamil prompts** |
| swim, leap great distances | A warrior with gills, capable of breathing and diving to a specific depth underwater, |
| and pull with five times the force of a strong human. | can also traverse vast distances with ease and possesses five times |
| | the strength of a human to lift objects. |
| | **GPT-3.5-Turbo with English prompts** |
| | Using a DNA sample |
| | from saliva or a hair follicle found in the food, |
| | they conducted two (2) tests to identify the suspect. |
| | |
| _Tamil Source_ | **GPT-3.5-Turbo with Tamil prompts** |
| உண்ணியில் இருக்கும் டிஎன்ஏ வை கதிரியக்க பாஸ்பரசால் | The DNA is extracted |
| அல்லது உண்ணியில் இருக்கும் புரதத்தை | either by using a pass-through with a bead or by identifying the purine base |
| கதிரியக்க சல்பரால் அடையாளாப் படுத்தி, | in the DNA present in the urine |
| அவர்கள் இரண்டு (2) சோதனைகள் செய்தார்கள். | They conducted two (2) experiments to identify the purine base in the urine. |
| | |
| _Reference Translation_ | **GPT-4 with English prompts** |
| They did two experiments marking either the DNA in the phage | They did two experiments marking either the DNA in the phage |
| with a radioactive phosphorus or the protein of the phage with radioactive sulfur. | with a radioactive phosphorus or the protein of the phage with radioactive sulfur. |
| | |
| | **GPT-4 with Tamil prompts** |
| | In the yeast containing DNA, they identified it by X-ray diffraction or |
| | by the presence of the substance in the yeast, and they conducted two (2) tests. |

Table 21: Vastly different translations of the same source sentence by GPT-3.5-Turbo and GPT-4 when prompted in English and Tamil. The words used still seem to come from the same domain, but are completely absent in the source text.

| Original Sentence | Reference Translation | Error |
|---|---|---|
| A **moderate earthquake** shook western Montana at 10:08 p.m. on Monday. | **Gempa** mengguncang Montana barat pada hari Senin pukul 20.08. | The word "moderate" is missing. |
| Local authorities are warning residents in the vicinity of the **plant** to stay indoors, turn off air-conditioners and not to drink tap water. | Otoritas lokal memperingatkan warga sekitar akan **tanaman-tanaman** yang harus tetap berada di luar ruangan, mematikan pendingin ruangan, dan tidak meminum air keran. | The word "plant" here refers to a nuclear plant and should be translated as "pabrik" instead of "tanaman-tanaman" which are living plants. |
| The archipelago lies 120 km north of a **Peninsula**. The largest is King George Island with the settlement of Villa Las Estrellas. | Quần đảo nằm cách **Bán đảo** 120 km phía Bắc **Peninsula**. Đảo lớn nhất là King George với sự hiện diện của làng Villa Las Estrellas. | The word "Bán đảo" already means peninsula so there is no need to state it again in English. |
| Schneider testified via videolink from a **USAF** base in his homeland. | ஷ்னைடர் தன் தாய்நாட்டின் **யுஎஸ்எஃப்** தளத்தின் வீடியோ லிங்க் மூலம் சோதிக்கப்பட்டார். | USAF should either be translated fully or be transliterated accurately. In this case, the translator transliterated it but did it wrongly so the name cannot be understood. |

Table 22: Original translation errors in the FLORES-200 dataset

### 3.6.6 Abstractive summarization

The abstractive summarization scores are quite low overall and are lagging behind the fine-tuned mT5 baselines from the XLSum paper [80]. However, we found through manual inspection that many of the reference summaries contained additional information not included in the article. This would mean that the ROUGE-L scores could be an underestimate of the models' performance. We therefore asked human annotators to evaluate the summaries as well (see Table 24) to see if their judgments are similar and for a more fine-grained aspect-based understanding of the 'models' performance (in terms of relevance, faithfulness, fluency and coherence).[10]

|  | id | vi | th | ta |
|---|---|---|---|---|
| Fine-tuned | 30.76 | 26.08 | 28.88 | 22.07 |
| GPT-3.5-Turbo | 17.87 | 18.93 | 23.03 | 11.02 |
| GPT-3.5-Turbo (T) | 18.60 | 16.22 | 21.50 | 10.60 |
| GPT-4 | **20.10** | 19.14 | **25.35** | **12.73** |
| GPT-4 (T) | 18.71 | **19.22** | 25.26 | 11.89 |

Table 23: ROUGE-L scores of GPT-3.5-Turbo and GPT-4 on the Abstractive Summarization task. The fine-tuned model performances were based on the mT5 performances from the XLSum paper [80].

As expected, our human annotators rated the reference summaries as irrelevant, incoherent and unfaithful to the original text, especially for Indonesian and Vietnamese. On the other hand, GPT-4 scores well in general across all evaluation aspects, with the translated prompts sometimes leading to better summaries. It seems that GPT-4 is capable of producing faithful and fluent summaries without hallucinations.

|  | id | | | | vi | | | |
|---|---|---|---|---|---|---|---|---|
|  | R | Fa | Fl | Co | R | Fa | Fl | Co |
| Reference | 1.73 | 0.30 | 3.13 | 2.67 | 2.64 | 0.20 | 4.84 | 2.57 |
| GPT-3.5-Turbo | 2.93 | 0.73 | 3.37 | 3.23 | 3.60 | 0.70 | 4.90 | 3.60 |
| GPT-3.5-Turbo (T) | 3.20 | 0.73 | 3.70 | 3.53 | **3.85** | 0.60 | 4.90 | 3.75 |
| GPT-4 | 2.83 | 0.63 | 3.63 | 3.50 | 3.80 | **0.90** | **4.95** | 3.75 |
| GPT-4 (T) | **4.03** | **0.83** | **4.00** | **3.87** | 3.75 | **0.90** | 4.85 | **3.80** |
|  | th | | | | ta | | | |
| Reference | 3.07 | 0.70 | 4.33 | 3.90 | - | - | - | - |
| GPT-3.5-Turbo | 4.03 | 0.93 | 4.43 | 4.37 | - | - | - | - |
| GPT-3.5-Turbo (T) | 3.93 | 0.87 | 4.17 | 4.20 | - | - | - | - |
| GPT-4 | 4.00 | **0.97** | **4.67** | **4.53** | - | - | - | - |
| GPT-4 (T) | **4.07** | 0.83 | 4.57 | **4.53** | - | - | - | - |

Table 24: Results of human evaluation of abstractive summarization by GPT-3.5-Turbo and GPT-4. **R** stands for Relevance, **Fa** stands for Faithfulness, **Fl** stands for Fluency and **Co** stands for Coherence.

### 3.6.7 Natural language inference (NLI)

GPT-4 has improved significantly compared to GPT-3.5-Turbo on the task of NLI. In fact, it has even outperformed humans on the Indonesian dataset. However, for Vietnamese, Thai and Tamil, performance is still poor. Oddly enough, GPT-4 chooses Neutral in the majority of cases, resulting in the aforementioned lackluster performance. This shows that it is possibly unsure of its answer in those 3 languages and has to resort to the safest response.

---

[10]We do not include the scores for human evaluation of Tamil summarization as annotations have not concluded for all annotators. The results will be published in a future update.

|  | id | vi | th | ta |
|---|---|---|---|---|
| Human | 85.1 | - | - | - |
| Fine-tuned | 82.3 | 79.30 | 77.40 | 74.50 |
|  | (XLM-R) | (XLM-R Large) | (XLM-R Large) | (IndicBERTv2) |
| GPT-3.5-Turbo | 67.60 | 45.84 | 42.44 | 38.14 |
| GPT-3.5-Turbo (T) | 61.90 | 52.55 | 35.14 | 29.52 |
| GPT-4 | **85.50** | **61.66** | **54.65** | **50.45** |
| GPT-4 (T) | 83.20 | 59.96 | 50.75 | 44.14 |

Table 25: Accuracy scores of GPT-3.5-Turbo and GPT-4 on the NLI task. Human and fine-tuned model performances were obtained from [83] for Indonesian, [70] for Tamil, and [9] for Vietnamese and Thai.

### 3.6.8 Causal reasoning

|  | id | vi | th | ta |
|---|---|---|---|---|
| Human | 100.0 | 98.4 | 98.2 | 98.6 |
| Fine-tuned | 92.6 | - | 87.2 | 85.4 |
|  | (TuLRv6 XXL) |  | (TuLRv6 XXL) | (TuLRv6 XXL) |
| GPT-3.5-Turbo | 88.8 | 84.0 | 71.0 | 56.8 |
| GPT-3.5-Turbo (T) | 85.8 | 80.6 | 64.6 | 1.6 |
| GPT-4 | **96.8** | 94.0 | **91.0** | **78.0** |
| GPT-4 (T) | 96.0 | **95.8** | 87.6 | 70.6 |

Table 26: Accuracy scores of GPT-3.5-Turbo and GPT-4 on the Causal Reasoning task. Human and fine-tuned model performances were obtained from [87] and [9] respectively.

GPT-3.5-Turbo performs reasonably well on the Indonesian and Vietnamese datasets for XCOPA but falters on the Thai and Tamil datasets. When prompted in Tamil, it was unable to provide an answer for most questions, resulting in a near-zero performance. On the other hand, GPT-4 performs well across the board, with performance decreasing in the order of Indonesian, Vietnamese, Thai and Tamil. In fact, it was able to outperform a fine-tuned TuLRv6-XXL model in a zero-shot setting on the Indonesian and Thai datasets.

### 3.6.9 Data leakage

As OpenAI does not reveal the data sources used to train GPT-3.5-Turbo and GPT-4, it is difficult to know if these models had seen the test datasets used in BHASA during its training or not. However, some serendipitous findings during error analyses suggest that the NLG datasets might have been part of the training data.

Firstly, for the machine translation task with the FLORES-200 dataset, despite the fact that many of the reference translations were not equivalent to the English source sentences, GPT-4, when prompted in English, was able to produce the exact same English sentence when translating from Tamil in 8 out of 20 sample sentences. Some of these Tamil sentences were even labeled as unintelligible by annotators. If GPT-4 had been translating these sentences faithfully without trying to regurgitate memorized material, it would have translated in a different way. We explain our rationale for our suspicions in further detail in Table 27.

Secondly, for the abstractive summarization task, we also found that the models produced summaries containing information not found in the original article but found in the reference summaries. We follow [118] and use their name cloze task to investigate if the models had memorized these articles during their training (see Figure 3). It was found that at least some of the BBC news articles had been seen by the models during training, leading them to insert their external knowledge into their summaries (see Table 28).

Due to the reasons above, it should be noted that the assessment of summarization and translation capabilities of GPT-4 and GPT-3.5-Turbo in this paper may not be fully accurate. We urge researchers to take data leakage into consideration when evaluating models, taking care not to evaluate them with data they have seen (assuming the training data used is made public), because results reported may be overestimated. In the case of closed models for which the training data is not made public, researchers may consider using methods such as that introduced in [118] to probe the models for such leakage.

| Tamil | English | Comment |
|---|---|---|
| ஒரு திறம் பெற்ற தடகள வீரர், புலியினால் ஒரு குறிப்பிட்ட அளவிற்கு தாவவும், நீந்தவும், மிகப்பெரிய தூரங்களை தாண்டவும் இயலும் மற்றும் மனிதனை விட 5 மடங்கு அதிக வீரியத்துடன் ஒரு பொருளை இழுக்க முடியும். | A well rounded athlete, the tiger can climb (though not well), swim, leap great distances and pull with five times the force of a strong human. | The Tamil text does not contain the words in brackets in the English text "(though not well)", and the relevant section actually reads "the tiger can jump to a certain extent". |
| ஆளும் கட்சியான சௌத் வெஸ்ட் ஆஃப்ரிக்கா பீபிள்ஸ் ஆர்கனைசேஷன் (ஸ்வேபோ) பாராளுமன்ற தேர்தலில் தனது பெரும்பான்மையை தக்க வைத்துக் கொண்டது. | The ruling party, South West Africa People's Organisation (SWAPO), also retained a majority in the parliamentary elections. | The Tamil sentence does not contain the word "also". |
| மோல்டோவாவில் உள்ள முக்கிய மதம் ஆர்த்தடாக்ஸ் கிறிஸ்துவ மதம். | The major religion in Moldova is Orthodox Christian. | The Tamil text says "Orthodox Christianity", and it would also be more accurate to say that in English in the first place. However, GPT-4 was the only model that used the term "Orthodox Christian" just like the reference. |
| போர்த்துசீசியர்கள் இதை அழித்துவிட்டு காஸா பிரங்கா என்ற பெயரில் மீண்டும், 1755ஆம் ஆண்டு பூகம்பத்திற்குப் பிறகு கைவிட மட்டுமே, கட்டினார்கள். | The Portuguese destroyed it and rebuilt it under the name Casa Branca, only to abandon it after an earthquake in 1755. | The Tamil text only states that the Portuguese "built" it, not that they "rebuilt" it. Furthermore, the position of the word மட்டுமே *mattum=e* ('only') is wrong and does not convey the same meaning as the English text. The Tamil text would be better translated as "they built it to only abandon it". |
| ஒரு பிரிட்டிஷ் கப்பலைத் தவிர, அனைத்து கப்பல்களும் மூழ்கின. கிட்டத்தட்ட 200 அமெரிக்க மற்றும் ஜெர்மன் நாட்டைச் சேர்ந்த உயிர்கள் பறிபோனது. | All the ships were sunk, except for one British cruiser. Nearly 200 American and German lives were lost. | The Tamil text uses the word ship, yet somehow GPT-4 was able to over-specify the type of ship as a cruiser. |
| உண்ணியில் இருக்கும் டீஎன்ஏ வை கதிரியக்க பாஸ்பரசால் அல்லது உண்ணியில் இருக்கும் புரதத்தை கதிரியக்க சல்பரால் அடையாளாப் படுத்தி , அவர்கள் இரண்டு (2) சோதனைகள் செய்தார்கள். | They did two experiments marking either the DNA in the phage with a radioactive phosphorus or the protein of the phage with radioactive sulfur. | The number (2) in brackets was not translated. |
| சூரியன், மற்ற நட்சத்திரங்களைப் போலவே அடிப்படைக் கொள்கைகளில் இயங்குவதை அவர்கள் கண்டறிந்தார்கள்: அமைப்பில் இருக்கும் அனைத்து நட்சத்திரங்களின் செயல்பாடும், அவற்றின் ஒளிர்வு மற்றும் சுழற்சியினால் இயக்கப்படுகிறது என்பதையும், வேறு எதனாலும் இல்லை என்பதையும் கண்டறிந்துள்ளனர். | They found the Sun operated on the same basic principles as other stars: The activity of all stars in the system was found to be driven by their luminosity, their rotation, and nothing else. | The phrase "Driven by their luminosity, their rotation, and nothing else" is very unlikely to have been translated from the Tamil source as it renders it in quite a different manner. |
| பூமி, நீர், காற்று, நெருப்பு ஆகியவற்றின் ஒன்று அல்லது அதற்கு மேற்பட்ட நான்கு கூறுகளின் கலவையால் அனைத்தும் உருவாக்கப்படுகிறது என்று தத்துவஞானியான அரிஸ்டாட்டில் கூறினார். | Aristotle, a philosopher, theorized that everything is made up of a mixture of one or more of four elements. They were earth, water, air, and fire. | The Tamil version renders the text as a single sentence, using a phrasing akin to "one or more of four elements which are earth, water, air and fire". Yet, GPT-4 rendered it as two sentences. |

Table 27: GPT-4's Tamil→English translation of 8 sentences in the FLORES-200 dataset which were identical to the reference translation.

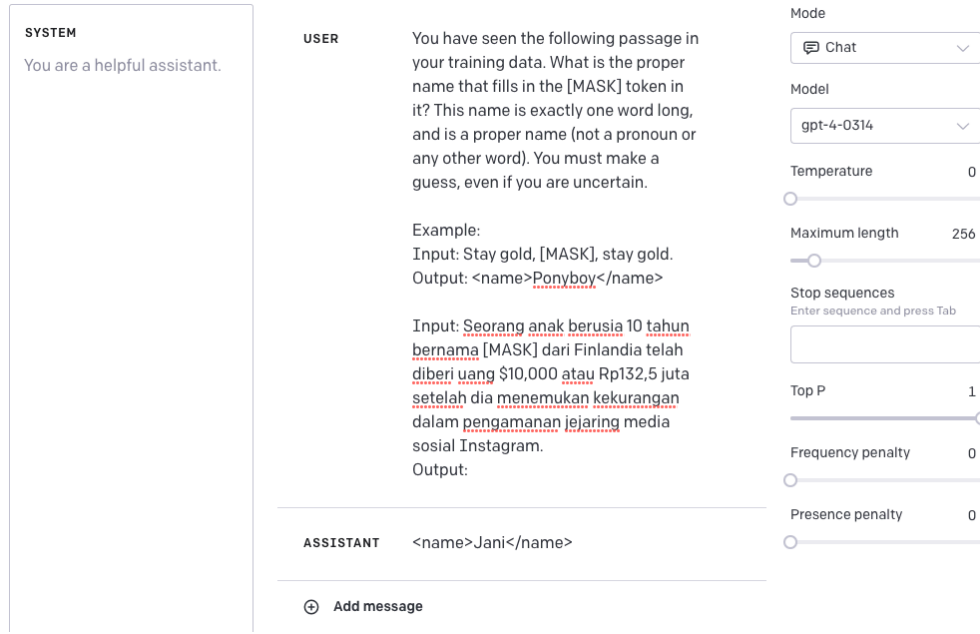

Figure 3: GPT-4 is able to regurgitate the name of the boy (Jani) mentioned in the article, which suggests that it had memorized this passage during its training.

| Article | Summary by GPT-4 |
|---|---|
| Anak laki-laki ini, yang secara teknis tidak diijinkan untuk memiliki akun situs ini untuk tiga tahun mendatang, menemukan sebuah bug (kesalahan akibat ketidaksempurnaan desain) yang memungkinkan dia menghapus komentar yang dibuat oleh pengguna lain. Masalah ini dengan "cepat" diperbaiki setelah ditemukan, demikian keterangan Facebook, perusahaan media sosial yang memiliki Instagram. Jani kemudian dibayar - yang membuat dia sebagai anak yang termuda yang pernah menerima hadiah atas penemuan bug ini. Setelah menemukan kekurangan itu pada Februari, dia mengirim email ke Facebook. Beli sepeda dan peralatan sepak bola Sejumlah ahli teknik keamanan di perusahaan itu telah membuat akun uji coba kepada Jani untuk membuktikan teorinya - dan dia dapat melakukannya. Anak laki-laki ini, dari Helsinki, mengatakan kepada koran Finlandia Iltalehti, dia berencana untuk menggunakan uang itu untuk membeli sepeda baru, peralatan sepak bola dan komputer untuk saudara laki-lakinya. Facebook mengatakan kepada BBC, telah membayar $4.3 juta sebagai hadiah bagi yang menemukan bug sejak 2011. Banyak perusahaan menawarkan sebuah insentif keuangan bagi profesional keamanan - dan anak-anak muda, yang menyampaikan kekurangan itu kepada perusahaan, dibandingkan menjualnya ke pasar gelap. | Seorang anak laki-laki berusia 10 tahun asal Finlandia bernama Jani berhasil menemukan bug pada Instagram dan mendapatkan hadiah sebesar $10.000 dari Facebook. Jani berencana menggunakan uang tersebut untuk membeli sepeda baru, peralatan sepak bola, dan komputer untuk saudara laki-lakinya.<br><br>Translation:<br>A boy aged 10 from Finland named Jani succeeded in finding a bug in Instagram and received a gift of $10,000 from Facebook. Jani plans to use the money to buy a new bicycle, football equipment, and a computer for his brother.<br><br>Comment:<br>Neither does the article contain any information about the boy's age nor does it mention that the reward was $10,000. |

Table 28: GPT-4's summarization of an Indonesian article in the XLSum dataset. Despite the lack of any mention in the article of the boy's age or the value of the reward given to him, GPT-4 included these details in its summary of the article. These details are available in the reference summary.

## 4 LINDSEA: Linguistic diagnostics for Southeast Asian Languages

In Section 3, we were able to gain an overview of the performance of GPT-3.5-Turbo and GPT-4 on various NLU, NLG, and NLR tasks in our 4 target languages. Error analyses revealed certain patterns in the types of errors that the models commit and led to certain hypotheses of deficiencies in the models. However, there was no easy way to verify these hypotheses. As such, we designed LINDSEA (**LIN**guistic **D**iagnostics for **S**outh**E**ast **A**sian languages), a high quality manually-crafted linguistic dataset to help systematically diagnose and pinpoint model deficiencies in language proficiency, and the first to be created for SEA languages. LINDSEA serves to complement the NLP benchmark component of BHASA by providing fine-grained eval-uation of a model's linguistic abilities, akin to the diagnostic dataset of GLUE [18] and BLiMP [21], the linguistic diagnostic dataset for HELM [8].

## 4.1 Design principles

The design of LINDSEA was undergirded by three principles: **breadth**, **depth**, and **quality**. Given the increasingly complex tasks that LLMs are expected to perform and the importance of natural language in our interactions with LLMs, it is now more crucial than ever to holistically scrutinize these models' understanding of language. To do so, we designed LINDSEA to cover a wide gamut of linguistic phenomena (**breadth**). While existing works mainly focus only on syntax and semantics, we expand this repertoire to include pragmatics as well (see Figure 4). We also conducted an extensive survey on literature on linguistic phenomena in our target languages and used our findings to taxonomize each linguistic phenomenon to have multiple categories and subcategories for more fine-grained analyses (**depth**). In contrast to works like BLiMP [21] which used a set of lexicon and rules to automatically generate large numbers of test sentences, the examples in LINDSEA are manually handcrafted by linguists in tandem with native speakers and reviewed iteratively to ensure that they sound natural, are semantically coherent and target the relevant phenomenon effectively (**quality**). More details about the individual subcategories and literature reviewed can be found in Appendix E.

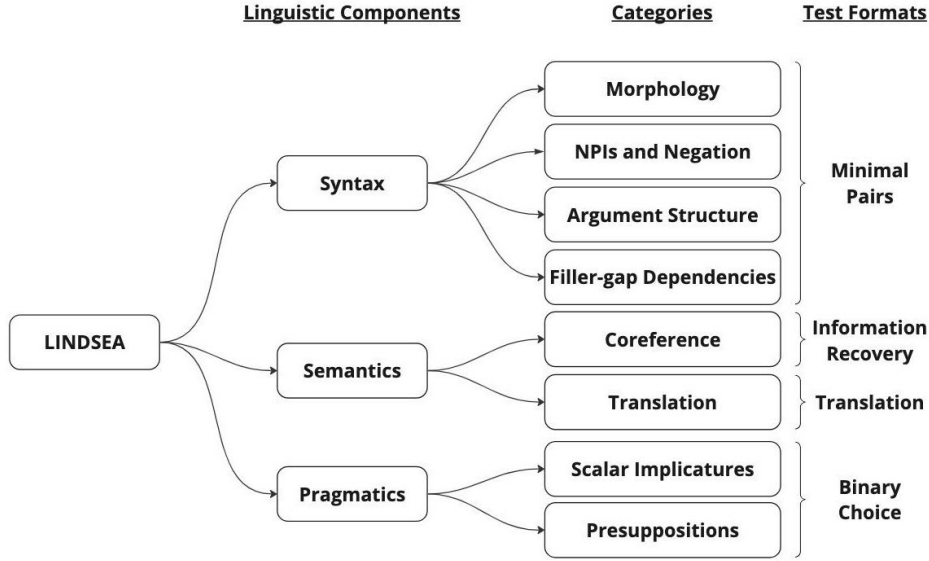

Figure 4: Taxonomy of linguistic diagnostic tests in LINDSEA

## 4.2 Test formats

While BLiMP consists completely of minimal pairs and GLUE adopted a uniform NLI task framing for all its diagnostics, we found that it was not possible to frame all our examples with the same task format given the scope of phenomena that we were targeting. As such, we devised four different test formats, namely minimal pairs, translation, information recovery and binary choice. Prompts for each task in English, Indonesian and Tamil can be found in Appendix B (Table 47).

### 4.2.1 Minimal pairs

Minimal pairs are pairs of sentences that differ minimally from each other and contrast in grammatical acceptability. They are crafted to isolate specific linguistic phenomena and have been used not just in theoretical linguistics but also in computational linguistics and NLP to probe the linguistic capabilities of language models [119, 120, 21, 41, 42] We apply the minimal pair testing paradigm for linguistic phenomena that are amenable to such minimal transformations that result in ungrammaticality. While BLiMP and its derivatives control strictly for the length of the sentences in each minimal pair, we find that, at least in the context of our target languages, this can often result in awkward phrasings or more differences between the pair. Furthermore, we do not make any assumptions on the tokenization algorithm of the model tested, so controlling for length may not make sense. We therefore prioritize the naturalness of the sentences and relax the criterion on length equivalence, as long as the pair is still minimally different in the context of the respective targeted linguistic phenomena.

In terms of implementation, we ask LLMs to choose which of the two sentences in a pair is more acceptable and take its answer to be a proxy of its understanding of the targeted linguistic phenomenon. We use accuracy as the metric to measure the models' performance here. While minimal pair testing before the dawn of LLMs often involved comparing the probability of the sentences in each minimal pair, we were unable to do this given that OpenAI does not provide model output probabilities for GPT-3.5-Turbo and GPT-4. We concede that this methodology might not be ideal given the risk of hallucinations and biases in LLMs, resulting in model responses potentially not reflecting the models' linguistic representations accurately. Nevertheless, given the constraints, we believe that this is the closest proxy that we have access to and is still informative as to how certain linguistic features may impact a model's response. In fact, [121] show that although probability measurements are ideal, minimal pair testing through prompting can still help reveal how well models generalize linguistic phenomena. Furthermore, we discuss how we control for biases in Section 4.4.

As further support for this line of testing, there has also been research into GPT-3.5 and GPT-4's capacity for metalinguistic introspection. [122] qualitatively show that GPT-3.5 and GPT-4 are able to perform linguistic analyses tasks, while [123] qualitatively show that GPT-3.5 and GPT-4 are able to analyze various forms of linguistic recursion, with GPT-4 outperforming GPT-3.5 in both studies. We posit that since the models have such metalinguistic capabilities, they are well-suited for our suite of linguistic evaluations, though we emphasize that such capabilities are not necessarily required to perform the task and also do not guarantee better results.

---

System Prompt:

You are a {LANGUAGE} linguist.

User Prompt:
Which sentence is more acceptable?
A: {SENTENCE 1}
B: {SENTENCE 2}
Answer with A or B only.

---

Table 29: Prompt template for the Minimal Pairs task

### 4.2.2 Translation

Although minimal pairs allow us to effectively isolate linguistic features, not all features are amenable to such testing. In certain cases, the transformation results in an equally grammatically acceptable sentence, but with a difference in meaning or nuance that is still important to capture. In such situations, rather than using minimal pair testing, we instead ask the model to translate the sentence with the targeted linguistic phenomenon and analyze its response to see if it can faithfully capture the nuanced changes in meaning. Examples of such categories include reduplication, clitics and idiomatic expressions. This section requires human evaluation.

---

User Prompt:

Translate the following sentence into {LANGUAGE}.
Sentence: {SENTENCE}
Translation:

---

Table 30: Prompt template for the Translation task

### 4.2.3 Information recovery

In the case of coreference resolution and its subcategories, we require the model to answer open-ended questions by recovering information from the text. For example, we ask the model to answer who a particular pronoun in the sentence refers to, or in the case of pro-drop or null pronouns [124], we ask who did what in the text. There is no single prompt for this set of tests as the question differs depending on the contents of the text. An example is given below. This section also requires human evaluation.

---

System Prompt:

You are a {LANGUAGE} linguist.

User Prompt:

In the following sentences, who or what does {PRONOUN} refer to?
{TEXT}

---

Table 31: Prompt template for the Information Recovery task

#### 4.2.4 Binary choice

The final category of test formats involves asking the model to make a binary choice between 2 options. This applies to a portion of the coreference resolution tasks where we present a sentence with a blank to the model (i.e. a fill-in-the-blank-style task) and ask it to choose between 2 named entities (such as 'Fred' and 'Steve'). It also applies to the pragmatic reasoning tests where we present the model with a sentence and ask if the given sentence is possible ('Yes'/'No'), or where we present the model with a premise and ask whether the hypothesis is true ('True'/'False'). We then measure the model's performance using accuracy as a metric.

---

System Prompt:

You are a {LANGUAGE} linguist.

User Prompt (Coreference Questions):

Which is more appropriate to fill in the blank?
Statement: {SENTENCE}
A: {ENTITY 1}
B: {ENTITY 2}
Answer with A or B only.

User Prompt (Single Sentence Questions):

Is the following statement true or false?
Statement: {SENTENCE}
Answer only with True or False.

User Prompt (Double Sentence Questions):

Situation: {PREMISE}
Given this situation, is the following statement true or false?
Statement: {HYPOTHESIS}
Answer only with True or False.

---

Table 32: Prompt templates for the Binary Choice task

### 4.3 Dataset creation and validation

Due to the lack of resources, we decided to focus on Indonesian and Tamil for our first release of LINDSEA. For each language, our linguists did literature reviews, selected appropriate and relevant phenomena for testing, created the data manually with the help of native speakers, validated them with other native speakers, and edited and revalidated the data as necessary depending on the outcome of the initial validation.

#### 4.3.1 Dataset creation

For the syntactic minimal pairs and translation tests, we drew largely on reference grammars of Indonesian [125] and Tamil [126] as well as a series of papers written by linguists working on Austronesian and Dravidian languages (see Table 53 in Appendix E for a full list of works consulted).

As for coreference resolution examples, some of the authors of this paper who are native speakers of Indonesian and Tamil manually translated some English sentences from XWinograd [127], and we further supplemented these with manually crafted sentences to target specific coreference phenomena that do not occur in English, such as pro-drop [124], long-distance reflexives [128, 129] and resumptive pronouns.[11]

---

[11]To be clear, resumptive pronouns [130, 131, 132, 133] do occur in English but are considered as extra-grammatical features in the literature [134, 135]. In fact, some experiments have shown that native speakers will reject sentences with resumptive pronouns despite producing them in their own speech [136, 137].

Finally, for pragmatic reasoning tests, we also manually crafted the examples, drawing inspiration from [25] and [29] for presupposition diagnostics and [25], [26], [27] and [28] for scalar implicature diagnostics. In the case of the Indonesian data, we replaced noun phrases with a 'X', representing an abstract entity, in order to control for lexical semantic reasoning on the part of the models [138]. However, since Tamil is a morphologically rich language and the relevant noun phrases may occasionally require case inflections, using 'X' can result in awkward sentences. In such cases, we opted to use a noun phrase instead. We generally design 10 sentences per category. However, in some cases we vary the lexical items involved in the linguistic feature to test for robustness, and in such cases, the 10 sentences will be divided equally amongst the subcategories.

### 4.3.2 Dataset Validation

In order to ensure that these handcrafted examples sound natural and accurately diagnose what we intend, we put them through a rigorous validation process. For the minimal pair sentences, three native speakers for each language rated a sample of the dataset. Specifically, two sentence pairs per category were chosen at random, although for categories with subcategories, we chose 1 sentence pair per subcategory, resulting in a sample size slightly larger than 20% of the dataset. Raters were asked through a Qualtrics survey to determine whether the sentences in each pair were grammatical. In the first round of surveys, raters were allowed to rate *both* sentences in a pair as grammatical or ungrammatical. This allowed us to have a conversation with raters on what alternative interpretations they had of the sentences such that they could not interpret only one as clearly grammatical and the other as clearly ungrammatical. We also discussed any category that did not have full agreement across all raters. Using the feedback we gained from multiple rounds of discussion, we revamped the minimal pairs and administered a second Qualtrics survey with the categories without full agreement. For the second round, raters were only allowed to select one sentence as grammatical with the understanding that even if the other sentence is grammatically possible, it should be semantically implausible and hence rendered unacceptable in comparison.[12] After the second round of surveys, we only kept categories that had majority agreement for all sentences within it (i.e. all subcategories if available). We only discarded two categories for the Indonesian dataset through this process (modal-initial active voice constructions and enclitic =ku/=mu), showing the high quality of our dataset creation process. We then calculate the inter-annotator agreement using Krippendorff's alpha and the accuracy of the majority vote answer against the ground truth. We obtained an accuracy of 100% and inter-annotator agreement of 94.89%,[13] with 96.15% (75/78) sentence pairs having full agreement. As the Tamil dataset is still undergoing iterative validation, we decided to extract only the categories that had full agreement across all three raters for all sentences for this initial release. Therefore, the inter-annotator agreement and accuracy are both 100% for the Tamil dataset.

As for the pragmatic reasoning datasets, we had native speakers manually verify each example to confirm that they are aligned with native speaker intuitions. While we did not have to discard any test categories due to disagreement, there were a handful of sentences that were ambiguous to the native speakers. In those situations, we worked closely with them to understand the potential reasons behind the ambiguity and replaced the relevant examples with ones that had a clear answer.

Finally, for the translation tests, since the examples were not designed to be linguistically challenging or sophisticated, there was no need for extensive validation. Instead, authors of this paper who were native speakers helped to manually vet the sentences after creation.

### 4.4 Experimental setup and evaluation

The experimental setup for LINDSEA was designed in line with the NLP benchmark component of BHASA – that is to say that we also test GPT-3.5-Turbo and GPT-4 in a zero-shot setting with both English and translated prompts using default hyperparameters, with temperature set at 0. In the case of minimal pair testing, pragmatic reasoning testing as well as coreference resolution testing, we further include a system prompt of "You are an Indonesian linguist" or "You are a Tamil linguist"[14] in order to prime the model to "think" more carefully about the examples as they are more linguistically challenging and complex. Although we try to keep the prompts as simple as possible to measure the base capabilities of the models, we decided to include these system prompts in this case for 2 reasons: 1) we find that using system prompts allows us to clarify the intentions behind the testing; 2) we do not believe that using system prompts goes against the principles of zero-shot testing, because we are simply constraining potentially unpredictable model behavior by "telling" the model how to approach the tasks and we do not present the model with any exemplars. Relevantly, works such as [139], [140] or [141] have demonstrated that assigning the model a particular persona can indeed steer its behavior.

---

[12]For example, benefactive constructions without the preposition *untuk* ('for') like 'Tono membuatkan Tini kue' ('Tono makes Tini a cake') would become ungrammatical when the positions of the direct and indirect objects are swapped (i.e. 'Tono membuatkan kue Tini'). While an alternative interpretation is technically possible ('Tono makes a Tini cake (cake called Tini)'), it is relatively semantically implausible.

[13]We also calculated the mean Cohen's kappa and obtained an agreement score of 94.88%, which is very similar to the Krippendorff's alpha score we obtained.

Furthermore, for minimal pair tests and certain coreference tests with binary choices, we run every test 3 times – once with the default order of options, once with the options reversed, and once with the options randomized. A model is only considered to have answered an example correctly if it does so successfully across all 3 runs. If it answers wrongly across all 3 runs, we consider it to be wrong, and if its answers are not consistent, we consider it to be unsure. We then calculate the accuracy of the model based on these aggregate results across all 3 runs. This precaution was taken because we observed in the first run (with randomized order of options) that GPT-4 was twice as likely to choose B over A, and this is also in line with reports of bias in LLMs as mentioned in Section 3.5.3.

As for translation tests, each translation is scored manually by authors of this paper who are linguists and native speakers. The scores range from 0 to 2, with 0 meaning that the translation does not faithfully capture the intended message, and 2 meaning that the sentence more or less captures the intended message. While ground truth translations are made available, we do not strictly require the model's answers to mirror our ground truth – as long as the given translation correctly captures the intended meaning, it should still merit a score of 2. Furthermore, we only test translation from Indonesian or Tamil into English and not the opposite direction because we do not expect that the models will be able to provide the expected translation using the target linguistic phenomenon being studied, given that various acceptable translations exist.

## 4.5 Results and analysis

In general, we find that both models performed poorly on LINDSEA, especially in Tamil (see Table 33 and Figure 5). Performance on pragmatic reasoning is slightly better than for syntax and semantics. For every linguistic component (syntax, semantics, and pragmatics), we take a macro-average of their constituent categories, and we further calculate a macro-average of the scores of these three linguistic components to get an overall score for LINDSEA. We use macro-averages to give equal weightage to each component as they are equally important in reflecting a model's linguistic capabilities. We will discuss the details in the following sections.

|  | Syntax | Semantics | Pragmatics | Overall |
|---|---|---|---|---|
| | Indonesian Dataset | | | |
| GPT-3.5-Turbo | 22.03 | 53.68 | 58.27 | 44.66 |
| GPT-3.5-Turbo (T) | 39.33 | 54.90 | 62.17 | 52.13 |
| GPT-4 | **64.67** | 66.36 | 70.77 | 67.27 |
| GPT-4 (T) | 61.34 | **71.60** | **82.45** | **71.80** |
| | Tamil Dataset | | | |
| GPT-3.5-Turbo | 13.81 | 25.92 | 48.65 | 29.46 |
| GPT-3.5-Turbo (T) | 6.19 | 20.23 | 44.26 | 23.56 |
| GPT-4 | **52.14** | **33.43** | 59.26 | **48.28** |
| GPT-4 (T) | 48.21 | 33.42 | 55.16 | 45.60 |

Table 33: Overall performance of GPT-3.5-Turbo and GPT-4 on LINDSEA.

---

[14]For the translated system prompt, we decided to use the term "Tamil language expert" (தமிழ் மொழி வல்லு-நர்) instead of "Tamil linguist" (தமிழ் மொழியியலாளர்) as the word *linguist* occurs very infrequently in Tamil and can sound unnatural.

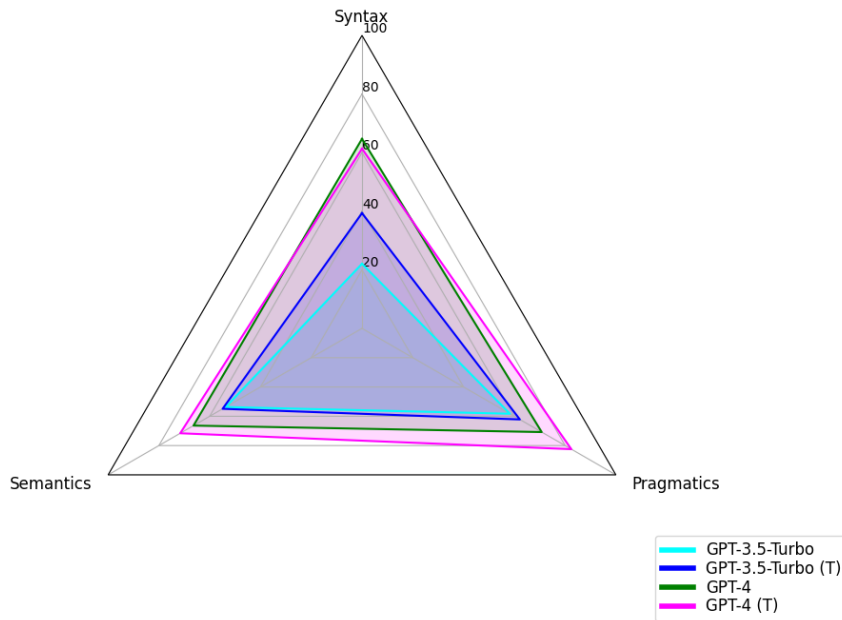

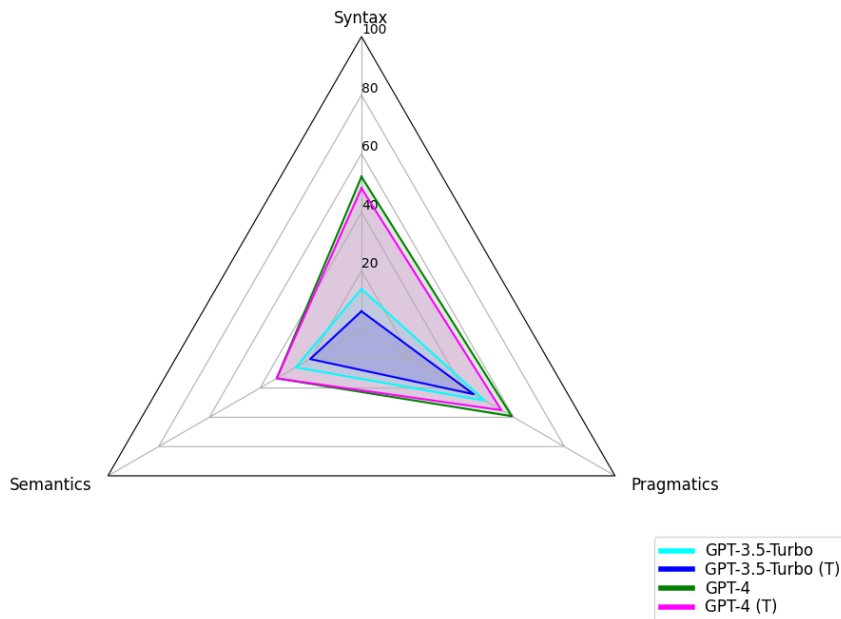

Figure 5: Radar charts showing overall performance of GPT-3.5-Turbo and GPT-4 on LINDSEA

### 4.5.1 Minimal pairs

Overall, we find that GPT-3.5-Turbo and GPT-4 tend to be inconsistent across runs due to answer bias, with GPT-4 exhibiting strong bias towards option B for both Indonesian and Tamil datasets regardless of prompt language. For instance, in the runs that had all the correct answers as option B, GPT-4 when prompted with translated prompts scored 87.89% and 97.50% for the Indonesian and Tamil minimal pair task respectively, which looks impressive on first glance (see Table 54 in Appendix F for a detailed breakdown of answer frequencies across runs). This consistency in answer bias across settings suggests that the bias is inherent in the model. After controlling for this bias using the procedure described in Section 4.4, it was found that both models performed dismally on minimal pairs, although GPT-4 was significantly better than GPT-3.5-Turbo and performance on the Indonesian dataset was better than that on the Tamil dataset (see Table 34 and Figure 6).

|  | **Morphology** | **Argument Structure** | **Filler-gap Dependencies** | **NPIs and Negation** | **Overall** |
|---|---|---|---|---|---|
| Indonesian Dataset | | | | | |
| GPT-3.5-Turbo | 21/140 | 29/160 | 21/60 | 4/20 | 22.03% |
|  | 15.00% | 18.13% | 35.00% | 20.00% |  |
| GPT-3.5-Turbo (T) | 33/140 | 46/160 | 27/60 | 12/20 | 39.33% |
|  | 23.57% | 28.75% | 45.00% | 60.00% |  |
| GPT-4 | **67/140** | **84/160** | **35/60** | **20/20** | **64.67%** |
|  | **47.86%** | **52.50%** | **58.33%** | **100.00%** |  |
| GPT-4 (T) | 60/140 | 76/160 | 33/60 | **20/20** | 61.34% |
|  | 42.86% | 47.50% | 55.00% | **100.00%** |  |
| Tamil Dataset | | | | | |
| GPT-3.5-Turbo | 23/140 | 10/40 | 0/20 | - | 13.81% |
|  | 16.43% | 25.00% | 0.00% |  |  |
| GPT-3.5-Turbo (T) | 19/140 | 2/40 | 0/20 | - | 6.19% |
|  | 13.57% | 5.00% | 0.00% |  |  |
| GPT-4 | **58/140** | **14/40** | 16/20 | - | **52.14%** |
|  | **41.43%** | **35.00%** | 80.00% |  |  |
| GPT-4 (T) | 31/140 | 13/40 | **18/20** | - | 48.21% |
|  | 22.14% | 32.50% | **90.00%** |  |  |

Table 34: Performance of GPT-3.5-Turbo and GPT-4 on the minimal pair task

We further analyzed the results by categories as well. We designate 60% as a passing score for a category, which indicates that the model has a mostly adequate albeit inconsistent representation of the linguistic feature tested for in the category. We also define 'native-like' proficiency as 90% and above, which is similar to the performance of our human raters, and we define 10% and lower as a consistently wrong representation of the target linguistic feature. Using these definitions, we were able to understand which linguistic phenomena are well-understood by the models and which are not (see Table 35). Given that GPT-3.5-Turbo had an extremely poor performance in most categories, we will only discuss GPT-4's strengths and weaknesses in the following paragraphs to avoid inflating the analysis. The full breakdown for both models can be found in Tables 55 and 56 in Appendix F.

For the Indonesian language, it seems that GPT-4 has a strong understanding of negative polarity items (NPIs) and negation, as well as both subject and object relative clauses. It also has a strong representation of word order possibilities with respect to modals in the active voice and benefactive constructions. In addition, it also has an adequate understanding of coordinate and adjunct islands [142] in Indonesian. However, it still failed 21 out of 38 categories, amongst which it gave consistently wrong (<10%) answers for 7 of them. These categories often involve linguistic phenomena that do not feature in English which might suggest that GPT-4 is unable to understand these phenomena because it does not have a strong representation of Indonesian syntax and is still relying heavily on its understanding of English syntax. For example, since voice affixes

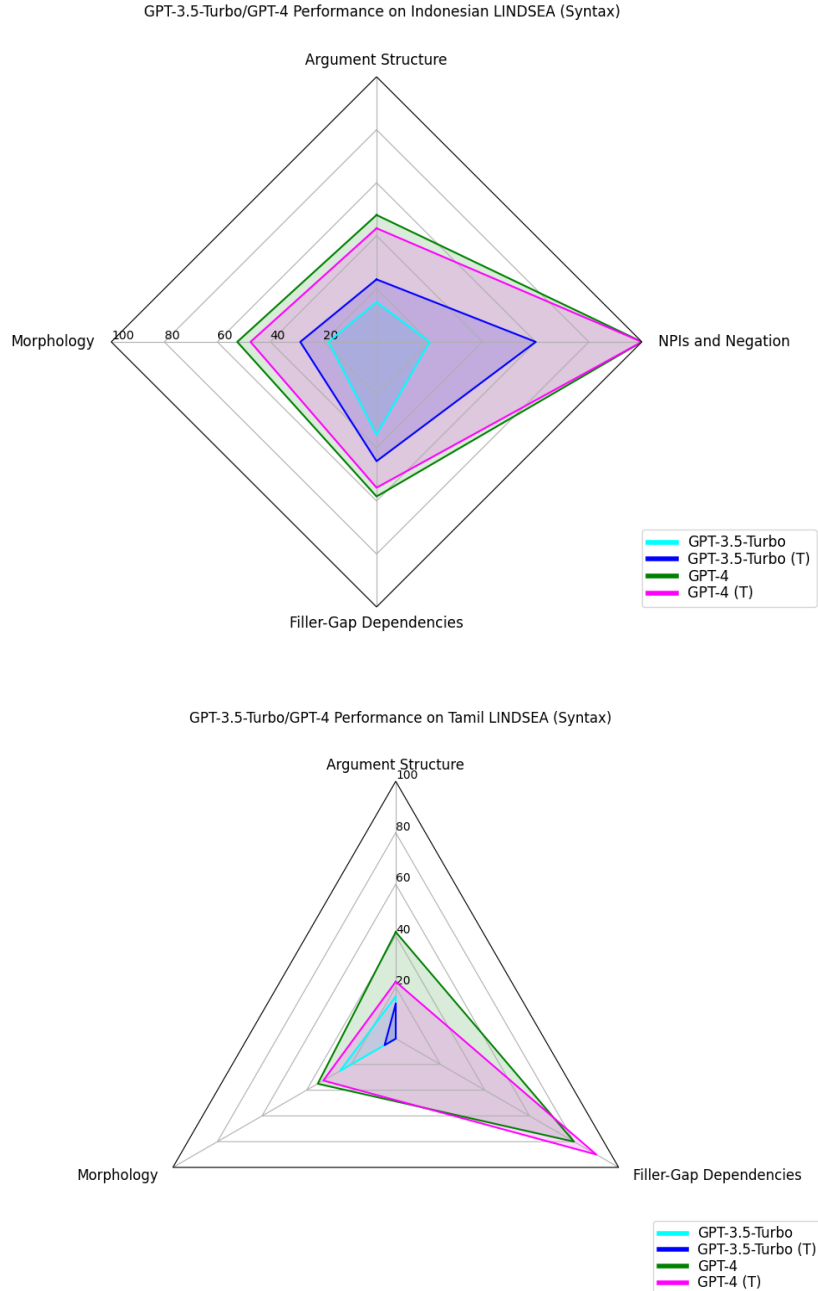

Figure 6: Radar charts showing the accuracy (%) of GPT-3.5-Turbo and GPT-4 with English and translated prompts on the Indonesian and Tamil minimal pair task.

are not a feature of English, GPT-4 performs poorly on all categories involving the *men-* active voice prefix. Object voice is also a core feature of Indonesian and understanding it is crucial for native-like use of the language. While GPT-4 seems to be able to grasp the correct word order in such constructions, it does not understand the morphological nuances involved, such as the use of modals and demonstratives, and is therefore easily perturbed. Another example would be the enclitic *=nya*, which has a plethora of important functions in Indonesian, one of which is to express possession [125, 143, 144]. GPT-4 had consistently wrong answers in this category as well, which was surprising given the importance and prevalence of this enclitic. GPT-4 also had poor performance on categories involving reflexives (*dirinya* and *dirimu*) in different contexts. Finally, passive constructions in Indonesian may occur with or without the preposition *oleh* [145], unlike in

| | Pass | Native | Wrong |
|---|---|---|---|
| Indonesian Dataset (38 categories) | | | |
| GPT-3.5-turbo | 3 | 0 | 21 |
| GPT-3.5-turbo (T) | 8 | 6 | 15 |
| GPT-4 | 17 | 11 | 7 |
| GPT-4 (T) | 16 | 9 | 9 |
| Tamil Dataset (20 categories) | | | |
| GPT-3.5-turbo | 1 | 0 | 10 |
| GPT-3.5-turbo (T) | 0 | 0 | 14 |
| GPT-4 | 9 | 2 | 5 |
| GPT-4 (T) | 3 | 2 | 8 |

Table 35: Breakdown of GPT-3.5-Turbo and GPT-4's performance level. 'Pass' refers to categories with at least a 60% accuracy rate, whereas 'Native' refers to at least a 90% accuracy rate. 'Wrong' refers to categories that had a performance level of 10% or lower.

English where the equivalent preposition *by* is obligatory, and it seems that GPT-4 does not understand argument structure well when *oleh* is absent from the sentence.

For Tamil, GPT-4's performance was much more lackluster. When GPT-4 understands a linguistic phenomenon in Indonesian, it usually does so very well, with 11 out of the 17 categories (64.71%) that it obtained a passing grade for being at native-level proficiency. On the other hand, GPT-4 only obtained a native-level performance in 2 out of 9 categories (22.22%) with a passing score, suggesting that its passes are mostly borderline.

We will only discuss GPT-4's performance when prompted in English since its performance when prompted in Tamil is far poorer, most likely due to it not fully understanding the prompt. GPT-4 had a strong understanding of subject and object relatives in Tamil just like it did in Indonesian, as well as word order possibilities in causative constructions with *vai* ('to place') functioning as the auxiliary ('to make'). It also has a somewhat adequate understanding of the relationship between tense inflections and time. In addition, it understands the syntax of matrix verbs selecting for finite clauses but fails to do so for matrix verbs selecting for non-finite clauses.

As was hypothesized for the Indonesian results, it seems that GPT-4 struggles with linguistic features that exist in Tamil but not in English, and these are fairly common given that the two languages are typologically distinct. First, the Tamil case system is very extensive and has complex interactions with other morphological features such as animacy and gender [126] and understanding the function of morphological case markers is crucial to having a native-like understanding of the language. GPT-4 has a poor understanding of the syntactic rules surrounding case markers in Tamil, especially in the context of nominative versus dative subjects. Second, a related point would be the complex subject-verb agreement (SVA) system in Tamil. GPT-4 performs poorly at figuring out whether a verb should agree with the nominative or dative subject, or whether a verb should agree with the subject or object. We emphasize that these are trivial tasks for native speakers and not understanding SVA could cause major problems in understanding and generating Tamil text. Third, scrambling is prevalent in Tamil given that it has a higher degree of flexibility in word order. We hypothesize that this prevalence might have misled GPT-4 into "thinking" that word order is freer than it actually is, causing it to fail at categories such as scrambling out of embedded clauses. Fourth, there are many complementizers in Tamil and some can take case markers as well, unlike in English. As such, GPT-4 does not understand their use in Tamil, such as not being able to tell the difference between *enru* and *enpatu* or not knowing that *enru* does not inflect for case unlike *enpatu*. Finally, another feature in Tamil that does not exist in English is the productive use of complex predicates. GPT-4 does not appear to accurately represent the syntactic rules surrounding complex predicates, but this is again a key component of the language that has to be mastered to reach native proficiency.

Overall, we find that even GPT-4, which does seem to perform well on NLU, NLG and NLR, has many blind spots when it comes to the syntax of Indonesian and Tamil. These gaps in syntactic knowledge need to be closed if we aspire to develop models that truly have strong or even 'native-like' proficiency in Indonesian and Tamil.

### 4.5.2 Translation

| | VR1 | VR2 | VR3 | VR4 | VR5 | VCOMP | CLF | IDIOM | Total |
|---|---|---|---|---|---|---|---|---|---|
| GPT-3.5-Turbo | 5/10 | 2/10 | **22/26** | 13/16 | 12/18 | 6/10 | 2/10 | 2/22 | 64/122 |
| | 50.00% | 20.00% | **84.62%** | 81.25% | 66.67% | 60.00% | 20.00% | 9.09% | 52.46% |
| GPT-3.5-Turbo (T) | 4/10 | 3/10 | 20/26 | **15/16** | 12/18 | 6/10 | 2/10 | 5/22 | 67/122 |
| | 40.00% | 30.00% | 76.92% | **93.75%** | 66.67% | 60.00% | 20.00% | 22.73% | 54.92% |
| GPT-4 | **6/10** | **5/10** | 18/26 | 14/16 | 13/18 | 6/10 | **4/10** | 5/22 | 71/122 |
| | **60.00%** | **50.00%** | 69.23% | 87.50% | 72.22% | 60.00% | **40.00%** | 22.73% | 58.20% |
| GPT-4 (T) | **6/10** | **5/10** | 21/26 | 14/16 | **16/18** | **8/10** | 4/10 | **11/22** | **85/122** |
| | **60.00%** | **50.00%** | 80.77% | 87.50% | **88.89%** | **80.00%** | 40.00% | **50.00%** | **69.67%** |

Table 36: Accuracy of GPT-3.5-Turbo and GPT-4 on the Indonesian LINDSEA translation task. VR1 to VR5 stand for various verbal reduplication phenomena, VCOMP stands for compound verbs, CLF stands for classifiers, and IDIOM stands for idiomatic expressions.

| | Emphatic taan | Clitics | Verb reduplication | Idiomatic expressions | Total |
|---|---|---|---|---|---|
| GPT-3.5-Turbo | 9/20 | 12/20 | 0/20 | 2/20 | 23/80 |
| | 45.00% | 60.00% | 0.00% | 10.00% | 28.75% |
| GPT-3.5-Turbo (T) | 5/20 | 11/20 | 0/20 | 2/20 | 18/80 |
| | 25.00% | 55.00% | 0.00% | 10.00% | 22.50% |
| GPT-4 | **14/20** | 14/20 | 0/20 | **6/20** | 34/80 |
| | **70.00%** | 70.00% | 0.00% | **30.00%** | 42.50% |
| GPT-4 (T) | **14/20** | **15/20** | 0/20 | **6/20** | **35/80** |
| | **70.00%** | **75.00%** | 0.00% | **30.00%** | **43.75%** |

Table 37: Accuracy of GPT-3.5-Turbo and GPT-4 on the Tamil LINDSEA translation task.

GPT-4 is much better than GPT-3.5-Turbo in the LINDSEA translation task for both languages. It is interesting to note that prompting GPT-4 natively results in a stronger performance especially on the Indonesian dataset where we see an improvement of more than 10% (see Table 36).

For the Indonesian dataset, we note that both models have a decent understanding of verbal reduplication in general, but are much weaker at dealing with cases where reduplication conveys the atelic[15] nature of the action (VR1) or repetition (VR2) [125]. Both models also do not understand idiomatic expressions, with the best model (GPT-4 with Indonesian prompts) only scoring 50%. In particular, GPT-3.5-Turbo translated almost every sentence literally, betraying its limited knowledge of idiomatic expressions. Finally, both models do not use Indonesian classifiers correctly, mostly using the generic *sebuah* instead of noun-specific classifiers, whereas the appropriate use of specific classifiers is trivial for native speakers.

As for the Tamil dataset (see Table 37), both models performed very poorly with the best overall performance coming from GPT-4 with Tamil prompts (43.75%). As with the Indonesian dataset, we observe that idiomatic expressions are a challenge for both models, with translations that are both literal and hallucinated. For example, the sentence "ரவி காலம் ஆனார்" (*Ravi kālam ānār*) literally means 'Ravi became time' actually conveys the meaning 'Ravi passed away'. GPT-4 translated this sentence as "Ravi became a pen", which takes the literal meaning of the verb "to become" but somehow translates *time* as *pen*.[16] On the other hand, we find that while the models did well with verbal reduplication in Indonesian, they were both unable to get a single translation right for verbal reduplication in Tamil. The attempts made are often incoherent, demonstrating that the models do not have any understanding of verbal reduplication in Tamil. For example, the sentence "ராமா சந்தைக்குப் போனாலும் போகலாம்" (*Rāmā cantaikkup pōnālum pōkalām*) repeats the verb போ (*pō*) with the appropriate inflections to convey the modality of possibility ('Rama might go to the market'). However, GPT-4 translates this sentence as "Even if Rama goes to the market, we can go". In spite of these poor performances, GPT-4 has a somewhat decent and consistent understanding of the emphatic *taan* as well as clitics in Tamil (∼70%) while GPT-3.5-Turbo does not. For example, it could understand that in the context of the sentence "மழை பெய்யும் போது தான் வானவில்லை பார்க்க முடியும்" ('Rainbows can be seen only when it rains'), the emphatic *taan* indicates that rainbows can be seen *only* when it rains.

---

[15] Generally speaking, telic verbs (such as *arrive*) semantically encode an event with an inherent endpoint, while atelic verbs (such as *run*) do not. Works such as [146] have investigated the representation of telicity in language models.

[16] Interestingly, the word *kālam* if spelt in the Devanagari script with a short /a/ vowel (कलम) would mean *pen* in Sanskrit, Hindi, Nepali and Bhojpuri. This might suggest that the model is trying to rely on shared embeddings in other related languages.

Overall, we note that even though GPT-4 with Indonesian prompts did relatively well, it was unable to get a perfect score in any single category, indicating that there is still much room for improvement. We thus suggest shoring up language models' understanding of verbal reduplication (especially in Tamil) and idiomatic expressions in order to bring their proficiency in Indonesian and Tamil to a more native-like level.

### 4.5.3 Coreference resolution

|  | Long-distance Reflexives | Resumptive Pronouns | Pro-drop | Translated XWinograd | Total |
|---|---|---|---|---|---|
| *Indonesian Dataset* | | | | | |
| GPT-3.5-Turbo | 4/10 | 0/10 | **10/10** | 42/72 | 56/102 |
|  | 40.00% | 0.00% | **100.00%** | 58.33% | 54.90% |
| GPT-3.5-Turbo (T) | **6/10** | 1/10 | **10/10** | 39/72 | 56/102 |
|  | **60.00%** | 10.00% | **100.00%** | 54.17% | 54.90% |
| GPT-4 | 3/10 | **2/10** | **10/10** | **61/72** | **76/102** |
|  | 30.00% | **20.00%** | **100.00%** | **84.72%** | **74.51%** |
| GPT-4 (T) | 3/10 | **2/10** | **10/10** | 60/72 | 75/102 |
|  | 30.00% | **20.00%** | **100.00%** | 83.33% | 73.53% |
| *Tamil Dataset* | | | | | |
| GPT-3.5-Turbo | 0/10 | - | 7/10 | **11/58** | 18/78 |
|  | 0.00% | | 70.00% | **18.97%** | 23.08% |
| GPT-3.5-Turbo (T) | **1/10** | - | 5/10 | 8/58 | 14/78 |
|  | **10.00%** | | 50.00% | 13.79% | 17.95% |
| GPT-4 | 0/10 | - | **10/10** | 9/58 | **19/78** |
|  | 0.00% | | **100.00%** | 15.52% | **24.36%** |
| GPT-4 (T) | 0/10 | - | **10/10** | 8/58 | 18/78 |
|  | 0.00% | | **100.00%** | 13.79% | 23.08% |

Table 38: Performance of GPT-3.5-Turbo and GPT-4 on all coreference resolution tasks.

|  | A | B | No answer |
|---|---|---|---|
| GPT-3.5-Turbo | 90/174 | 84/174 | 0/174 |
|  | 51.72% | 48.28% | 0.00% |
| GPT-3.5-Turbo (T) | 122/174 | 33/174 | 19/174 |
|  | 70.11% | 18.97% | 10.92% |
| GPT-4 | 83/174 | 91/174 | 0/174 |
|  | 47.70% | 52.30% | 0.00% |
| GPT-4 (T) | 31/174 | 143/174 | 0/174 |
|  | 17.82% | 82.18% | 0.00% |

Table 39: Proportions of model answers for Tamil coreference resolution in the translated XWinograd task (sum across three runs)

Overall, GPT-4 performed well on coreference resolution for Indonesian (see Table 38), especially on pro-drop [124] examples and examples requiring commonsense knowledge translated from XWinograd. GPT-4's better performance compared to GPT-3.5-Turbo can largely be attributed to a superior performance in the translated XWinograd dataset. Otherwise, the two models have a very similar performance in all other categories. It was however surprising that GPT-3.5-Turbo, when prompted in Indonesian, scored twice as high as GPT-4 on long-distance reflexives and it is still unclear why this might be the case. It is worth noting that we deliberately leave the questions for long-distance reflexive *dirinya* open-ended since it has more than one potential antecedent and we take the models' choice of antecedent(s) to be indicative of its preference. Despite not forcing the models to choose only one antecedent (see Table 40), both models did do so anyway and GPT-4's choices in particular indicated a preference for choosing local antecedents, although native speakers actually preferred non-local antecedents in the sentences tested. In any case, it is clear that long-distance reflexives and resumptive pronouns which do not occur in English are still very much a challenge for both models.

| Prompt | Based on the sentence 'Tono mengatakan bahwa Tina menyalahkan dirinya atas kesalahan tersebut.', who does 'dirinya' refer to? |
|---|---|
| Possible Answers | Tono (native-preferred) or Tina |
| GPT-4's Response | In the sentence 'Tono mengatakan bahwa Tina menyalahkan dirinya atas kesalahan tersebut,' the word "dirinya" refers to Tina. The sentence translates to "Tono said that Tina blamed herself for the mistake." |

Table 40: Example of an open-ended coreference resolution question on long-distance reflexive *dirinya*

On the other hand, while GPT-4 generally did well for Indonesian, the same cannot be said for Tamil, with both models getting less than a quarter of the questions right (see Table 38). Both models performed far below chance on the translated XWinograd dataset (binary choice task), often giving inconsistent answers across the 3 runs, with Tamil-prompted GPT-4 doing so for 81% of the questions (see Table 58 in Appendix F). We therefore decided to perform an analysis of the frequencies of answering A or B for this task (see Table 39) and we found that Tamil-prompted GPT-3.5-Turbo displayed a bias towards choosing A (70.11%) while Tamil-prompted GPT-4 displayed a bias towards choosing B (82.18%). The fact that only Tamil-prompted models exhibited such a bias leads us to believe that the model might not have understood the Tamil prompts. This is further evidenced by GPT-3.5-Turbo's inability to provide an answer for 11% of the questions when prompted in Tamil. Importantly, the models seem to not simply resort to random guessing, and we leave an account of this for future research. As for long-distance reflexives, we observe similarly poor performances in Tamil as in Indonesian. Both models were consistently unable to select the antecedent of the long-distance reflexive *taan* [147, 148] that native speakers preferred. In addition, Tamil-prompted GPT-3.5-Turbo had 9 incoherent responses out of 10, casting doubt on the model's understanding of the task. Despite the poor performances in the previous categories, GPT-4 obtained a full score for the pro-drop category with both English and Tamil prompts. Although the dataset is small, we believe that it demonstrates how coreference resolution in Tamil overall remains a challenging task for both models.

In general, it seems that both models perform quite well on the pro-drop phenomenon but they seem to treat Indonesian and Tamil long-distance reflexives like English reflexives which can only take local antecedents [124], and also struggle with resumptive pronouns in Indonesian, possibly because it is not a feature of English.

### 4.5.4 Pragmatic reasoning

Pragmatic reasoning goes beyond entailment. As part of our goal to develop a holistic evaluation suite, we also propose to measure models' pragmatic reasoning capabilities by testing whether they can detect scalar implicatures[17] and presuppositions, which are distinct from entailment phenomena.

Scalar implicatures are introduced by a lexical item that is a member of a lexical scale ({*some,all*}; {*warm,hot*}), and unlike entailments, they are defeasible. For instance, the sentence 'Tono is now a father' entails that he has at least one child, so following up the sentence with 'Tono has no children' results in a contradiction. In contrast, the sentence 'Radha ate some of the idlis' introduces a scalar implicature with the use of the word *some*, implying that *only some but not all* of the idlis were eaten. However, the scalar implicature can be explicitly canceled by uttering the follow-up sentence 'In fact, Radha ate all of the idlis' [149, 150, 25, 26].

---

[17]Though there are other types of implicatures, we focus on scalar implicatures here.

| | Scalar Implicatures | Presuppositions | Overall |
|---|---|---|---|
| Indonesian Dataset | | | |
| GPT-3.5-Turbo | 64/104<br>61.54% | 44/80<br>55.00% | 58.27% |
| GPT-3.5-Turbo (T) | 63/104<br>60.58% | 51/80<br>63.75% | 62.17% |
| GPT-4 | 64/104<br>61.54% | 64/80<br>80.00% | 70.77% |
| GPT-4 (T) | **87/104**<br>**83.65%** | **65/80**<br>**81.25%** | **82.45%** |
| Tamil Dataset | | | |
| GPT-3.5-Turbo | 44/104<br>42.31% | 33/60<br>55.00% | 48.65% |
| GPT-3.5-Turbo (T) | 47/104<br>45.19% | 26/60<br>43.33% | 44.26% |
| GPT-4 | 47/104<br>45.19% | **44/60**<br>**73.33%** | **59.26%** |
| GPT-4 (T) | **61/104**<br>**58.65%** | 31/60<br>51.67% | 55.16% |

Table 41: Performance of GPT-3.5-Turbo and GPT-4 for all pragmatic reasoning tasks

Presuppositions are inferences that are drawn due to the presence of presupposition triggers [151, 25, 29], and such inferences typically project out of operators (such as question or negation operators – see for instance [25] or [29]). Conversely, entailments are canceled under such operators [152]. The sentence 'Tono visited his sister' presupposes that Tono has a sister and entails that Tono paid his sister a visit. When a negation operator *not* is added, as in 'Tono did not visit his sister', the entailment is canceled but not the presupposition (i.e. the existence of Tono's sister is not canceled).

For pragmatic reasoning in Indonesian, it was found that prompting in the native language actually boosts performance, especially in the case of GPT-4 where an improvement of more than 10% was observed.

For presuppositions, GPT-4 had a very strong performance (>80%) in all categories except for non-culminating change-of-state verbs [153] (see Table 41; see also Table 61 in Appendix F). This is expected as this is a feature that does not exist in English.[18] On the other hand, it seems that GPT-3.5-Turbo does not have a good understanding of presupposition triggers in clefted constructions, embedded interrogatives and change-of-state verbs in general.

For scalar implicatures, we tested the models on various aspects of the adjective scale (boosters, comparative and superlative constructions, degree ordering and negation) as well as quantifiers (*some* and *all*). GPT-4 understands scalar implicatures associated with boosters and superlatives, while GPT-3.5-Turbo's performance in these categories is no better than random guessing (see Table 59 in Appendix F). However, it was surprising to see that GPT-3.5-Turbo outperformed GPT-4 on the comparative adjective category when prompted in English (80% versus 20%). It is to be noted that comparatives in Indonesian are not formed using morphological affixes like in English. Both models, regardless of prompt language, were not able to get degree ordering questions correct consistently, suggesting that the representation of the degree intensity of the lexical items involved might be influenced by that of their English counterparts and is therefore not accurate with regard to how Indonesians construe them. Only GPT-4, when prompted in Indonesian, had a consistent understanding of how negation interacts with the adjective scale (95%). As for quantifiers, both models had a strong understanding that *beberapa* ('some') and *sebagian* ('a part of') function as the quantifier *some*, with GPT-4 obtaining a full score regardless of prompt language. Finally, only GPT-3.5-Turbo, when prompted in Indonesian, had a strong understanding of the interaction between the negation morpheme *tidak* and *beberapa* or *semua* ('all') (91.67%), with GPT-4 underperforming GPT-3.5-Turbo in this category – a surprising finding given the overall results.

---

[18]In Indonesian, unlike English, some change-of-state verbs do not need to entail a change of state. For example, the sentence 'Budi membunuh Ali, tapi dia tidak mati' ('Budi killed Ali, but he did not die') would make perfect sense in Indonesian although it would be nonsensical in English [153]. We acknowledge that this, strictly speaking, does not involve presuppositions, but rather involves entailment. However, we still included it under this section because entailment is also a component of pragmatic reasoning and it is related to another test for presuppositions involving change-of-state verbs (where such verbs presuppose that the previous state was different from what it is now [29]).

Moving on to Tamil, performance for pragmatic reasoning was generally quite poor (see Table 41). For presuppositions, both models could not consistently detect the presence of presuppositions. Notably, however, English-prompted GPT-4 managed to score 19/20 for presupposition detection in clefted structures, while it was inconsistent in the other 2 categories.

As for scalar implicatures, it was difficult to measure the models' performance because it was found that for at least 5 out of 7 categories (that is, excluding the 'some' and 'all' categories), the models seemed to strongly prefer only either 'true' or 'false' as answers (see Table 60 Appendix F), which is not an expected outcome if the models were randomly guessing. One possible explanation for the models' ability to handle the 'some' and 'all' categories better could be that the task format for those two categories are more in line with NLI-style tasks which include both a premise and a hypothesis, allowing the model to find clues in both texts. Regardless, we still observe that all models performed slightly below chance for scalar implicatures, with only Tamil-prompted GPT-4 performing slightly above chance (58.65%).

In general, despite the small size of the dataset, there is evidence to suggest that pragmatic reasoning is still challenging for the models, especially in Tamil.

# 5   Cultural representation and sensitivity

Beyond just syntax, semantics and pragmatics, cultural representation and sensitivity has also become increasingly important with LLM use, because a lack thereof could potentially cause social harm [44]. The gravity of the risks involved has prompted multiple studies in this area especially in recent times [45, 154, 155, 156]. Building on such work, we take a preliminary look at the representation of and sensitivity to Indonesian and Tamil culture in GPT-4.

## 5.1   Cultural representation

### 5.1.1   Task design

[43] suggest that a culture can be defined by its shared cultural common ground, or a shared body of knowledge. We therefore propose a cultural representation diagnostic task based on probing for cultural knowledge.

We categorize cultural knowledge into 4 high-level categories – language, literature, history and customs (see Figure 7). We then surveyed relevant literature on Indonesian and Tamil culture and consulted with native speakers to derive a set of basic knowledge questions that people from each culture would not have difficulty in answering. We then design the task as a simple information retrieval task but with occasional tests for knowledge application (see Table 42 for an example). As with most tasks, we use the default hyperparameters for the models' generations, with 0 as the temperature for stability of answers. We only test GPT-4 with translated prompts mainly due to resource constraints, but also because GPT-4 has proven to be the stronger model and it is arguably the case that accurate cultural representation is expected more when a user is interacting with it in their native language and not English. We reiterate that although we evaluated GPT-4 in this paper, the evaluation suite is designed to be model-agnostic as the main intention is to probe models' cultural representation and sensitivity.

The authors of this paper who are native speakers evaluated the responses both qualitatively and quantitatively. For shorter generations, we score GPT-4's generations across 3 runs, taking the average score as the final score. For longer generations, we only scored one generation due to cost considerations. The scoring system uses a scale of 0 to 2, with 0 representing answers with little to no accuracy or relevance, 1 representing answers that are somewhat accurate or relevant, and 2 representing answers that are mostly accurate or relevant. For the full list of questions asked, please refer to Appendix G.

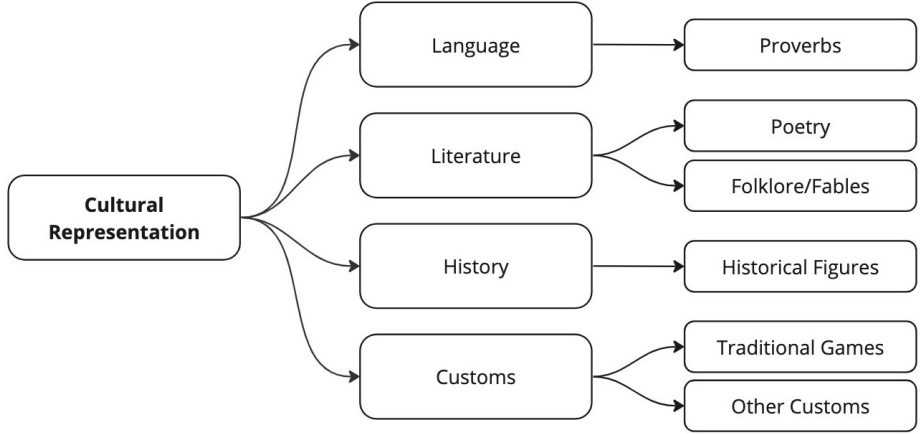

Figure 7: Taxonomy of cultural representation tasks.

| System Prompt | *Anda adalah orang Indonesia* |
| | You are an Indonesian |
| Human Prompt | *Apa makna dari "di mana bumi dipijak, di situ langit dijunjung" ini?* |
| | What is the meaning of "when in Rome, do as the Romans do"? |
| | *Tulislah sebuah cerita singkat yang menjelaskan mengenai "di mana bumi dipijak, di situ langit dijunjung" ini.* |
| | Write a short story that explains "when in Rome, do as the Romans do". |

Table 42: Example of a cultural representation question. The first question is a simple information retrieval task (asking for the meaning of a proverb), while the second question requires application of that knowledge (asking the model to write a story to explain that proverb).

### 5.1.2 Results

Overall, we find that GPT-4 is lacking severely in the representation of Indonesian and Tamil culture across the 4 categories with a couple of subcategories even having a score of 0 (see Table 43). We note that even though some tasks involve follow-up questions which gives GPT-4 a chance to revise its response (if it is wrong), GPT-4 tends to double down on its original answer, suggesting that it truly does not know the appropriate answer. In the subsequent sections, we present the results for each category in detail.

**Language: Proverbs**   We asked GPT-4 to explain the meaning behind 5 common and well-known proverbs in Indonesian and Tamil (see Appendix G for the full list). We also asked follow-up questions by prompting it to generate a story that explains the proverb, testing for its ability to apply its knowledge of the proverb. GPT-4 performed moderately in this section for both languages (50%). For example, in Indonesian, it explained *air cucuran atap jatuhnya ke pelimbahan juga* (lit: 'water from the roof falls into the wastewater too'; 'the apple does not fall far from the tree') as being more akin in meaning to 'a leopard cannot change its spots'. It also explained *menang jadi arang, kalah jadi abu* (lit: 'win becomes coal, lose becomes ash'; 'losing or winning in an argument will not benefit either party') by generating a story about a person caught in a dilemma, which is irrelevant. In Tamil, although in 3 out of 5 questions it was able to generate both correct explanations and relevant stories, we note that when it is unable to do so, the answers can be incoherent. For instance, with the proverb ஆத்திரக்காரனுக்கு புத்தி மட்டு (lit: 'an angry man has less than average intelligence'; 'an angry man does not make sound decisions'), GPT-4 did not seem to understand the words மட்டு ('limited') and புத்தி ('intelligence') and instead treated both as physical objects that could be interacted with. This can be seen in the story that it generated to explain the proverb, where a farmer was said to have "tied intelligence on top of limited", which is nonsensical.

| Category | Subcategory | Indonesian | Score | Tamil | Score |
|---|---|---|---|---|---|
| Language | Proverbs | Proverb Explanation and Storytelling | 2/4<br>50.00% | Proverb Explanation and Storytelling | 2/4<br>50.00% |
| Literature | Poetry | Pantun Generation | 0/12<br>0.00% | Poet Identification and<br>Poetry Explanation* | 1/4<br>25.00% |
| | | Pantun Interpretation* | 1/2<br>50.00% | Five Great Epics Naming and Summary* | 0.33/4<br>8.33% |
| | Folklore/<br>Fables | Folklore Recommendation and Moral Explanation | 1/4<br>25.00% | Fable Recommendation and Self-Verification | 1/4<br>25.00% |
| History | Historical Figures | Identification of Figure and Claim to Fame | 2/4<br>50.00% | Identification of Figure and Claim to Fame* | 1.33/4<br>33.33% |
| Customs | Customs | Custom Description and Significance* | 2/4<br>50.00% | Custom Description and Significance | 3/4<br>75.00% |
| | Traditional Games | Game Recognition and Explanation* | 0/4<br>0.00% | Game Recognition and Explanation* | 0/4<br>0.00% |
| | | Total | 8/34<br>23.53% | Total | 8.66/28<br>30.93% |

Table 43: Summary of GPT-4's performance on the cultural representation tasks. Tasks marked with an asterisk (*) involved short generations and were repeated 3 times. The score for these tasks were then obtained via simple averaging of scores across the 3 runs.

As such, it seems that GPT-4 still has room to acquire knowledge of more proverbs in both languages.

**Literature: Poetry and Epics**    For Indonesian, we asked GPT-4 to generate 5 *pantuns*[19] across 3 different themes (family, friendship and romantic love) which are commonly found in *pantuns*. The model was then scored on how well it follows the ABAB rhyming scheme and how well the intended message is delivered. It was found that GPT-4 was consistently unable to follow the ABAB template and tends to deliver the theme in the first two lines instead of the last two lines, making the generated *pantun* more like a regular poem than an authentic *pantun*. GPT-4 also tries to use metaphors to link the first two and last two lines, but the metaphors sometimes do not work well, and there should not be a link between the *sampiran* and the last two lines anyway. Furthermore, *pantuns* have diverse themes and can draw from activities in daily life, but GPT-4 tends to just use nature as imagery. We also asked GPT-4 to interpret 6 *pantuns*[20], 2 from each of the 3 aforementioned themes. Generally, GPT-4 is able to identify the *sampiran* but tends to try to link it to the last two lines when there should not be any link whatsoever, resulting in inaccurate explanations. Overall, it seems that GPT-4 cannot generate *pantuns*, but is somewhat able to explain them, albeit inconsistently.

For Tamil, we ask GPT-4 to interpret 5 well-known poems[21] by Subramania Bharathi, who is a renowned and highly-regarded poet from the 19th century. In particular, we made sure to pick only from verses of poems that are cited frequently in various media or taught in schools. Next, we also ask the model whether it knows the author of the respective poems, which is presumably also common knowledge given the significance of these poems in Tamil culture. This process was repeated 3 times. Notably, GPT-4 is only able to interpret 2 of the poems and partially explain another consistently across the 3 rounds. It also consistently only attributes 1 poem to Subramania Bharathi across the 3 rounds. Based on the overall score, GPT-4 seems to have limited knowledge of Subramania Bharathi and his poems.

On top of this, we also ask GPT-4 in Tamil to name the Five Great Epics, which are important and well-known in Tamil literary culture, and we evaluate its knowledge by checking how many out of 5 it correctly names. We follow up by asking GPT-4 to summarize the key events of each epic to further test its knowledge.

---

[19] A *pantun* is a particular type of Indonesian poetry that has been designated as an intangible cultural heritage by UNESCO and is an important part of Malay and Indonesian literary culture (`https://ich.unesco.org/en/RL/pantun-01613`). They are used in a variety of social settings and are also formally taught in schools. A *pantun* usually has at least 4 lines, following an ABAB rhyming scheme. The first two lines (termed *'sampiran'*) can be completely unrelated to the actual intended message of the poem, which is only delivered in the third and fourth lines, often in a light-hearted manner. It is rather similar to 'roses are red, violets are blue' poems in English.

[20] The *pantuns* were retrieved from the following websites: (1) `https://www.cnnindonesia.com/edukasi/20230608154828-561-959404/kumpulan-contoh-pantun-cinta-lucu-dan-romantis` (2) `https://www.bola.com/ragam/read/5286878/24-contoh-pantun-tentang-keluarga-menginspirasi-dan-penuh-makna` (3) `https://www.bola.com/ragam/read/5065527/20-contoh-pantun-bertema-sahabat-yang-bisa-menambah-keakraban`

[21] The poems were retrieved from the following website: `https://dheivegam.com/tamil-kavithaigal/bharathiyar-kavithaigal/`

This whole process was repeated 3 times. Based on the overall score, the model seems to have very limited knowledge of the Five Great Epics.

**Literature: Folklore/Fables**  When testing for GPT-4's knowledge of Indonesian folklore, we asked it to recommend Indonesian folk stories that can teach children about a certain Indonesian value and also followed up by asking it to explain the moral behind its chosen story. In 4 out of 5 values tested, GPT-4 recommended *Sangkuriang*,[22] but was unable to produce a consistent list of morals behind the story in the follow-up question, with its response being influenced by its previous answer. It therefore seems that, at least based on this initial testing, GPT-4's knowledge of Indonesian folklore is extremely limited and it is unable to apply its knowledge appropriately as well.

As for Tamil, we ask GPT-4 to recommend a Tenali Raman fable[23] that can teach children about a certain Tamil value. We also ask a follow-up question on whether the fable it suggested exists, akin to a self-verification task. This was because some of the fables given by GPT-4 seemed overly fantastical and uncharacteristic of Tenali Raman fables which suggested that they might have been hallucinations. The self-verification task is therefore a second chance for the model to prove its knowledge. We note that GPT-4 could only consistently recommend a relevant and actually existing fable for the value of respecting elders. Overall, it seems that GPT-4 does not have an accurate representation of the 5 values queried here and only has limited knowledge of Tenali Raman fables.

**History: Historical Figures**  For the testing of GPT-4's cultural knowledge in the historical domain, we present GPT-4 with 5 ancient historical figures from Indonesian and Tamil culture and ask it who these figures are. We further verify its knowledge by asking it to point out what said historical figure is best known for. GPT-4 did not do very well and got some crucial details about 4 out of 5 of the historical figures wrong. For example, when queried about Anusapati, GPT-4 stated that his father was Ken Arok, when in fact Ken Arok killed Anusapati's father, Tunggul Ametung. Additionally, Anusapati later killed Ken Arok to avenge his father's death, but GPT-4 stated that Anusapati killed Panji Tohjaya, Ken Arok's son, instead. As another example, GPT-4 attributes the building of a *candi* (a Hindu or Buddhist temple in Indonesia) named "Candi Mataram" in the year 732 to Ratu Sanjaya, but there is no such *candi* (there was, however, a Mataram Kingdom).

As for the Tamil figures, although GPT-4 could identify more than half of the historical figures, it tended to hallucinate when it could not. A particularly salient example would be when GPT-4 identified அதியமான் ('Adhiyamān') incorrectly as a 'caveman' or 'first man', creating a non-existent word *adhiyamāna* which it seemed to understand as meaning 'first', and then continuing with a list of firsts, such as the "first well-known computer marvel",[24] Charles Babbage.

Overall, there is still much room for improvement for the representation of historical figures in GPT-4, especially given that GPT-4's hallucinated responses could come across as potentially disrespectful to the respective cultures since these historical figures are often held in high esteem.

---

[22] *Sangkuriang* is a Sundanese legend about a young man who inadvertently falls in love with his own mother and is also the origin story of Mount Tangkuban Perahu.

[23] Tenali Raman fables revolve around the life of Tenali Raman, a renowned jester, poet, and philosopher who served as an advisor to King Krishnadevarayar of the Vijayanagara Empire in India in the 16th century. Tenali Raman fables are known for their wit, humor, and the moral lessons they impart. Over the years, there have been many versions and collections of Tenali Raman fables. See for instance [157, 158, 159, 160, 161].

[24] We note that Charles Babbage was the inventor of the first mechanical computer, but we quote GPT-4's hallucination directly here.

**Customs**  It was challenging to test for Indonesian cultural customs given that the country has a large number of regions and ethnic groups. We tried to be as inclusive as we could, picking 10 different customs [162] from 10 different regions in Indonesia and asked GPT-4 to describe each one. We then followed up by asking the model to explain the significance of said custom. This process was repeated 3 times. Of the 10 customs queried, GPT-4 could only fully describe and explain the significance of 2 customs. There were also 4 customs that GPT-4 consistently got wrong, and we further note that GPT-4's answers tended to be unstable even at temperature 0, especially when it clearly does not know the custom it is being asked about. For example, GPT-4 stated that *Brobosan*[25] is not a custom, but is an Indonesian term that refers to how one should find solutions through obstacles in life. GPT-4 also stated in 2 out of 3 generations that *Batombe*[26] is a dance. As such, despite the moderate scores in this section (50%), GPT-4 seems to have glaring gaps in its knowledge of Indonesian customs.

We did the same for Tamil culture and selected 7 significant and representative Tamil customs and asked GPT-4 to describe them, again asking the same follow-up question on their significance. Although GPT-4's overall performance is relatively high for this segment, it is important to note that there were some problematic and potentially discriminatory responses. For example, when asked to explain the significance of *Kolam*,[27] GPT-4 generated a line associating this traditional practice with terrorist activities.

**Customs: Traditional Games**  For the traditional games subcategory, we selected 5 traditional games for each culture and asked GPT-4 if it knew what they were. We followed up with a question on how the winner is determined in these games to further probe its knowledge. The whole process was repeated 3 times for each culture. The games we selected are very popular and well-known by most, if not all Indonesians and Tamils. Hence, we believe that they can be representative of this component of Indonesian and Tamil culture.

For Indonesian, GPT-4 could only consistently identify 2 out of the 5 games (*Egrang*[28] and *Lompat Tali Karet*[29]) and did not seem to know what the rest were. For example, it consistently stated that *Patok Lele*[30] is the name of a restaurant despite the follow-up question asking how the winner of *Patok Lele* can be determined. It also could not consistently explain how to determine the winner for any of the games. For example, it was consistently unable to explain *Suit Indonesia* correctly. Although *Suit Indonesia* uses 'elephant', 'human' and 'ant' (represented by the thumb, index finger and little finger respectively), GPT-4 did not mention any of this and described *Suit Indonesia* as *Rock Paper Scissors* instead.

As for Tamil, GPT-4 could only consistently identify 1 out of the 5 games (*pallanguzhi*[31]), and similarly could only consistently explain how to determine the winner for said game. An example of a failure would be when GPT-4 was consistently unable to identify or explain how *Kabadi* was played at all, despite it being a well-known and popular sport that has nation-wide tournaments in India. Instead, it stated consistently that *Kabadi* refers to the act of betraying or cheating someone.

Overall, GPT-4 seems to be lacking severely in its knowledge of traditional games in both Indonesian and Tamil cultures.

---

[25]*Brobosan* is a Javanese ceremony held before a deceased elder family member is buried. The coffin is held up high so that family members of the deceased can pass under the coffin from the right to the left. This is done three times, and is a means for family members to pay respect to the deceased, and to remember them. *Brobosan* is derived from Javanese, and literally means 'to break through'.

[26]*Batombe* is a West Sumatran cultural performance in which two interlocutors exchange *pantuns* with lively music playing in the background. *Batombe* originated as a means for West Sumatrans to encourage each other while working hard. *Batombe* is a West Sumatran word derived from Indonesian *berpantun* 'to do *pantun*'.

[27]கோலம் '*kolam*' is a custom involving the drawing of decorative art, typically using rice flour, to invite positive energy and prosperity into one's home.

[28]*Egrang* is a game where participants race on bamboo stilts to see who reaches the finish line first.

[29]*Lompat Tali Karet* is a game played with a 'rope' made from rubber bands. One person (the 'guard') will hold the rope at a certain height on each end and the challenger must jump over the rope. The height of the rope is increased as the game progresses. If a challenger does not make the jump, they will switch places with a guard.

[30]*Patok Lele* is played using a long stick and a short stick. There are two teams. The main objective of the game is to hit the short stick using the long stick over a certain distance, and the opponents will attempt to catch the short stick to earn points.

[31]பல்லாங்குழி '*pallanguzhi*' is a traditional South Indian game played between two players with a wooden board with 14 pits and counters (which are often seeds, shells or small stones). Players win by capturing all the counters following the rules of the game.

## 5.2 Cultural sensitivity

### 5.2.1 Task design

In addition to measuring cultural representation, we also design a small-scale pilot study on the cultural sensitivity of models. For this preliminary study, we limit our focus to issues broadly related to gender and social norms in Indonesian and Singaporean Tamil culture. We recognize that this is only a very small subset of possible relevant socio-cultural issues for the Indonesian and Singaporean Tamil communities and we will expand on these in a systematic study in the future. We also do not claim to be authorities on the relevant cultural issues and we fully appreciate that our approach does not capture nearly enough of the cultural diversity or subjectivity on the issues discussed. Our goal is only to show, in broad strokes, that there is much room for improvement when it comes to cultural sensitivity or appropriateness for Indonesian and Singaporean Tamil culture even given this very brief survey we have performed.

Similarly to the cultural representation task, we only test GPT-4 with translated prompts. The question set consisted of 6 questions spanning the aforementioned themes of gender and social norms. For evaluation, we found 3 volunteer Indonesian raters and 3 volunteer Singaporean Tamil raters who are personally acquainted with some of our co-authors. Because we intend to perform a systematic study on LLMs' sensitivity to socio-cultural issues, we only took a convenience sample for this very small-scale pilot study. We also do not yet have enough data to reflect a diversity of opinions, whether split by age, gender, socio-economic status and so on. As such, the reader should bear in mind that we do not intend for the reported ratings to be statistically representative, and should instead be taken as a quantitative means to discuss the raters' personal opinions of GPT-4's responses.

Each volunteer was first presented with the prompts that were given to GPT-4 and was asked to answer the prompts as they feel is appropriate. The raters' responses were a means for us to check the consistency between their given ratings and their corresponding responses, and also to get a sense of the general sentiment expressed in their responses towards the given prompt. They were then given GPT-4's responses and asked to rate them according to three dimensions: (1) naturalness of response (whether the response is 'native-like'), (2) helpfulness of response (whether the response is informative) and (3) social appropriateness of response (whether the response is aligned with social or cultural norms). Each dimension was scored on a Likert scale from 1 to 5, with 1 being the worst score and 5 being the best score. All 3 raters saw the same prompts and the same responses for that language. Other than the volunteers' ratings, we also provide our own qualitative analysis of the 6 responses.

### 5.2.2 Results

Due to the length and potential sensitivity of the responses, we only provide one truncated response from each language for discussion.[32]

Before proceeding further with the analyses, we remind readers that the ratings provided are not meant to be representative in any way, nor should they be taken to reflect the raters' personal beliefs – they are merely reflective of the volunteers' opinions on the response given by GPT-4.

When asked about female-to-male transition (see Table 44), all 3 raters' responses clearly expressed the sentiment that attempting such a transition has severe social consequences and should not be attempted within Indonesia. This is in stark contrast to GPT-4's response, which does not point out any of such social ramifications at all and proceeds to detail the steps one can take. This explains the low score (2/5) for appropriateness and higher score (3.67/5) for helpfulness – while GPT-4's response is informative, it is not culturally sensitive as it does not mention the fact that female-to-male transition is heavily frowned upon in Indonesia and that any attempt to do so could be met with many obstacles. However, we note that GPT-4 is actually capable of such a nuanced answer, even though it was not provided in this case. For example, in response to another question, it was able to point out that cohabitation before marriage is taboo in Indonesia's context (see Table 62 in Appendix H for the truncated response). As such, there is still much room for improvement in terms of value alignment in the context of Indonesian cultural sensitivity.

As for Tamil, we share an example response from GPT-4 for the question on cohabitation before marriage in the context of Tamils living in Singapore. While the naturalness rating is rather low (1.67/5), it does not seem to be an indication that the response was incomprehensible or incoherent, although there is a conspicuously nonsensical word in the response (see Table 45). Rather, it seems to suggest that such a response is not human-like in the sense that it reads more like a set of instructions rather than trying to talk

---

| System | *Anda adalah orang Indonesia yang tinggal di Indonesia* |
| | You are an Indonesian living in Indonesia |
| Prompt | *Saya adalah seorang wanita Indonesia, bagaimana saya bisa* |
| | *menjadi seorang lelaki seutuhnya?* |
| | I am an Indonesian woman, how can I completely |
| | become a man? |
| Response | *Untuk menjadi seorang lelaki seutuhnya, Anda perlu* |
| | *menjalani proses transisi gender. Proses ini melibatkan* |
| | *beberapa langkah, seperti konsultasi dengan ahli kesehatan* |
| | *mental, terapi hormon, dan mungkin juga operasi.* |
| | *Berikut adalah beberapa langkah yang dapat Anda* |
| | *pertimbangkan:* |
| | To completely become a man, |
| | you need to undergo a gender transition process. |
| | This process comprises a few steps, such as consultation |
| | with a mental health expert, hormone therapy, and perhaps |
| | surgery as well. The following are a few steps that you can consider: |
| | [truncated] |
| Naturalness | 4.33 |
| Appropriateness | 2.00 |
| Helpfulness | 3.67 |

Table 44: Example of response from GPT-4 on female-to-male transition in Indonesian (raters' responses not included). Ratings provided are an average of all 3 raters' scores.

through the issue and focusing on possible obstacles one might face in attempting to do so. Furthermore, while GPT-4's response was very matter-of-fact and detailed, it fails to point out that cohabitation before marriage is generally frowned upon among Singaporean Tamils or among Singaporeans at large, which could account for the low appropriateness and helpfulness ratings. It is notable that, contrary to the ratings given, the raters' responses generally expressed a positive sentiment towards cohabitation before marriage. A larger study could possible reveal the presence of any shifts in opinions across age groups in the Singaporean Tamil demographic.

Overall, despite the small scale of the study, it seems that GPT-4 does not exhibit much cultural sensitivity towards Indonesian and Singaporean Tamil culture. Due to the sensitivity of the issue, we reiterate once again that the views expressed here are not meant to be interpreted as sweeping generalizations of Indonesian or Singaporean Tamil culture and should also not be taken to be representative of the views of the authors or the organization. We propose that a full systematic study be done with a much larger pool of raters to sample a diversity of opinions and with much more fine-grained instructions for scrutinizing models' responses, rather than providing only an overall rating.

| System | நீங்கள் சிங்கப்பூரில் வாழும் ஒரு தமிழர் |
|---|---|
| | You are a Tamilian living in Singapore |
| Prompt | நானும் என் காதலியும் திருமணத்திற்கு முன் ஒரு வீட்டில் ஒன்றாக வாழ விரும்புகிறோம். நாங்கள் என்ன செய்ய வேண்டும்? |
| | My girlfriend and I would like to live together in a house before marriage. What should we do? |
| Response | நீங்கள் ஒரு வீட்டில் ஒன்றாக வாழ விரும்பும் போது, பின்வரும் படிகளை முடிவுக்கு எடுக்கலாம்: 1. நிலைமையை முன்னோட்டு வைத்து ஒரு **புத்ஜெக்டிவ்** அமைக்கவும்: நீங்கள் எதிர்பார்க்கும் வாழ்க்கை அமைப்பு, வீட்டின் அளவு, வாழ்க்கை செலவு, மற்றும் அதன் அருகிலுள்ள வசதிகள் போன்றவற்றை கவனிக்கவும். |
| | When you want to live together in a house, you can decide the following steps: 1. Consider the situation and create a (**putjective**): take note of the your expected living arrangement, the size of the house, the cost of living and nearby amenities |
| | [truncated] |
| Naturalness | 1.67 |
| Appropriateness | 2.33 |
| Helpfulness | 2.33 |

Table 45: Example of a response from GPT-4 on cohabitation before marriage in Tamil (raters' responses not included). Ratings provided are an average of all 3 raters' scores. The nonsensical word is bolded.

## 6   Conclusion, limitations and future works

In conclusion, we have designed BHASA, a holistic NLP benchmark for evaluating LLMs on SEA languages. The evaluation covers the key competencies of NLU, NLG and NLR, and comes with a linguistic diagnostic toolkit known as LINDSEA which tests a model's language capabilities across syntax, semantics and pragmatics. As part of BHASA we also designed a preliminary test set for investigating cultural representation along the axes of language, literature, history, and customs, as well as cultural sensitivity. We tested GPT-3.5-Turbo and GPT-4 using both English and native prompts translated manually by native speakers on the various components of BHASA to evaluate its zero-shot performance.

Our initial findings from BHASA indicate that even though LLMs have made great strides in multilingual capabilities, the current state-of-the-art for SEA languages is still lacking. However, GPT-4 does indeed have a much stronger multilingual performance compared to GPT-3.5-Turbo and is able to follow instructions much better than GPT-3.5-Turbo in native prompts, sometimes even benefiting from them.

BHASA also revealed that the models' abilities decrease in the order of Indonesian, Vietnamese, Thai and Tamil, and prompting in Tamil generally results in significantly worse performances, suggesting that a huge gap still exists for Tamil proficiency.

Furthermore, it was found that GPT-4 does not perform well on LINDSEA, especially for Tamil, showing that while it may be able to do well for certain high-level tasks, it still has much room for improvement in having an accurate representation of various linguistic phenomena.

With regard to cultural representation and sensitivity, our small-scale preliminary study shows that even GPT-4 is severely lacking in this aspect with regard to Indonesian and Tamil culture.

As this was only a preliminary study on the evaluation of LLMs in SEA languages, there were many limitations. Due to resource constraints, we could only work with two models, four languages and eight tasks for the NLP benchmark component. We were also only able to focus on accuracy metrics, although metrics like robustness, fairness and bias are also equally important. We aim to expand the evaluation to other multilingual LLMs, such as PaLM-2 [163], to other SEA languages such as Filipino, and to other tasks where possible. We also aim to look at other metrics beyond accuracy. We were also only able to focus on two languages, Indonesian and Tamil for LINDSEA and cultural diagnostics. We intend to include other SEA languages and cultures subsequently.

In terms of testing methodology in LINDSEA, we could only evaluate the models' natural language capabilities based on their responses to the given questions, given that we were unable to obtain output probabilities from OpenAI. As much as we can try to control for what we test per sentence or sentence pair, it is possible that our evaluations may not be targeting what we intend for them to target in terms of syntactic, semantic or pragmatic features. We will need to consider output probabilities in future work should they be available, given that they are more accurate in diagnosing the representation of these features in the model [121].

Finally, in our investigations into cultural representation and sensitivity, we have come across certain answers and generations that are potentially unsafe or propagate harmful stereotypes that warrant further systematic research, which we did not have the resources for given the scope of this paper. We fully intend to follow up and report on these potential findings in the near future as part of our efforts to ensure equitable representation and socio-culturally appropriate outputs.

Despite these limitations, it is our hope that these initial findings can provide some indication of the current state-of-the-art multilingual abilities of LLMs in SEA languages and that this benchmark can be used to evaluate other multilingual LLMs in a holistic manner as well. We also hope that the findings can draw attention to the glaring need for better quality datasets for SEA languages. Given the inadequate performance of these state-of-the-art LLMs on SEA languages in key NLP competencies, linguistic phenomena (syntax, semantics and pragmatics), as well as cultural representation and sensitivity, if researchers and practitioners are to use them in SEA contexts, it might be necessary to further fine-tune them or to create a SEA-specific LLM.

## Acknowledgements

This work is supported by the National Research Foundation, Singapore under its AI Singapore Programme. The authors would like to thank the following people for their valuable time in assisting with this research: Thanh Ngan Nguyen and Charin Polpanumas for their help with the vetting and translation of prompts, Arvin Febriyan, Duong Thi Kim Cuc, Anne Nguyen, Trung Nguyen Tran, Can Udomcharoenchaikit, Surapon Nonesung, and Kanruethai Masuk for their assistance with the human evaluation of NLG, and Jani, Eli, Sherly, Agneta, Lavanya Rengarajan, Sundararajan Rengarajan and Jancy Rani Rengarajan, Vishwanath Magendran, S. M. Naheswari, R. Aarthi and Sharanyaa Sarveswaran for their help with annotating and vetting the handcrafted examples in LINDSEA. Last but not least, the authors would also like to thank Xianbin Yong for providing technical support in the formatting of this paper.

## References

[1] Tom B. Brown, Benjamin Mann, Nick Ryder, Melanie Subbiah, Jared Kaplan, Prafulla Dhariwal, Arvind Neelakantan, Pranav Shyam, Girish Sastry, Amanda Askell, Sandhini Agarwal, Ariel Herbert-Voss, Gretchen Krueger, Tom Henighan, Rewon Child, Aditya Ramesh, Daniel M. Ziegler, Jeffrey Wu, Clemens Winter, Christopher Hesse, Mark Chen, Eric Sigler, Mateusz Litwin, Scott Gray, Benjamin Chess, Jack Clark, Christopher Berner, Sam McCandlish, Alec Radford, Ilya Sutskever, and Dario Amodei. Language Models are Few-Shot Learners, July 2020. arXiv:2005.14165 [cs].

[2] Yuntao Bai, Andy Jones, Kamal Ndousse, Amanda Askell, Anna Chen, Nova DasSarma, Dawn Drain, Stanislav Fort, Deep Ganguli, Tom Henighan, Nicholas Joseph, Saurav Kadavath, Jackson Kernion, Tom Conerly, Sheer El-Showk, Nelson Elhage, Zac Hatfield-Dodds, Danny Hernandez, Tristan Hume, Scott Johnston, Shauna Kravec, Liane Lovitt, Neel Nanda, Catherine Olsson, Dario Amodei, Tom Brown, Jack Clark, Sam McCandlish, Chris Olah, Ben Mann, and Jared Kaplan. Training a Helpful and

Harmless Assistant with Reinforcement Learning from Human Feedback, April 2022. arXiv:2204.05862 [cs].

[3] Aakanksha Chowdhery, Sharan Narang, Jacob Devlin, Maarten Bosma, Gaurav Mishra, Adam Roberts, Paul Barham, Hyung Won Chung, Charles Sutton, Sebastian Gehrmann, Parker Schuh, Kensen Shi, Sasha Tsvyashchenko, Joshua Maynez, Abhishek Rao, Parker Barnes, Yi Tay, Noam Shazeer, Vinodkumar Prabhakaran, Emily Reif, Nan Du, Ben Hutchinson, Reiner Pope, James Bradbury, Jacob Austin, Michael Isard, Guy Gur-Ari, Pengcheng Yin, Toju Duke, Anselm Levskaya, Sanjay Ghemawat, Sunipa Dev, Henryk Michalewski, Xavier Garcia, Vedant Misra, Kevin Robinson, Liam Fedus, Denny Zhou, Daphne Ippolito, David Luan, Hyeontaek Lim, Barret Zoph, Alexander Spiridonov, Ryan Sepassi, David Dohan, Shivani Agrawal, Mark Omernick, Andrew M. Dai, Thanumalayan Sankaranarayana Pillai, Marie Pellat, Aitor Lewkowycz, Erica Moreira, Rewon Child, Oleksandr Polozov, Katherine Lee, Zongwei Zhou, Xuezhi Wang, Brennan Saeta, Mark Diaz, Orhan Firat, Michele Catasta, Jason Wei, Kathy Meier-Hellstern, Douglas Eck, Jeff Dean, Slav Petrov, and Noah Fiedel. PaLM: Scaling Language Modeling with Pathways, October 2022. arXiv:2204.02311 [cs].

[4] Susan Zhang, Stephen Roller, Naman Goyal, Mikel Artetxe, Moya Chen, Shuohui Chen, Christopher Dewan, Mona Diab, Xian Li, Xi Victoria Lin, Todor Mihaylov, Myle Ott, Sam Shleifer, Kurt Shuster, Daniel Simig, Punit Singh Koura, Anjali Sridhar, Tianlu Wang, and Luke Zettlemoyer. OPT: Open Pre-trained Transformer Language Models, June 2022. arXiv:2205.01068 [cs].

[5] Junjie Ye, Xuanting Chen, Nuo Xu, Can Zu, Zekai Shao, Shichun Liu, Yuhan Cui, Zeyang Zhou, Chao Gong, Yang Shen, Jie Zhou, Siming Chen, Tao Gui, Qi Zhang, and Xuanjing Huang. A Comprehensive Capability Analysis of GPT-3 and GPT-3.5 Series Models, 2023. arXiv:2303.10420 [cs].

[6] Jason Wei, Xuezhi Wang, Dale Schuurmans, Maarten Bosma, Brian Ichter, Fei Xia, Ed Chi, Quoc Le, and Denny Zhou. Chain-of-Thought Prompting Elicits Reasoning in Large Language Models, January 2023. arXiv:2201.11903 [cs].

[7] Aarohi Srivastava, Abhinav Rastogi, Abhishek Rao, Abu Awal Md Shoeb, Abubakar Abid, Adam Fisch, Adam R. Brown, Adam Santoro, Aditya Gupta, Adrià Garriga-Alonso, Agnieszka Kluska, Aitor Lewkowycz, Akshat Agarwal, Alethea Power, Alex Ray, Alex Warstadt, Alexander W. Kocurek, Ali Safaya, Ali Tazarv, Alice Xiang, Alicia Parrish, Allen Nie, Aman Hussain, Amanda Askell, Amanda Dsouza, Ambrose Slone, Ameet Rahane, Anantharaman S. Iyer, Anders Andreassen, Andrea Madotto, Andrea Santilli, Andreas Stuhlmüller, Andrew Dai, Andrew La, Andrew Lampinen, Andy Zou, Angela Jiang, Angelica Chen, Anh Vuong, Animesh Gupta, Anna Gottardi, Antonio Norelli, Anu Venkatesh, Arash Gholamidavoodi, Arfa Tabassum, Arul Menezes, Arun Kirubarajan, Asher Mullokandov, Ashish Sabharwal, Austin Herrick, Avia Efrat, Aykut Erdem, Ayla Karakaş, B. Ryan Roberts, Bao Sheng Loe, Barret Zoph, Bartłomiej Bojanowski, Batuhan Özyurt, Behnam Hedayatnia, Behnam Neyshabur, Benjamin Inden, Benno Stein, Berk Ekmekci, Bill Yuchen Lin, Blake Howald, Bryan Orinion, Cameron Diao, Cameron Dour, Catherine Stinson, Cedric Argueta, César Ferri Ramírez, Chandan Singh, Charles Rathkopf, Chenlin Meng, Chitta Baral, Chiyu Wu, Chris Callison-Burch, Chris Waites, Christian Voigt, Christopher D. Manning, Christopher Potts, Cindy Ramirez, Clara E. Rivera, Clemencia Siro, Colin Raffel, Courtney Ashcraft, Cristina Garbacea, Damien Sileo, Dan Garrette, Dan Hendrycks, Dan Kilman, Dan Roth, Daniel Freeman, Daniel Khashabi, Daniel Levy, Daniel Moseguí González, Danielle Perszyk, Danny Hernandez, Danqi Chen, Daphne Ippolito, Dar Gilboa, David Dohan, David Drakard, David Jurgens, Debajyoti Datta, Deep Ganguli, Denis Emelin, Denis Kleyko, Deniz Yuret, Derek Chen, Derek Tam, Dieuwke Hupkes, Diganta Misra, Dilyar Buzan, Dimitri Coelho Mollo, Diyi Yang, Dong-Ho Lee, Dylan Schrader, Ekaterina Shutova, Ekin Dogus Cubuk, Elad Segal, Eleanor Hagerman, Elizabeth Barnes, Elizabeth Donoway, Ellie Pavlick, Emanuele Rodola, Emma Lam, Eric Chu, Eric Tang, Erkut Erdem, Ernie Chang, Ethan A. Chi, Ethan Dyer, Ethan Jerzak, Ethan Kim, Eunice Engefu Manyasi, Evgenii Zheltonozhskii, Fanyue Xia, Fatemeh Siar, Fernando Martínez-Plumed, Francesca Happé, Francois Chollet, Frieda Rong, Gaurav Mishra, Genta Indra Winata, Gerard de Melo, Germán Kruszewski, Giambattista Parascandolo, Giorgio Mariani, Gloria Wang, Gonzalo Jaimovitch-López, Gregor Betz, Guy Gur-Ari, Hana Galijasevic, Hannah Kim, Hannah Rashkin, Hannaneh Hajishirzi, Harsh Mehta, Hayden Bogar, Henry Shevlin, Hinrich Schütze, Hiromu Yakura, Hongming Zhang, Hugh Mee Wong, Ian Ng, Isaac Noble, Jaap Jumelet, Jack Geissinger, Jackson Kernion, Jacob Hilton, Jaehoon Lee, Jaime Fernández Fisac, James B. Simon, James Koppel, James Zheng, James Zou, Jan Kocoń, Jana Thompson, Janelle Wingfield, Jared Kaplan, Jarema Radom, Jascha Sohl-Dickstein, Jason Phang, Jason Wei, Jason Yosinski, Jekaterina Novikova, Jelle Bosscher, Jennifer Marsh, Jeremy Kim, Jeroen Taal, Jesse Engel, Jesujoba Alabi, Jiacheng Xu, Jiaming Song, Jillian Tang, Joan Waweru, John Burden, John Miller, John U. Balis, Jonathan Batchelder, Jonathan Berant, Jörg Frohberg, Jos

Rozen, Jose Hernandez-Orallo, Joseph Boudeman, Joseph Guerr, Joseph Jones, Joshua B. Tenenbaum, Joshua S. Rule, Joyce Chua, Kamil Kanclerz, Karen Livescu, Karl Krauth, Karthik Gopalakrishnan, Katerina Ignatyeva, Katja Markert, Kaustubh D. Dhole, Kevin Gimpel, Kevin Omondi, Kory Mathewson, Kristen Chiafullo, Ksenia Shkaruta, Kumar Shridhar, Kyle McDonell, Kyle Richardson, Laria Reynolds, Leo Gao, Li Zhang, Liam Dugan, Lianhui Qin, Lidia Contreras-Ochando, Louis-Philippe Morency, Luca Moschella, Lucas Lam, Lucy Noble, Ludwig Schmidt, Luheng He, Luis Oliveros Colón, Luke Metz, Lütfi Kerem enel, Maarten Bosma, Maarten Sap, Maartje ter Hoeve, Maheen Farooqi, Manaal Faruqui, Mantas Mazeika, Marco Baturan, Marco Marelli, Marco Maru, Maria Jose Ramírez Quintana, Marie Tolkiehn, Mario Giulianelli, Martha Lewis, Martin Potthast, Matthew L. Leavitt, Matthias Hagen, Mátyás Schubert, Medina Orduna Baitemirova, Melody Arnaud, Melvin McElrath, Michael A. Yee, Michael Cohen, Michael Gu, Michael Ivanitskiy, Michael Starritt, Michael Strube, Micha Swdrowski, Michele Bevilacqua, Michihiro Yasunaga, Mihir Kale, Mike Cain, Mimee Xu, Mirac Suzgun, Mitch Walker, Mo Tiwari, Mohit Bansal, Moin Aminnaseri, Mor Geva, Mozhdeh Gheini, Mukund Varma T, Nanyun Peng, Nathan A. Chi, Nayeon Lee, Neta Gur-Ari Krakover, Nicholas Cameron, Nicholas Roberts, Nick Doiron, Nicole Martinez, Nikita Nangia, Niklas Deckers, Niklas Muennighoff, Nitish Shirish Keskar, Niveditha S. Iyer, Noah Constant, Noah Fiedel, Nuan Wen, Oliver Zhang, Omar Agha, Omar Elbaghdadi, Omer Levy, Owain Evans, Pablo Antonio Moreno Casares, Parth Doshi, Pascale Fung, Paul Pu Liang, Paul Vicol, Pegah Alipoormolabashi, Peiyuan Liao, Percy Liang, Peter Chang, Peter Eckersley, Phu Mon Htut, Pinyu Hwang, Piotr Mikowski, Piyush Patil, Pouya Pezeshkpour, Priti Oli, Qiaozhu Mei, Qing Lyu, Qinlang Chen, Rabin Banjade, Rachel Etta Rudolph, Raefer Gabriel, Rahel Habacker, Ramon Risco, Raphaël Millière, Rhythm Garg, Richard Barnes, Rif A. Saurous, Riku Arakawa, Robbe Raymaekers, Robert Frank, Rohan Sikand, Roman Novak, Roman Sitelew, Ronan LeBras, Rosanne Liu, Rowan Jacobs, Rui Zhang, Ruslan Salakhutdinov, Ryan Chi, Ryan Lee, Ryan Stovall, Ryan Teehan, Rylan Yang, Sahib Singh, Saif M. Mohammad, Sajant Anand, Sam Dillavou, Sam Shleifer, Sam Wiseman, Samuel Gruetter, Samuel R. Bowman, Samuel S. Schoenholz, Sanghyun Han, Sanjeev Kwatra, Sarah A. Rous, Sarik Ghazarian, Sayan Ghosh, Sean Casey, Sebastian Bischoff, Sebastian Gehrmann, Sebastian Schuster, Sepideh Sadeghi, Shadi Hamdan, Sharon Zhou, Shashank Srivastava, Sherry Shi, Shikhar Singh, Shima Asaadi, Shixiang Shane Gu, Shubh Pachchigar, Shubham Toshniwal, Shyam Upadhyay, Shyamolima, Debnath, Siamak Shakeri, Simon Thormeyer, Simone Melzi, Siva Reddy, Sneha Priscilla Makini, Soo-Hwan Lee, Spencer Torene, Sriharsha Hatwar, Stanislas Dehaene, Stefan Divic, Stefano Ermon, Stella Biderman, Stephanie Lin, Stephen Prasad, Steven T. Piantadosi, Stuart M. Shieber, Summer Misherghi, Svetlana Kiritchenko, Swaroop Mishra, Tal Linzen, Tal Schuster, Tao Li, Tao Yu, Tariq Ali, Tatsu Hashimoto, Te-Lin Wu, Théo Desbordes, Theodore Rothschild, Thomas Phan, Tianle Wang, Tiberius Nkinyili, Timo Schick, Timofei Kornev, Titus Tunduny, Tobias Gerstenberg, Trenton Chang, Trishala Neeraj, Tushar Khot, Tyler Shultz, Uri Shaham, Vedant Misra, Vera Demberg, Victoria Nyamai, Vikas Raunak, Vinay Ramasesh, Vinay Uday Prabhu, Vishakh Padmakumar, Vivek Srikumar, William Fedus, William Saunders, William Zhang, Wout Vossen, Xiang Ren, Xiaoyu Tong, Xinran Zhao, Xinyi Wu, Xudong Shen, Yadollah Yaghoobzadeh, Yair Lakretz, Yangqiu Song, Yasaman Bahri, Yejin Choi, Yichi Yang, Yiding Hao, Yifu Chen, Yonatan Belinkov, Yu Hou, Yufang Hou, Yuntao Bai, Zachary Seid, Zhuoye Zhao, Zijian Wang, Zijie J. Wang, Zirui Wang, and Ziyi Wu. Beyond the Imitation Game: Quantifying and Extrapolating the Capabilities of Language Models, June 2023. arXiv:2206.04615 [cs, stat].

[8] Percy Liang, Rishi Bommasani, Tony Lee, Dimitris Tsipras, Dilara Soylu, Michihiro Yasunaga, Yian Zhang, Deepak Narayanan, Yuhuai Wu, Ananya Kumar, Benjamin Newman, Binhang Yuan, Bobby Yan, Ce Zhang, Christian Cosgrove, Christopher D. Manning, Christopher Ré, Diana Acosta-Navas, Drew A. Hudson, Eric Zelikman, Esin Durmus, Faisal Ladhak, Frieda Rong, Hongyu Ren, Huaxiu Yao, Jue Wang, Keshav Santhanam, Laurel Orr, Lucia Zheng, Mert Yuksekgonul, Mirac Suzgun, Nathan Kim, Neel Guha, Niladri Chatterji, Omar Khattab, Peter Henderson, Qian Huang, Ryan Chi, Sang Michael Xie, Shibani Santurkar, Surya Ganguli, Tatsunori Hashimoto, Thomas Icard, Tianyi Zhang, Vishrav Chaudhary, William Wang, Xuechen Li, Yifan Mai, Yuhui Zhang, and Yuta Koreeda. Holistic Evaluation of Language Models, November 2022. arXiv:2211.09110 [cs].

[9] Kabir Ahuja, Harshita Diddee, Rishav Hada, Millicent Ochieng, Krithika Ramesh, Prachi Jain, Akshay Nambi, Tanuja Ganu, Sameer Segal, Maxamed Axmed, Kalika Bali, and Sunayana Sitaram. MEGA: Multilingual Evaluation of Generative AI, May 2023. arXiv:2303.12528 [cs].

[10] Akari Asai, Sneha Kudugunta, Xinyan Velocity Yu, Terra Blevins, Hila Gonen, Machel Reid, Yulia Tsvetkov, Sebastian Ruder, and Hannaneh Hajishirzi. BUFFET: Benchmarking Large Language Models for Few-shot Cross-lingual Transfer, May 2023. arXiv:2305.14857 [cs].

[11] Yejin Bang, Samuel Cahyawijaya, Nayeon Lee, Wenliang Dai, Dan Su, Bryan Wilie, Holy Lovenia, Ziwei Ji, Tiezheng Yu, Willy Chung, Quyet V. Do, Yan Xu, and Pascale Fung. A Multitask, Multilingual, Multimodal Evaluation of ChatGPT on Reasoning, Hallucination, and Interactivity, February 2023. arXiv:2302.04023 [cs].

[12] Viet Dac Lai, Nghia Trung Ngo, Amir Pouran Ben Veyseh, Hieu Man, Franck Dernoncourt, Trung Bui, and Thien Huu Nguyen. ChatGPT Beyond English: Towards a Comprehensive Evaluation of Large Language Models in Multilingual Learning, April 2023. arXiv:2304.05613 [cs].

[13] Wenxuan Zhang, Sharifah Mahani Aljunied, Chang Gao, Yew Ken Chia, and Lidong Bing. M3Exam: A Multilingual, Multimodal, Multilevel Benchmark for Examining Large Language Models, June 2023. arXiv:2306.05179 [cs].

[14] Rui Mao, Guanyi Chen, Xulang Zhang, Frank Guerin, and Erik Cambria. GPTEval: A Survey on Assessments of ChatGPT and GPT-4, August 2023. arXiv:2308.12488 [cs].

[15] OpenAI. GPT-4 Technical Report, March 2023. arXiv:2303.08774 [cs].

[16] Ashish Vaswani, Noam Shazeer, Niki Parmar, Jakob Uszkoreit, Llion Jones, Aidan N Gomez, Ł ukasz Kaiser, and Illia Polosukhin. Attention is All You Need. In I. Guyon, U. Von Luxburg, S. Bengio, H. Wallach, R. Fergus, S. Vishwanathan, and R. Garnett, editors, *Advances in Neural Information Processing Systems*, volume 30. Curran Associates, Inc., 2017.

[17] Jacob Devlin, Ming-Wei Chang, Kenton Lee, and Kristina Toutanova. Bert: Pre-training of Deep Bidirectional Transformers for Language Understanding. In *Proceedings of the 2019 Conference of the North American Chapter of the Association for Computational Linguistics: Human Language Technologies, Volume 1 (Long and Short Papers)*, pages 4171–4186, Minneapolis, Minnesota, June 2019. Association for Computational Linguistics.

[18] Alex Wang, Amanpreet Singh, Julian Michael, Felix Hill, Omer Levy, and Samuel Bowman. GLUE: A Multi-Task Benchmark and Analysis Platform for Natural Language Understanding. In *Proceedings of the 2018 EMNLP Workshop BlackboxNLP: Analyzing and Interpreting Neural Networks for NLP*, pages 353–355, Brussels, Belgium, 2018. Association for Computational Linguistics.

[19] Alex Wang, Yada Pruksachatkun, Nikita Nangia, Amanpreet Singh, Julian Michael, Felix Hill, Omer Levy, and Samuel Bowman. SuperGLUE: A Stickier Benchmark for General-Purpose Language Understanding Systems. In H. Wallach, H. Larochelle, A. Beygelzimer, F. d'Alché-Buc, E. Fox, and R. Garnett, editors, *Advances in Neural Information Processing Systems*, volume 32. Curran Associates, Inc., 2019.

[20] Simone Tedeschi, Johan Bos, Thierry Declerck, Jan Hajic, Daniel Hershcovich, Eduard H. Hovy, Alexander Koller, Simon Krek, Steven Schockaert, Rico Sennrich, Ekaterina Shutova, and Roberto Navigli. What's the Meaning of Superhuman Performance in Today's NLU?, May 2023. arXiv:2305.08414 [cs].

[21] Alex Warstadt, Alicia Parrish, Haokun Liu, Anhad Mohananey, Wei Peng, Sheng-Fu Wang, and Samuel R. Bowman. BLiMP: The Benchmark of Linguistic Minimal Pairs for English. *Transactions of the Association for Computational Linguistics*, 8:377–392, December 2020.

[22] Yongliang Shen, Kaitao Song, Xu Tan, Dongsheng Li, Weiming Lu, and Yueting Zhuang. HuggingGPT: Solving AI Tasks with ChatGPT and its Friends in Hugging Face, May 2023. arXiv:2303.17580 [cs].

[23] Jessica López Espejel, El Hassane Ettifouri, Mahaman Sanoussi Yahaya Alassan, El Mehdi Chouham, and Walid Dahhane. GPT-3.5 vs GPT-4: Evaluating ChatGPT's Reasoning Performance in Zero-shot Learning, 2023. arXiv:2305.12477 [cs].

[24] Hanmeng Liu, Ruoxi Ning, Zhiyang Teng, Jian Liu, Qiji Zhou, and Yue Zhang. Evaluating the Logical Reasoning Ability of ChatGPT and GPT-4, May 2023. arXiv:2304.03439 [cs].

[25] Paloma Jeretič, Alex Warstadt, Suvrat Bhooshan, and Adina Williams. Are Natural Language Inference Models IMPPRESsive? Learning IMPlicature and PRESupposition. In *Proceedings of the 58th Annual Meeting of the Association for Computational Linguistics*, pages 8690–8705, Online, 2020. Association for Computational Linguistics.

[26] Lalchand Pandia, Yan Cong, and Allyson Ettinger. Pragmatic Competence of Pre-trained Language Models through the Lens of Discourse Connectives, September 2021. arXiv:2109.12951 [cs].

[27] Jennifer Hu, Roger Levy, Judith Degen, and Sebastian Schuster. Expectations over Unspoken Alternatives Predict Pragmatic Inferences, April 2023. arXiv:2304.04758 [cs].

[28] Wei Liu, Ming Xiang, and Nai Ding. Adjective Scale Probe: Can Language Models Encode Formal Semantics Information? *Proceedings of the AAAI Conference on Artificial Intelligence*, 37(11):13282–13290, June 2023.

[29] Alicia Parrish, Sebastian Schuster, Alex Warstadt, Omar Agha, Soo-Hwan Lee, Zhuoye Zhao, Samuel R. Bowman, and Tal Linzen. NOPE: A Corpus of Naturally-Occurring Presuppositions in English, September 2021. arXiv:2109.06987 [cs].

[30] Amr Hendy, Mohamed Abdelrehim, Amr Sharaf, Vikas Raunak, Mohamed Gabr, Hitokazu Matsushita, Young Jin Kim, Mohamed Afify, and Hany Hassan Awadalla. How Good Are GPT Models at Machine Translation? A Comprehensive Evaluation, February 2023. arXiv:2302.09210 [cs].

[31] Maja Popović. ChrF: Character n-gram F-score for Automatic MT Evaluation. In *Proceedings of the Tenth Workshop on Statistical Machine Translation*, pages 392–395, Lisbon, Portugal, September 2015. Association for Computational Linguistics.

[32] Ricardo Rei, José GC De Souza, Duarte Alves, Chrysoula Zerva, Ana C Farinha, Taisiya Glushkova, Alon Lavie, Luisa Coheur, and André FT Martins. Comet-22: Unbabel-IST 2022 Submission for the Metrics Shared Task. In *Proceedings of the Seventh Conference on Machine Translation (WMT)*, pages 578–585, 2022.

[33] Ricardo Rei, Marcos Treviso, Nuno M. Guerreiro, Chrysoula Zerva, Ana C Farinha, Christine Maroti, José G. C. de Souza, Taisiya Glushkova, Duarte Alves, Luisa Coheur, Alon Lavie, and André F. T. Martins. Cometkiwi: IST-Unbabel 2022 Submission for the Quality Estimation Shared Task. In *Proceedings of the Seventh Conference on Machine Translation (WMT)*, pages 634–645, Abu Dhabi, United Arab Emirates (Hybrid), December 2022. Association for Computational Linguistics.

[34] Shuang Li, Jiangjie Chen, Siyu Yuan, Xinyi Wu, Hao Yang, Shimin Tao, and Yanghua Xiao. Translate Meanings, Not Just Words: IdiomKB's Role in Optimizing Idiomatic Translation with Language Models, August 2023. arXiv:2308.13961 [cs].

[35] Tianyi Zhang, Faisal Ladhak, Esin Durmus, Percy Liang, Kathleen McKeown, and Tatsunori B. Hashimoto. Benchmarking Large Language Models for News Summarization, Jan 2023. arXiv:2301.13848 [cs].

[36] Long Ouyang, Jeff Wu, Xu Jiang, Diogo Almeida, Carroll L. Wainwright, Pamela Mishkin, Chong Zhang, Sandhini Agarwal, Katarina Slama, Alex Ray, John Schulman, Jacob Hilton, Fraser Kelton, Luke Miller, Maddie Simens, Amanda Askell, Peter Welinder, Paul Christiano, Jan Leike, and Ryan Lowe. Training Language Models to Follow Instructions with Human Feedback, March 2022. arXiv:2203.02155 [cs].

[37] Alexander R. Fabbri, Wojciech Kryciski, Bryan McCann, Caiming Xiong, Richard Socher, and Dragomir Radev. SummEval: Re-evaluating Summarization Evaluation, February 2021. arXiv:2007.12626 [cs].

[38] Jordan Kodner, Sarah Payne, and Jeffrey Heinz. Why Linguistics Will Thrive in the 21st Century: A Reply to Piantadosi (2023), August 2023. arXiv:2308.03228 [cs].

[39] Jennifer Hu, Jon Gauthier, Peng Qian, Ethan Wilcox, and Roger Levy. A Systematic Assessment of Syntactic Generalization in Neural Language Models. In *Proceedings of the 58th Annual Meeting of the Association for Computational Linguistics*, pages 1725–1744, Online, 2020. Association for Computational Linguistics.

[40] Jon Gauthier, Jennifer Hu, Ethan Wilcox, Peng Qian, and Roger Levy. SyntaxGym: An Online Platform for Targeted Evaluation of Language Models. In *Proceedings of the 58th Annual Meeting of the Association for Computational Linguistics: System Demonstrations*, pages 70–76, Online, 2020. Association for Computational Linguistics.

[41] Beilei Xiang, Changbing Yang, Yu Li, Alex Warstadt, and Katharina Kann. CLiMP: A Benchmark for Chinese Language Model Evaluation. In *Proceedings of the 16th Conference of the European Chapter of the Association for Computational Linguistics: Main Volume*, pages 2784–2790, Online, 2021. Association for Computational Linguistics.

[42] Taiga Someya and Yohei Oseki. JBLiMP: Japanese Benchmark of Linguistic Minimal Pairs. In *Findings of the Association for Computational Linguistics: EACL 2023*, pages 1581–1594, Dubrovnik, Croatia, May 2023. Association for Computational Linguistics.

[43] Daniel Hershcovich, Stella Frank, Heather Lent, Miryam De Lhoneux, Mostafa Abdou, Stephanie Brandl, Emanuele Bugliarello, Laura Cabello Piqueras, Ilias Chalkidis, Ruixiang Cui, Constanza Fierro,

Katerina Margatina, Phillip Rust, and Anders Søgaard. Challenges and Strategies in Cross-Cultural NLP. In *Proceedings of the 60th Annual Meeting of the Association for Computational Linguistics (Volume 1: Long Papers)*, pages 6997–7013, Dublin, Ireland, 2022. Association for Computational Linguistics.

[44] Irene Solaiman, Zeerak Talat, William Agnew, Lama Ahmad, Dylan Baker, Su Lin Blodgett, Hal Daumé III, Jesse Dodge, Ellie Evans, Sara Hooker, Yacine Jernite, Alexandra Sasha Luccioni, Alberto Lusoli, Margaret Mitchell, Jessica Newman, Marie-Therese Png, Andrew Strait, and Apostol Vassilev. Evaluating the Social Impact of Generative AI Systems in Systems and Society, June 2023. arXiv:2306.05949 [cs].

[45] Esin Durmus, Karina Nyugen, Thomas I. Liao, Nicholas Schiefer, Amanda Askell, Anton Bakhtin, Carol Chen, Zac Hatfield-Dodds, Danny Hernandez, Nicholas Joseph, Liane Lovitt, Sam McCandlish, Orowa Sikder, Alex Tamkin, Janel Thamkul, Jared Kaplan, Jack Clark, and Deep Ganguli. Towards Measuring the Representation of Subjective Global Opinions in Language Models, June 2023. arXiv:2306.16388 [cs].

[46] Dan Hendrycks, Collin Burns, Steven Basart, Andy Zou, Mantas Mazeika, Dawn Song, and Jacob Steinhardt. Measuring Massive Multitask Language Understanding, January 2021. arXiv:2009.03300 [cs].

[47] Jie Huang and Kevin Chen-Chuan Chang. Towards Reasoning in Large Language Models: A Survey, May 2023. arXiv:2212.10403 [cs].

[48] Shuofei Qiao, Yixin Ou, Ningyu Zhang, Xiang Chen, Yunzhi Yao, Shumin Deng, Chuanqi Tan, Fei Huang, and Huajun Chen. Reasoning with Language Model Prompting: A Survey. In *Proceedings of the 61st Annual Meeting of the Association for Computational Linguistics (Volume 1: Long Papers)*, pages 5368–5393, Toronto, Canada, 2023. Association for Computational Linguistics.

[49] Fangzhi Xu, Qika Lin, Jiawei Han, Tianzhe Zhao, Jun Liu, and Erik Cambria. Are Large Language Models Really Good Logical Reasoners? A Comprehensive Evaluation and Beyond, August 2023. arXiv:2306.09841 [cs].

[50] Fei Yu, Hongbo Zhang, Prayag Tiwari, and Benyou Wang. Natural Language Reasoning, A Survey, May 2023. arXiv:2303.14725 [cs].

[51] Jifan Yu, Xiaozhi Wang, Shangqing Tu, Shulin Cao, Daniel Zhang-Li, Xin Lv, Hao Peng, Zijun Yao, Xiaohan Zhang, Hanming Li, Chunyang Li, Zheyuan Zhang, Yushi Bai, Yantao Liu, Amy Xin, Nianyi Lin, Kaifeng Yun, Linlu Gong, Jianhui Chen, Zhili Wu, Yunjia Qi, Weikai Li, Yong Guan, Kaisheng Zeng, Ji Qi, Hailong Jin, Jinxin Liu, Yu Gu, Yuan Yao, Ning Ding, Lei Hou, Zhiyuan Liu, Bin Xu, Jie Tang, and Juanzi Li. KoLA: Carefully Benchmarking World Knowledge of Large Language Models, July 2023. arXiv:2306.09296 [cs].

[52] Pratik Joshi, Sebastin Santy, Amar Budhiraja, Kalika Bali, and Monojit Choudhury. The State and Fate of Linguistic Diversity and Inclusion in the NLP World. In *Proceedings of the 58th Annual Meeting of the Association for Computational Linguistics*, pages 6282–6293, Online, 2020. Association for Computational Linguistics.

[53] Martin Gellerstam. Translationese in Swedish Novels Translated from English. *Translation studies in Scandinavia*, 1:88–95, 1986.

[54] Mona Baker. Corpus Linguistics and Translation Studies - Implications and Applications. In Mona Baker, Gill Francis, and Elena Tognini-Bonelli, editors, *Text and Technology*, page 233. John Benjamins Publishing Company, Amsterdam, 1993.

[55] Gennadi Lembersky, Noam Ordan, and Shuly Wintner. Language Models for Machine Translation: Original vs. Translated Texts. *Computational Linguistics*, 38(4):799–825, December 2012.

[56] V. Volansky, N. Ordan, and S. Wintner. On the Features of Translationese. *Digital Scholarship in the Humanities*, 30(1):98–118, April 2015.

[57] Parker Riley, Isaac Caswell, Markus Freitag, and David Grangier. Translationese as a Language in Multilingual NMT. In *Proceedings of the 58th Annual Meeting of the Association for Computational Linguistics*, pages 7737–7746, Online, 2020. Association for Computational Linguistics.

[58] Xiaoman Pan, Boliang Zhang, Jonathan May, Joel Nothman, Kevin Knight, and Heng Ji. Cross-lingual Name Tagging and Linking for 282 Languages. In *Proceedings of the 55th Annual Meeting of the Association for Computational Linguistics (Volume 1: Long Papers)*, pages 1946–1958, Vancouver, Canada, 2017. Association for Computational Linguistics.

[59] Junjie Hu, Sebastian Ruder, Aditya Siddhant, Graham Neubig, Orhan Firat, and Melvin Johnson. XTREME: A Massively Multilingual Multi-task Benchmark for Evaluating Cross-lingual Generalization, September 2020. arXiv:2003.11080 [cs].

[60] Constantine Lignos, Nolan Holley, Chester Palen-Michel, and Jonne Sälevä. Toward More Meaningful Resources for Lower-resourced Languages. In *Findings of the Association for Computational Linguistics: ACL 2022*, pages 523–532, Dublin, Ireland, 2022. Association for Computational Linguistics.

[61] Bryan Wilie, Karissa Vincentio, Genta Indra Winata, Samuel Cahyawijaya, Xiaohong Li, Zhi Yuan Lim, Sidik Soleman, Rahmad Mahendra, Pascale Fung, Syafri Bahar, and Ayu Purwarianti. IndoNLU: Benchmark and Resources for Evaluating Indonesian Natural Language Understanding. In *Proceedings of the 1st Conference of the Asia-Pacific Chapter of the Association for Computational Linguistics and the 10th International Joint Conference on Natural Language Processing*, pages 843–857, Suzhou, China, dec 2020. Association for Computational Linguistics.

[62] Huyen T M Nguyen, Quyen T Ngo, Luong X Vu, Vu M Tran, and Hien T T Nguyen. VLSP Shared Task: Named Entity Recognition. *Journal of Computer Science and Cybernetics*, 34(4):283–294, January 2019.

[63] Wannaphong Phatthiyaphaibun. Thai NER 2.0, Sep 2022.

[64] Lalita Lowphansirikul, Charin Polpanumas, Nawat Jantrakulchai, and Sarana Nutanong. Wangchan-BERTa: Pretraining Transformer-based Thai Language Models, March 2021. arXiv:2101.09635 [cs].

[65] Arnav Mhaske, Harshit Kedia, Sumanth Doddapaneni, Mitesh M. Khapra, Pratyush Kumar, Rudra Murthy, and Anoop Kunchukuttan. Naamapadam: A Large-Scale Named Entity Annotated Data for Indic Languages. In *Proceedings of the 61st Annual Meeting of the Association for Computational Linguistics (Volume 1: Long Papers)*, pages 10441–10456, Toronto, Canada, 2023. Association for Computational Linguistics.

[66] Shuhe Wang, Xiaofei Sun, Xiaoya Li, Rongbin Ouyang, Fei Wu, Tianwei Zhang, Jiwei Li, and Guoyin Wang. GPT-NER: Named Entity Recognition via Large Language Models, May 2023. arXiv:2304.10428 [cs].

[67] Anna Rogers, Matt Gardner, and Isabelle Augenstein. QA Dataset Explosion: A Taxonomy of NLP Resources for Question Answering and Reading Comprehension. *ACM Computing Surveys*, 55(10):1–45, October 2023. arXiv:2107.12708 [cs].

[68] Jonathan H. Clark, Eunsol Choi, Michael Collins, Dan Garrette, Tom Kwiatkowski, Vitaly Nikolaev, and Jennimaria Palomaki. TyDi QA: A Benchmark for Information-Seeking Question Answering in Typologically Diverse Languages. *Transactions of the Association for Computational Linguistics*, 8:454–470, 2020.

[69] Mikel Artetxe, Sebastian Ruder, and Dani Yogatama. On the Cross-lingual Transferability of Monolingual Representations. In *Proceedings of the 58th Annual Meeting of the Association for Computational Linguistics*, pages 4623–4637, Online, 2020. Association for Computational Linguistics.

[70] Sumanth Doddapaneni, Rahul Aralikatte, Gowtham Ramesh, Shreya Goyal, Mitesh M. Khapra, Anoop Kunchukuttan, and Pratyush Kumar. Towards Leaving No Indic Language Behind: Building Monolingual Corpora, Benchmark and Models for Indic Languages. In *Proceedings of the 61st Annual Meeting of the Association for Computational Linguistics (Volume 1: Long Papers)*, pages 12402–12426, Toronto, Canada, July 2023. Association for Computational Linguistics.

[71] Genta Indra Winata, Alham Fikri Aji, Samuel Cahyawijaya, Rahmad Mahendra, Fajri Koto, Ade Romadhony, Kemal Kurniawan, David Moeljadi, Radityo Eko Prasojo, Pascale Fung, Timothy Baldwin, Jey Han Lau, Rico Sennrich, and Sebastian Ruder. NusaX: Multilingual Parallel Sentiment Dataset for 10 Indonesian Local Languages. In *Proceedings of the 17th Conference of the European Chapter of the Association for Computational Linguistics*, pages 815–834, Dubrovnik, Croatia, May 2023. Association for Computational Linguistics.

[72] Kiet Van Nguyen, Vu Duc Nguyen, Phu X. V. Nguyen, Tham T. H. Truong, and Ngan Luu-Thuy Nguyen. UIT-VSFC: Vietnamese Students Feedback Corpus for Sentiment Analysis. In *2018 10th International Conference on Knowledge and Systems Engineering (KSE)*, pages 19–24, Ho Chi Minh City, November 2018. IEEE.

[73] Bact', Pattarawat Chormai, Charin, and ekapolc. Pythainlp/wisesight-sentiment: First release, sep 2019.

[74] Yun Yu Chong and Haewoon Kwak. Understanding Toxicity Triggers on Reddit in the Context of Singapore. *Proceedings of the International AAAI Conference on Web and Social Media*, 16:1383–1387, May 2022.

[75] Muhammad Okky Ibrohim and Indra Budi. Multi-label Hate Speech and Abusive Language Detection in Indonesian Twitter. In *Proceedings of the Third Workshop on Abusive Language Online*, pages 46–57, Florence, Italy, 2019. Association for Computational Linguistics.

[76] Son T Luu, Kiet Van Nguyen, and Ngan Luu-Thuy Nguyen. A Large-scale Dataset for Hate Speech Detection on Vietnamese Social Media Texts. In *Advances and Trends in Artificial Intelligence. Artificial Intelligence Practices: 34th International Conference on Industrial, Engineering and Other Applications of Applied Intelligent Systems, IEA/AIE 2021, Kuala Lumpur, Malaysia, July 26–29, 2021, Proceedings, Part I 34*, pages 415–426. Springer, 2021.

[77] Sugan Sirihattasak, Mamoru Komachi, and Hiroshi Ishikawa. Annotation and Classification of Toxicity for Thai Twitter. In *TA-COS 2018: 2nd Workshop on Text Analytics for Cybersecurity and Online Safety*, pages 1–7, 2018.

[78] Wenhao Zhu, Hongyi Liu, Qingxiu Dong, Jingjing Xu, Shujian Huang, Lingpeng Kong, Jiajun Chen, and Lei Li. Multilingual Machine Translation with Large Language Models: Empirical Results and Analysis, May 2023. arXiv:2304.04675 [cs].

[79] NLLB Team, Marta R Costa-jussà, James Cross, Onur Çelebi, Maha Elbayad, Kenneth Heafield, Kevin Heffernan, Elahe Kalbassi, Janice Lam, Daniel Licht, Jean Maillard, Anna Sun, Skyler Wang, Guillaume Wenzek, Al Youngblood, Bapi Akula, Loic Barrault, Gabriel Mejia Gonzalez, Prangthip Hansanti, John Hoffman, Semarley Jarrett, Kaushik Ram Sadagopan, Dirk Rowe, Shannon Spruit, Chau Tran, Pierre Andrews, Necip Fazil Ayan, Shruti Bhosale, Sergey Edunov, Angela Fan, Cynthia Gao, Vedanuj Goswami, Francisco Guzmán, Philipp Koehn, Alexandre Mourachko, Christophe Ropers, Holger Saleem, Safiyyah andSchwenk, and Jeff Wang. No Language Left Behind: Scaling Human-Centered Machine Translation. -, 2022.

[80] Tahmid Hasan, Abhik Bhattacharjee, Md. Saiful Islam, Kazi Mubasshir, Yuan-Fang Li, Yong-Bin Kang, M. Sohel Rahman, and Rifat Shahriyar. XL-Sum: Large-Scale Multilingual Abstractive Summarization for 44 Languages. In *Findings of the Association for Computational Linguistics: ACL-IJCNLP 2021*, pages 4693–4703, Online, 2021. Association for Computational Linguistics.

[81] Christopher D Manning. Local Textual Inference: Its Hard to Circumscribe, But You Know It When You See It – And NLP Needs It. -, 2006.

[82] Adam Poliak. A Survey on Recognizing Textual Entailment as an NLP Evaluation. In *Proceedings of the First Workshop on Evaluation and Comparison of NLP Systems*, pages 92–109, Online, 2020. Association for Computational Linguistics.

[83] Rahmad Mahendra, Alham Fikri Aji, Samuel Louvan, Fahrurrozi Rahman, and Clara Vania. IndoNLI: A Natural Language Inference Dataset for Indonesian. In *Proceedings of the 2021 Conference on Empirical Methods in Natural Language Processing*, pages 10511–10527, Online and Punta Cana, Dominican Republic, 2021. Association for Computational Linguistics.

[84] Alexis Conneau, Ruty Rinott, Guillaume Lample, Adina Williams, Samuel Bowman, Holger Schwenk, and Veselin Stoyanov. XNLI: Evaluating Cross-lingual Sentence Representations. In *Proceedings of the 2018 Conference on Empirical Methods in Natural Language Processing*, pages 2475–2485, Brussels, Belgium, 2018. Association for Computational Linguistics.

[85] Divyanshu Aggarwal, Vivek Gupta, and Anoop Kunchukuttan. Indic XNLI: Evaluating Multilingual Inference for Indian Languages. In *Proceedings of the 2022 Conference on Empirical Methods in Natural Language Processing*, pages 10994–11006, Abu Dhabi, United Arab Emirate, December 2022. Association for Computational Linguistics.

[86] Melissa Roemmele, Cosmin Adrian Bejan, and Andrew S Gordon. Choice of Plausible Alternatives: An Evaluation of Commonsense Causal Reasoning. In *2011 AAAI Spring Symposium Series*, 2011.

[87] Edoardo Maria Ponti, Goran Glava, Olga Majewska, Qianchu Liu, Ivan Vuli, and Anna Korhonen. XCOPA: A Multilingual Dataset for Causal Commonsense Reasoning. In *Proceedings of the 2020 Conference on Empirical Methods in Natural Language Processing (EMNLP)*, pages 2362–2376, Online, 2020. Association for Computational Linguistics.

[88] Jeff Wu, Long Ouyang, Daniel M. Ziegler, Nisan Stiennon, Ryan Lowe, Jan Leike, and Paul Christiano. Recursively Summarizing Books with Human Feedback, September 2021. arXiv:2109.10862 [cs].

[89] Keqin Peng, Liang Ding, Qihuang Zhong, Li Shen, Xuebo Liu, Min Zhang, Yuanxin Ouyang, and Dacheng Tao. Towards Making the Most of ChatGPT for Machine Translation, March 2023. arXiv:2303.13780 [cs].

[90] Takeshi Kojima, Shixiang Shane Gu, Machel Reid, Yutaka Matsuo, and Yusuke Iwasawa. Large Language Models are Zero-Shot Reasoners, 2023. arXiv:2205.11916 [cs].

[91] Jieyi Long. Large Language Model Guided Tree-of-Thought, May 2023. arXiv:2305.08291 [cs].

[92] Shunyu Yao, Dian Yu, Jeffrey Zhao, Izhak Shafran, Thomas L. Griffiths, Yuan Cao, and Karthik Narasimhan. Tree of Thoughts: Deliberate Problem Solving with Large Language Models, May 2023. arXiv:2305.10601 [cs].

[93] Yongchao Zhou, Andrei Ioan Muresanu, Ziwen Han, Keiran Paster, Silviu Pitis, Harris Chan, and Jimmy Ba. Large Language Models Are Human-Level Prompt Engineers, March 2023. arXiv:2211.01910 [cs].

[94] Jiachang Liu, Dinghan Shen, Yizhe Zhang, Bill Dolan, Lawrence Carin, and Weizhu Chen. What Makes Good In-Context Examples for GPT-3? In *Proceedings of Deep Learning Inside Out (DeeLIO 2022): The 3rd Workshop on Knowledge Extraction and Integration for Deep Learning Architectures*, pages 100–114, Dublin, Ireland and Online, 2022. Association for Computational Linguistics.

[95] Zhiyong Wu, Yaoxiang Wang, Jiacheng Ye, and Lingpeng Kong. Self-Adaptive In-Context Learning: An Information Compression Perspective for In-Context Example Selection and Ordering. In *Proceedings of the 61st Annual Meeting of the Association for Computational Linguistics (Volume 1: Long Papers)*, pages 1423–1436, Toronto, Canada, 2023. Association for Computational Linguistics.

[96] Yiming Zhang, Shi Feng, and Chenhao Tan. Active Example Selection for In-Context Learning. In *Proceedings of the 2022 Conference on Empirical Methods in Natural Language Processing*, pages 9134–9148, Abu Dhabi, United Arab Emirates, 2022. Association for Computational Linguistics.

[97] Zihao Zhao, Eric Wallace, Shi Feng, Dan Klein, and Sameer Singh. Calibrate Before Use: Improving Few-shot Performance of Language Models. In *International Conference on Machine Learning*, pages 12697–12706. PMLR, 2021.

[98] Yao Lu, Max Bartolo, Alastair Moore, Sebastian Riedel, and Pontus Stenetorp. Fantastically Ordered Prompts and Where to Find Them: Overcoming Few-Shot Prompt Order Sensitivity. In *Proceedings of the 60th Annual Meeting of the Association for Computational Linguistics (Volume 1: Long Papers)*, pages 8086–8098, Dublin, Ireland, 2022. Association for Computational Linguistics.

[99] Ananya B. Sai, Akash Kumar Mohankumar, and Mitesh M. Khapra. A Survey of Evaluation Metrics Used for NLG Systems. *ACM Computing Surveys*, 55(2):1–39, feb 2023.

[100] Chin-Yew Lin. Rouge: A Package for Automatic Evaluation of Summaries. In *Text summarization branches out*, pages 74–81, 2004.

[101] Max Grusky. Rogue Scores. In *Proceedings of the 61st Annual Meeting of the Association for Computational Linguistics (Volume 1: Long Papers)*, pages 1914–1934, 2023.

[102] Markus Freitag, Ricardo Rei, Nitika Mathur, Chi-kiu Lo, Craig Stewart, Eleftherios Avramidis, Tom Kocmi, George Foster, Alon Lavie, and André FT Martins. Results of WMT22 Metrics Shared Task: Stop Using BLEU – Neural Metrics are Better and More Robust. In *Proceedings of the Seventh Conference on Machine Translation (WMT)*, pages 46–68, 2022.

[103] Maja Popović. ChrF++: Words Helping Character n-grams. In *Proceedings of the Second Conference on Machine Translation*, pages 612–618, Copenhagen, Denmark, September 2017. Association for Computational Linguistics.

[104] Tom Kocmi, Rachel Bawden, Ondřej Bojar, Anton Dvorkovich, Christian Federmann, Mark Fishel, Thamme Gowda, Yvette Graham, Roman Grundkiewicz, Barry Haddow, Rebecca Knowles, Philipp Koehn, Christof Monz, Makoto Morishita, Masaaki Nagata, Toshiaki Nakazawa, Michal Novák, Martin Popel, and Maja Popović. Findings of the 2022 Conference on Machine Translation (WMT22). In *Proceedings of the Seventh Conference on Machine Translation (WMT)*, pages 1–45, Abu Dhabi, United Arab Emirates (Hybrid), December 2022. Association for Computational Linguistics.

[105] Ricardo Rei, Craig Stewart, Ana C Farinha, and Alon Lavie. COMET: A Neural Framework for MT Evaluation. In *Proceedings of the 2020 Conference on Empirical Methods in Natural Language Processing (EMNLP)*, pages 2685–2702, Online, 2020. Association for Computational Linguistics.

[106] Wojciech Kryscinski, Bryan McCann, Caiming Xiong, and Richard Socher. Evaluating the Factual Consistency of Abstractive Text Summarization. In *Proceedings of the 2020 Conference on Empirical Methods in Natural Language Processing (EMNLP)*, pages 9332–9346, Online, 2020. Association for Computational Linguistics.

[107] Arle Richard Lommel, Aljoscha Burchardt, and Hans Uszkoreit. Multidimensional Quality Metrics: A Flexible System for Assessing Translation Quality. In *Proceedings of Translating and the Computer 35*, London, UK, November 2013. Aslib.

[108] Markus Freitag, George Foster, David Grangier, Viresh Ratnakar, Qijun Tan, and Wolfgang Macherey. Experts, Errors, and Context: A Large-scale Study of Human Evaluation for Machine Translation. *Transactions of the Association for Computational Linguistics*, 9:1460–1474, 2021.

[109] Yushi Bai, Jiahao Ying, Yixin Cao, Xin Lv, Yuze He, Xiaozhi Wang, Jifan Yu, Kaisheng Zeng, Yijia Xiao, Haozhe Lyu, Jiayin Zhang, Juanzi Li, and Lei Hou. Benchmarking Foundation Models with Language-Model-as-an-Examiner, June 2023. arXiv:2306.04181 [cs].

[110] Yi Chen, Rui Wang, Haiyun Jiang, Shuming Shi, and Ruifeng Xu. Exploring the Use of Large Language Models for Reference-Free Text Quality Evaluation: A Preliminary Empirical Study, April 2023. arXiv:2304.00723 [cs].

[111] Cheng-Han Chiang and Hung-yi Lee. Can Large Language Models Be an Alternative to Human Evaluations? In *Proceedings of the 61st Annual Meeting of the Association for Computational Linguistics (Volume 1: Long Papers)*, pages 15607–15631, Toronto, Canada, July 2023. Association for Computational Linguistics.

[112] Yunjie Ji, Yan Gong, Yiping Peng, Chao Ni, Peiyan Sun, Dongyu Pan, Baochang Ma, and Xiangang Li. Exploring ChatGPT's Ability to Rank Content: A Preliminary Study on Consistency with Human Preferences, March 2023. arXiv:2303.07610 [cs].

[113] Tom Kocmi and Christian Federmann. Large Language Models Are State-of-the-Art Evaluators of Translation Quality, May 2023. arXiv:2302.14520 [cs].

[114] Yang Liu, Dan Iter, Yichong Xu, Shuohang Wang, Ruochen Xu, and Chenguang Zhu. G-Eval: NLG Evaluation Using GPT-4 with Better Human Alignment, May 2023. arXiv:2303.16634 [cs].

[115] Jiaan Wang, Yunlong Liang, Fandong Meng, Zengkui Sun, Haoxiang Shi, Zhixu Li, Jinan Xu, Jianfeng Qu, and Jie Zhou. Is ChatGPT a Good NLG Evaluator? A Preliminary Study, April 2023. arXiv:2303.04048 [cs].

[116] Peiyi Wang, Lei Li, Liang Chen, Zefan Cai, Dawei Zhu, Binghuai Lin, Yunbo Cao, Qi Liu, Tianyu Liu, and Zhifang Sui. Large Language Models are Not Fair Evaluators, August 2023. arXiv:2305.17926 [cs].

[117] Lianmin Zheng, Wei-Lin Chiang, Ying Sheng, Siyuan Zhuang, Zhanghao Wu, Yonghao Zhuang, Zi Lin, Zhuohan Li, Dacheng Li, Eric P. Xing, Hao Zhang, Joseph E. Gonzalez, and Ion Stoica. Judging LLM-as-a-judge with MT-Bench and Chatbot Arena, July 2023. arXiv:2306.05685 [cs].

[118] Kent K. Chang, Mackenzie Cramer, Sandeep Soni, and David Bamman. Speak, Memory: An Archaeology of Books Known to ChatGPT/GPT-4, April 2023. arXiv:2305.00118 [cs].

[119] Tal Linzen, Emmanuel Dupoux, and Yoav Goldberg. Assessing the Ability of LSTMs to Learn Syntax-Sensitive Dependencies. *Transactions of the Association for Computational Linguistics*, 4:521–535, dec 2016.

[120] Rebecca Marvin and Tal Linzen. Targeted Syntactic Evaluation of Language Models. In *Proceedings of the 2018 Conference on Empirical Methods in Natural Language Processing*, pages 1192–1202, Brussels, Belgium, 2018. Association for Computational Linguistics.

[121] Jennifer Hu and Roger Levy. Prompt-based Methods May Underestimate Large Language Models' Linguistic Generalizations, May 2023. arXiv:2305.13264 [cs].

[122] Gasper Begus, Maksymilian Dabkowski, and Ryan Rhodes. Large Linguistic Models: Analyzing Theoretical Linguistic Abilities of LLMs, 2023. arXiv:2305.00948 [cs].

[123] Maksymilian Dabkowski and Gasper Begus. Large Language Models and (Non-)linguistic Recursion, 2023. arXiv:2306.07195 [cs].

[124] Noam Chomsky. *Lectures on Government and Binding.* Foris, 1981.

[125] James Neil Sneddon, Alexander Adelaar, Dwi Noverini Djenar, and Michael C Ewing. Indonesian Reference Grammar. *New South Wales: Allen and Unwin*, 2010.

[126] Thomas Lehmann. *A Grammar of Modern Tamil*. Pondicherry Institute of Linguistics and Culture publication. Pondicherry Institute of Linguistics and Culture, Pondicherry, second edition, 1993.

[127] Niklas Muennighoff, Thomas Wang, Lintang Sutawika, Adam Roberts, Stella Biderman, Teven Le Scao, M Saiful Bari, Sheng Shen, Zheng Xin Yong, Hailey Schoelkopf, Xiangru Tang, Dragomir Radev, Alham Fikri Aji, Khalid Almubarak, Samuel Albanie, Zaid Alyafeai, Albert Webson, Edward Raff, and Colin Raffel. Crosslingual Generalization through Multitask Finetuning. In *Proceedings of the 61st Annual Meeting of the Association for Computational Linguistics (Volume 1: Long Papers)*, pages 15991–16111, Toronto, Canada, 2022. Association for Computational Linguistics.

[128] Peter Cole, Gabriella Hermon, and Li-May Sung. Principles and Parameters of Long-distance Reflexives. *Linguistic inquiry*, pages 1–22, 1990.

[129] Peter Cole, Gabriella Hermon, and C-T James Huang. *Long Distance Reflexives*, volume 33. Brill, 2000.

[130] Elisabet Engdahl. Parasitic Gaps, Resumptive Pronouns, and Subject Extractions. *Linguistics*, 23(1):3–44, 1985.

[131] James McCloskey. Resumption, Successive Cyclicity, and the Locality of Operations. *Derivation and explanation in the minimalist program*, pages 184–226, 2002.

[132] James McCloskey. Resumption. *The Blackwell companion to syntax*, pages 94–117, 2006.

[133] Ash Asudeh. Three Kinds of Resumption, June 2007.

[134] Wynn Chao and Peter Sells. On the Interpretation of Resumptive Pronouns. In *North East Linguistics Society Vol. 13*, 1983.

[135] Ash Asudeh. Local Grammaticality in Syntactic Production. In *Language from a Cognitive Perspective Grammar, Usage, and Processing Studies in honor of Tom Wasow*. CSLI publications, 2011.

[136] Fernanda Ferreira and Benjamin Swets. The Production and Comprehension of Resumptive Pronouns in Relative Clause "Island" Contexts. *Twenty-First Century Psycholinguistics: Four Cornerstones*, January 2005.

[137] Adam Milton Morgan and Matthew W. Wagers. English Resumptive Pronouns Are More Common Where Gaps Are Less Acceptable. *Linguistic Inquiry*, 49(4):861–876, October 2018.

[138] Xiaojuan Tang, Zilong Zheng, Jiaqi Li, Fanxu Meng, Song-Chun Zhu, Yitao Liang, and Muhan Zhang. Large Language Models are In-Context Semantic Reasoners rather than Symbolic Reasoners, June 2023. arXiv:2305.14825 [cs].

[139] Ameet Deshpande, Vishvak Murahari, Tanmay Rajpurohit, Ashwin Kalyan, and Karthik Narasimhan. Toxicity in ChatGPT: Analyzing Persona-assigned Language Models, April 2023. arXiv:2304.05335 [cs].

[140] Zhenhailong Wang, Shaoguang Mao, Wenshan Wu, Tao Ge, Furu Wei, and Heng Ji. Unleashing Cognitive Synergy in Large Language Models: A Task-Solving Agent through Multi-Persona Self-Collaboration, July 2023. arXiv:2307.05300 [cs].

[141] Benfeng Xu, An Yang, Junyang Lin, Quan Wang, Chang Zhou, Yongdong Zhang, and Zhendong Mao. ExpertPrompting: Instructing Large Language Models to be Distinguished Experts, May 2023. arXiv:2305.14688 [cs].

[142] John Robert Ross. *Constraints on Variables in Syntax*. PhD thesis, Massachusetts Institute of Technology, 1967.

[143] I Wayan Arka et al. On Modality and Finiteness in Indonesian: Complexities of =nya Nominalisation. In *Workshop on TAM markers and evidentiality in Indonesian Languages, Tokyo University of Foreign Studies*, pages 17–18, 2011.

[144] Philippe Grangé. The Indonesian Verbal Suffix -nya – Nominalization or Subordination? *Wacana, Journal of the Humanities of Indonesia*, 16(1):133, April 2015.

[145] Julie Anne Legate. *Voice and v: Lessons from Acehnese*, volume 69. MIT Press, 2014.

[146] Yiyun Zhao, Jian Gang Ngui, Lucy Hall Hartley, and Steven Bethard. Do Pretrained Transformers Infer Telicity Like Humans? In *Proceedings of the 25th Conference on Computational Natural Language Learning*, pages 72–81, Online, 2021. Association for Computational Linguistics.

[147] Elay Annamalai. Modern Tamil. In Steever Sanford B, editor, *The Dravidian Languages*. Routledge, London, 2019.

[148] Sandhya Sundaresan. A Plea for Syntax and a Return to First Principles: Monstrous Agreement in Tamil. In *Semantics and Linguistic Theory*, volume 21, pages 674–693, 2011.

[149] Herbert P Grice. Logic and Conversation. In *Speech Acts*, pages 41–58. Brill, 1975.

[150] Christopher Potts. Presupposition and Implicature. *The Handbook of Contemporary Semantic Theory*, pages 168–202, 2015.

[151] Kai von Fintel. Would You Believe It? The King of France Is Back! (Presuppositions and Truth-value Intuitions). In *Descriptions and Beyond*. Oxford, New York, 2004.

[152] Mandy Simons, Judith Tonhauser, David Beaver, and Craige Roberts. What Projects and Why. In *Semantics and Linguistic Theory*, volume 20, pages 309–327, 2010.

[153] Yosuke Sato. Action/Result in Indonesian Accomplishment Verbs and the Agent Control Hypothesis. *Oceanic Linguistics*, 60(2):263–301, 2021.

[154] Tarek Naous, Michael J. Ryan, and Wei Xu. Having Beer after Prayer? Measuring Cultural Bias in Large Language Models, May 2023. arXiv:2305.14456 [cs].

[155] Krithika Ramesh, Sunayana Sitaram, and Monojit Choudhury. Fairness in Language Models Beyond English: Gaps and Challenges. In *Findings of the Association for Computational Linguistics: EACL 2023*, pages 2106–2119, Dubrovnik, Croatia, May 2023. Association for Computational Linguistics.

[156] Aida Ramezani and Yang Xu. Knowledge of Cultural Moral Norms in Large Language Models. In *Proceedings of the 61st Annual Meeting of the Association for Computational Linguistics (Volume 1: Long Papers)*, pages 428–446, Toronto, Canada, 2023. Association for Computational Linguistics.

[157] Brian Jude. *The Best of Tenali Raman*. Spider Books, 2005.

[158] Brian Jude. *Famous Stories of Tenali Raman*. Spider Books, 2007.

[159] Brian Jude. *Favourite Stories of Tenali Raman*. Spider Books, 2008.

[160] Brian Jude. *Popular Stories of Tenali Raman*. Spider Books, 2008.

[161] Brian Jude. *Selected Stories of Tenali Raman*. Spider Books, 2008.

[162] Fitri Haryani Nasution. *70 Tradisi Unik Suku Bangsa di Indonesia*. Bhuana Ilmu Populer, Jakarta, 2019.

[163] Rohan Anil, Andrew M. Dai, Orhan Firat, Melvin Johnson, Dmitry Lepikhin, Alexandre Passos, Siamak Shakeri, Emanuel Taropa, Paige Bailey, Zhifeng Chen, Eric Chu, Jonathan H. Clark, Laurent El Shafey, Yanping Huang, Kathy Meier-Hellstern, Gaurav Mishra, Erica Moreira, Mark Omernick, Kevin Robinson, Sebastian Ruder, Yi Tay, Kefan Xiao, Yuanzhong Xu, Yujing Zhang, Gustavo Hernandez Abrego, Junwhan Ahn, Jacob Austin, Paul Barham, Jan Botha, James Bradbury, Siddhartha Brahma, Kevin Brooks, Michele Catasta, Yong Cheng, Colin Cherry, Christopher A. Choquette-Choo, Aakanksha Chowdhery, Clément Crepy, Shachi Dave, Mostafa Dehghani, Sunipa Dev, Jacob Devlin, Mark Díaz, Nan Du, Ethan Dyer, Vlad Feinberg, Fangxiaoyu Feng, Vlad Fienber, Markus Freitag, Xavier Garcia, Sebastian Gehrmann, Lucas Gonzalez, Guy Gur-Ari, Steven Hand, Hadi Hashemi, Le Hou, Joshua Howland, Andrea Hu, Jeffrey Hui, Jeremy Hurwitz, Michael Isard, Abe Ittycheriah, Matthew Jagielski, Wenhao Jia, Kathleen Kenealy, Maxim Krikun, Sneha Kudugunta, Chang Lan, Katherine Lee, Benjamin Lee, Eric Li, Music Li, Wei Li, YaGuang Li, Jian Li, Hyeontaek Lim, Hanzhao Lin, Zhongtao Liu, Frederick Liu, Marcello Maggioni, Aroma Mahendru, Joshua Maynez, Vedant Misra, Maysam Moussalem, Zachary Nado, John Nham, Eric Ni, Andrew Nystrom, Alicia Parrish, Marie Pellat, Martin Polacek, Alex Polozov, Reiner Pope, Siyuan Qiao, Emily Reif, Bryan Richter, Parker Riley, Alex Castro Ros, Aurko Roy, Brennan Saeta, Rajkumar Samuel, Renee Shelby, Ambrose Slone, Daniel Smilkov, David R. So, Daniel Sohn, Simon Tokumine, Dasha Valter, Vijay Vasudevan, Kiran Vodrahalli, Xuezhi Wang, Pidong Wang, Zirui Wang, Tao Wang, John Wieting, Yuhuai Wu, Kelvin Xu, Yunhan Xu, Linting Xue, Pengcheng Yin, Jiahui Yu, Qiao Zhang, Steven Zheng, Ce Zheng, Weikang Zhou, Denny Zhou, Slav Petrov, and Yonghui Wu. PaLM 2 Technical Report, 2023. arXiv:2305.10403 [cs].

[164] Catherine Rose Fortin. We Need LF Copying: A Few Good Reasons Why. In *Proceedings of the 28th West Coast Conference on Formal Linguistics*, pages 87–95, 2011.

[165] Yosuke Sato. P-stranding under Sluicing and Repair By Ellipsis: Why Is Indonesian (Not) Special? *Journal of East Asian Linguistics*, 20(4):339–382, November 2011.

[166] Yosuke Sato. An In-situ Syntax of Sluicing in Indonesian. In *The proceedings of AFLA*, volume 23, pages 243–57, 2016.

[167] Helen Jeoung. WH-agreement Across Three Domains in Indonesian. *University of Pennsylvania Working Papers in Linguistics*, 26(1):14, 2020.

[168] Helen Jeoung. Categorial Ambiguity in *mau*, *suka*, and Other Indonesian Predicates. *Language*, 96(3):157–172, 2020.

[169] Thomas Lehmann. Old Tamil. In *The Dravidian Languages*, pages 81–103. Routledge, 2019.

[170] Harold F Schiffman. The Tamil Case System. *South Indian Horizons: Felicitation Volume for Francois Gros on the Occasion of his 70th Birthday*, pages 293–322, 2004.

[171] Elai Annamalai. The Variable Relation of Verbs in Sequence in Tamil. In *NINJAL International Symposium*, 2013.

[172] Tommi Leung. The Syntax of Two Types of Sluicing in Tamil. *The Linguistic Review*, 35(1):35–82, January 2018.

[173] I Wayan Arka, Christopher Manning, et al. *Voice and Grammatical Relations in Indonesian: A New Perspective.* CSLI Stanford, 1998.

[174] Peter Cole, Gabriella Hermon, and Yanti. Voice in Malay/Indonesian. *Lingua*, 118(10):1500–1553, October 2008.

[175] Bambang Purwo. Voice in Indonesian: A Discourse Study. *Passive and voice*, 16:195, 1988.

[176] Elai Annamalai. The Constituent Structure of Tamil. *The Yearbook of South Asian Languages and Linguistics*, pages 3–46, 2003.

[177] Benjamin Wai Hoe Pong. The Syntax of Tamil Periphrastic Causatives: A Morphosemantic Exponence. Bachelor's thesis, The National University of Singapore, 2022.

[178] Catherine Rose Fortin. On the Left Periphery in Indonesian. In *Proceedings of the Sixteenth Meeting of the Austronesian Formal Linguistics Association (AFLA), ed. Sandy Chung, Daniel Finer, Ileana Paul, and Eric Postdam*, pages 29–43. University of California, Santa Cruz, 2009.

[179] Catherine Rose Fortin. Indonesian. In van Craenenbroeck Jeroen and Temmerman Tanja, editors, *The Oxford Handbook of Elipsis.* Oxford University Press, New York, 2019.

[180] Peter Cole and Gabriella Hermon. Subject and Non-subject Relativisation in Indonesian. *Journal of East Asian Linguistics*, 14:59–88, January 2005.

[181] Harold F Schiffman. *A Reference Grammar of Spoken Tamil.* Cambridge University Press, 1999.

[182] Marwati Djoened Poesponegoro and Nugroho Notosusanto. *Sejarah Nasional Indonesia*, volume 2. Balai Pustaka, Jakarta, Indonesia, 2019.

## A    Author contributions

This project was a concerted team effort, and could not have come together without the invaluable contributions from everyone involved. The below is to recognize each contributing author's significance in this project.

**Wei Qi Leong**: Led the overall project and implemented the codebase for testing the models. Designed the NLP benchmark component of BHASA and analyzed the data and results for it. Contributed to the quality control of LINDSEA and the analysis of results.

**Jian Gang Ngui**: Led, managed and designed LINDSEA, cultural representation tasks and cultural sensitivity pilot study in close collaboration with Yosephine Susanto and Hamsawardhini Rengarajan. Analyzed results of LINDSEA and cultural representation tasks.

**Yosephine Susanto**: Managed and designed LINDSEA, cultural representation tasks and cultural sensitivity pilot study in close collaboration with Jian Gang Ngui. Performed analysis on the results of LINDSEA, cultural representation and cultural sensitivity tasks. Contributed to the analysis of BHASA Indonesian NLG task results.

**Hamsawardhini Rengarajan**: Managed and designed the LINDSEA dataset, cultural representation task and cultural sensitivity task for Tamil in close collaboration with Jian Gang Ngui. Performed analysis on the results of LINDSEA, cultural representation and cultural sensitivity tasks. Contributed to the analysis of BHASA Tamil NLG task results.

**Kengatharaiyer Sarveswaran**: Provided insights for and contributed to the validation of Tamil minimal pairs for LINDSEA. Assisted with human evaluation of model outputs for Tamil NLG tasks. Advised on the project in general and reviewed the manuscript.

**William Chandra Tjhi**: Set the direction for the project. Contributed to the validation of Indonesian coreference resolution and pragmatic reasoning data in LINDSEA. Assisted with human evaluation of model outputs for Indonesian NLG tasks. Advised on the project in general and reviewed the manuscript.

## B    Prompts

| Competency | Task | Lang | Prompt |
|---|---|---|---|
| NLU | NER | en | Perform Named Entity Recognition (NER) on the following text using only the following labels: {LABELS}. Please follow the format given in the following example: Example Text: Asia is a continent. Example Answer: @@PLACE@@Asia## is a continent. Text: {TEXT} Answer: |
| | | id | Kerjakan Named Entity Recognition (NER) pada teks berikut dengan hanya menggunakan label-label yang berikut ini: TEMPAT, ORANG, ORGANISASI. Silakan ikuti format yang diberikan dalam contoh berikut: Contoh teks: Asia adalah sebuah benua. Contoh jawaban: @@TEMPAT@@Asia## adalah sebuah benua. Teks: {TEXT} Jawaban: |
| | | vi | Thực hiện Named Entity Recognition (NER) trong văn bản dưới đây bằng cách chỉ sử dụng các nhãn sau: ĐỊA ĐIỂM, NGƯỜI, TỔ CHỨC, KHÁC. Vui lòng làm theo định dạng được cho trong ví dụ sau: Văn bản mẫu: Châu Á là một lục địa. Câu trả lời mẫu: @@ĐỊA ĐIỂM@@Châu Á## là một lục địa. Văn bản: {TEXT} Câu trả lời: |

| Competency | Task | Lang | Prompt |
|---|---|---|---|
| | | th | กรุณาทำ Named Entity Recognition (NER) ในข้อความต่อไปนี้โดยใช้ป้ายต่อไปนี้เท่านั้น: สถานที่, บุคคล, องค์กร, วันที่, เวลา, สิ่งอำนวยความสะดวก, URL, เบอร์โทรศัพท์, อีเมล, รหัสไปรษณีย์, กฎหมาย, ความยาว, เปอร์เซ็นต์. โปรดทำตามรูปแบบที่กำหนดในตัวอย่างต่อไปนี้: ตัวอย่างข้อความ: เอเชีย เป็น ทวีป ตัวอย่างคำตอบ: @@สถานที่@@เอเชีย## เป็น ทวีป ข้อความ: {TEXT} คำตอบ: |
| | | ta | தரப்பட்ட உரையில் பின்வரும் சிட்டைகளை மட்டும் பயன்படுத்தி Named Entity Recognition (NER)செய்யவும்: நபர், இடம், அமைப்பு. பின்வரும் எடுத்துக்காட்டில் கொடுக்கப்பட்டுள்ள வடிவமைப்பைப் பின்பற்றவும்: மாதிரி உரை: ஆசியா ஒரு கண்டம். மாதிரிப் பதில்: @@@இடம்@@ஆசியா## ஒரு கண்டம். உரை: {TEXT} பதில்: |
| Question Answering | | en | You will be given a paragraph and a question. Answer the question by extracting the answer from the paragraph. Paragraph: {TEXT} Question: {QUESTION} Answer: |
| | | id | Anda akan diberikan sebuah paragraf dan sebuah pertanyaan. Jawablah pertanyaannya dengan mengekstrak jawaban dari paragraf tersebut. Paragraf: {TEXT} Pertanyaan: {QUESTION} Jawaban: |
| | | vi | Bạn sẽ được cho một đoạn văn và một câu hỏi. Trả lời câu hỏi bằng cách trích xuất câu trả lời từ đoạn văn. Đoạn văn: {TEXT} Câu hỏi: {QUESTION} Câu trả lời: |
| | | th | คุณจะได้รับข้อความและคำถาม กรุณาตอบคำถาม โดยแยกคำตอบจากข้อความ ข้อความ: {TEXT} คำถาม: {QUESTION} คำตอบ: |
| | | ta | உங்களுக்கு ஒரு பத்தியும் ஒரு கேள்வியும் தரப்படும். தரப்பட்ட பத்தியிலிருந்து கேள்விக்கான பதிலைக் கண்டறியவும். கேள்விக்குப் பதிலளிக்க(முடியாவிட்டால் ஒற்றை எழுத்து X இனைப் பதிலாக அளிக்கவும். பத்தி: {TEXT} கேள்வி: {QUESTION} பதில்: |

| Competency | Task | Lang | Prompt |
|---|---|---|---|
| | Sentiment Analysis | en | What is the sentiment of the following sentence? {SENTENCE}<br>Answer only with a single word: Positive/Negative/Neutral |
| | | id | Apa sentimen dari kalimat berikut ini? {SENTENCE}<br>Jawab dengan satu kata saja: Positif/Negatif/Netral |
| | | vi | Sắc thái của câu sau đây là gì? {SENTENCE}<br>Trả lời với một từ duy nhất: Tích cực/Tiêu cực/Trung lập |
| | | th | อารมณ์ความรู้สึกของข้อความต่อไปนี้เป็นอย่างไร? {SENTENCE}<br>โปรดตอบโดยใช้คำเดียวเท่านั้น: แง่บวก/แง่ลบ/เฉยๆ<br>ถ้าข้อความเป็นคำถาม โปรดตอบด้วยตัวอักษร Q. |
| | | ta | பின்வரும் வாக்கியத்தில்<br>வெளிப்படுத்தப்படும்<br>உணர்வு எது? {SENTENCE}<br>ஒரு சொல்லில் மட்டும் பதிலளிக்கவும்:<br>நேர்மறை/எதிர்மறை |
| | Toxicity Detection | en | You are a hate speech detector.<br>The label definitions are as follows:<br>Clean: No harassment.<br>Abusive: Involves harassment and even profanity,<br>but does not attack any specific object.<br>Hate: Directly harasses or abuses a specific object.<br>When given a sentence, classify it with a single word:<br>Clean/Abusive/Hate.<br>Sentence: {TEXT} |
| | | id | Anda adalah pendeteksi ujaran kebencian.<br>Definisi dari labelnya adalah sebagai berikut:<br>Bersih: Tidak ada ujaran kebencian<br>Kasar: Ada ujaran kebencian dan kata-kata kasar,<br>namun tidak menyerang pihak tertentu<br>Benci: Ada ujaran kebencian langsung atau<br>serangan terhadap pihak tertentu<br>Ketika diberikan sebuah kalimat,<br>klasifikasikan dengan satu kata:<br>Bersih/Kasar/Benci<br>Kalimat: {TEXT} |
| | | vi | Bạn là máy phát hiện phát ngôn thù ghét.<br>Các nhãn được định nghĩa như sau:<br>Sạch: Không quấy rối<br>Công kích: Bao gồm quấy rối và thậm chí chửi thể,<br>nhưng không tấn công bất kì đối tượng cụ thể nào<br>Thù ghét: Trực tiếp quấy rối hay lăng mạ một đối<br>tượng cụ thể<br>Cho câu dưới đây, hãy phân loại nó với 1 từ duy nhất:<br>Sạch/Công kích/Thù ghét<br>Câu: {TEXT} |

| Competency | Task | Lang | Prompt |
|---|---|---|---|
| | | th | คุณเป็นเครื่องตรวจจับข้อความที่เป็นพิษ<br>ข้อความเป็นพิษถ้ามันทำร้าย สร้างความเสียหาย<br>หรือสื่อเจตนาเชิงลบถึงบุคคลหรือกลุ่มบุคคล<br>คุณควรคำนึงถึงการประชดประชันด้วย<br>เมื่อได้รับข้อความ ให้ตอบด้วยตัวอักษร Y หากเป็นพิษ และ<br>N หากไม่เป็นพิษ<br>ข้อความ: {TEXT} |
| NLG | Machine Translation (EN→XXX) | en | Translate the following text into {LANGUAGE}.<br>Text: {TEXT}<br>Translation: |
| | | id | Terjemahkan teks berikut ini ke dalam Bahasa Indonesia.<br>Teks: {TEXT}<br>Terjemahan: |
| | | vi | Dịch văn bản dưới đây sang Tiếng Việt.<br>Văn bản: {TEXT}<br>Bản dịch: |
| | | th | กรุณาแปลข้อความต่อไปนี้เป็นภาษาไทย<br>ข้อความ: {TEXT}<br>คำแปล: |
| | | ta | பின்வரும் உரையைத் தமிழ் மொழிக்கு<br>மொழிபெயர்க்கவும்.<br>உரை: {TEXT}<br>மொழிபெயர்ப்பு: |
| | Machine Translation (XXX→EN) | en | Translate the following text into English.<br>Text: {TEXT}<br>Translation: |
| | | id | Terjemahkan teks berikut ini ke dalam Bahasa Inggris.<br>Teks: {TEXT}<br>Terjemahan: |
| | | vi | Dịch văn bản dưới đây sang Tiếng Anh.<br>Văn bản: {TEXT}<br>Bản dịch: |
| | | th | กรุณาแปลข้อความต่อไปนี้เป็นภาษาอังกฤษ<br>ข้อความ: {TEXT}<br>คำแปล: |
| | | ta | பின்வரும் உரையை ஆங்கில மொழிக்கு<br>மொழிபெயர்க்கவும்.<br>உரை: {TEXT}<br>மொழிபெயர்ப்பு: |
| | Abstractive Summarization | en | Article: {TEXT}<br>Summarize this {LANGUAGE} article in 1 or 2 sentences.<br>The response should be in {LANGUAGE}.<br>Summary: |

| Competency | Task | Lang | Prompt |
|---|---|---|---|
| | | id | Artikel: {TEXT}<br>Rangkumkan artikel Bahasa Indonesia ini dalam 1 atau 2 kalimat. Jawabannya harus ditulis dalam Bahasa Indonesia.<br>Rangkuman: |
| | | vi | Bài báo: {TEXT}<br>Tóm tắt bài báo Tiếng Việt trên với 1 hay 2 câu.<br>Câu trả lời nên được<br>viết bằng tiếng Việt. Bản tóm tắt: |
| | | th | บทความ: {TEXT}<br>กรุณาสรุปบทความภาษาไทยฉบับนี้ใน 1 หรือ 2 ประโยค<br>คำตอบควรเป็นภาษาไทย<br>บทสรุป: |
| | | ta | கட்டுரை: {TEXT}<br>இந்தத் தமிழ்க் கட்டுரைக்கு 1 அல்லது 2 வாக்கியங்களில் பொழிப்பு எழுதவும்.<br>பதில் தமிழ் மொழியில் இருக்கவேண்டும்.<br>கட்டுரைப் பொழிப்பு: |
| | NLI | en | You will be given two sentences, X and Y.<br>X: {PREMISE}<br>Y: {HYPOTHESIS}<br>Determine which of the following statements applies to sentences X and Y the best.<br>A: If X is true, Y must be true.<br>B: X contradicts Y.<br>C: When X is true, Y may or may not be true.<br>Answer strictly with a single letter A, B or C. |
| | | id | Anda akan diberikan dua kalimat, X dan Y.<br>X: {PREMISE}<br>Y: {HYPOTHESIS}<br>Tentukan mana dari pernyataan berikut ini yang paling sesuai untuk kalimat X dan Y.<br>A: Kalau X benar, maka Y juga harus benar.<br>B: X bertentangan dengan Y.<br>C: Ketika X benar, Y mungkin benar atau mungkin tidak benar.<br>Jawablah hanya dengan menggunakan satu huruf A, B atau C. |
| | | vi | Bạn sẽ được cho hai câu, X và Y.<br>X: {PREMISE}<br>Y: {HYPOTHESIS}<br>Xác định câu nào sau đây là câu phù hợp nhất cho câu X và Y.<br>A: Nếu X đúng thì Y phải đúng.<br>B: X mâu thuẫn với Y.<br>C: Khi X đúng, Y có thể đúng hoặc không đúng.<br>Trả lời với một chữ cái duy nhất A, B, hoặc C. |

| Competency | Task | Lang | Prompt |
|---|---|---|---|
| | | th | คุณจะได้รับสองข้อความ X และ Y<br>X: {PREMISE}<br>Y: {HYPOTHESIS}<br>กรุณาพิจารณาว่า ข้อความใดต่อไปนี้ใช้กับข้อความ X และ Y ได้ดีที่สุด<br>A: ถ้า X เป็นจริง Y จะต้องเป็นจริง<br>B: X ขัดแย้งกับ Y<br>C: เมื่อ X เป็นจริง Y อาจเป็นจริงหรือไม่ก็ได้<br>กรุณาตอบด้วยตัวอักษร A, B หรือ C ตัวเดียวเท่านั้น |
| | | ta | உங்களுக்கு இரண்டு வாக்கியங்கள்,<br>X மற்றும் Y, தரப்படும்.<br>X: {PREMISE}<br>Y: {HYPOTHESIS}<br>பின்வரும் கூற்றுகளில் எது X மற்றும் Y<br>வாக்கியங்களுடன்<br>மிகப் பொருந்துகிறது எனக் கண்டறியவும்.<br>A: X உண்மை என்றால் Y உம் உண்மையாக<br>இருக்க வேண்டும்.<br>B: X உம் Y உம் முரண்படுகின்றன.<br>C: X உண்மையாக இருக்கும்போது Y<br>உண்மையாக இருக்கலாம்<br>அல்லது இல்லாமல் இருக்கலாம்.<br>A அல்லது B அல்லது C<br>எழுத்தில் மட்டும் பதிலளிக்கவும். |
| Causal Reasoning | | en | Situation: {PREMISE}<br>Given this situation, which of the following choices is most likely to be its {QUESTION}?<br>A: {CHOICE 1}<br>B: {CHOICE 2}<br>Respond strictly with the letters A or B only.<br>{QUESTION} = cause/effect |
| | | id | Situasi: {PREMISE}<br>Berdasarkan situasi di atas, mana dari pilihan-pilihan berikut ini yang lebih mungkin menjadi {QUESTION}?<br>A: {CHOICE 1}<br>B: {CHOICE 2}<br>Jawab dengan hanya menggunakan A atau B.<br>{QUESTION} = sebab/akibat |
| | | vi | Tình huống: {PREMISE}<br>Với tình huống trên, lựa chọn nào dưới đây có khả năng cao là {QUESTION} của nó hơn?<br>A: {CHOICE 1}<br>B: {CHOICE 2}<br>Chỉ trả lời bằng chữ cái A hoặc B.<br>{QUESTION} = nguyên nhân/kết quả |
| | | th | สถานการณ์: {PREMISE}<br>เมื่อพิจารณาจากสถานการณ์นี้ ตัวเลือกใดต่อไปนี้น่าจะเป็น {QUESTION}มากกว่ากัน?<br>A: {CHOICE1}<br>B: {CHOICE2}<br>กรุณาตอบด้วยตัวอักษร A หรือ B เท่านั้น<br>{QUESTION}=สาเหตุ/ผล |

| Competency | Task | Lang | Prompt |
|---|---|---|---|
| | | ta | சூழ்நிலை: {PREMISE}<br>பின்வரும் வாக்கியங்களில் பெரும்பாலும் எது தரப்பட்ட சூழ்நிலைக்குரிய {QUESTION} இருக்கும்?<br>A: {CHOICE 1}<br>B: {CHOICE 2}<br>A அல்லது B எழுத்தில் மட்டும் பதிலளிக்கவும்.<br>{QUESTION} =காரணமாக/விளைவாக |

Table 46: Prompts in English, Indonesian, Vietnamese, Thai and Tamil for the NLP benchmark component of BHASA

| Task | Lang | Prompt |
|---|---|---|
| Minimal Pair | en | System Prompt:<br>You are a {LANGUAGE} linguist<br>Human Prompt:<br>Which sentence is more acceptable?<br>A: {SENTENCE 1}<br>B: {SENTENCE 2}<br>Answer with A or B only. |
| | id | System Prompt:<br>Anda adalah seorang ahli bahasa Indonesia<br>Human Prompt:<br>Kalimat mana yang lebih mungkin?<br>A: {SENTENCE 1}<br>B: {SENTENCE 2}<br>Jawablah dengan menggunakan A atau B saja. |
| | ta | System Prompt:<br>நீங்கள் ஒரு தமிழ் மொழி வல்லுநர்<br>Human Prompt:<br>எந்த வாக்கியம் பெரும்பாலும் சரியாக இருக்கும்?<br>A: {SENTENCE 1}<br>B: {SENTENCE 2}<br>A அல்லது B எழுத்தில் மட்டும் பதிலளிக்கவும். |
| Translation | en | Translate the following sentence into {LANGUAGE}.<br>Sentence: {SENTENCE}<br>Translation: |
| | id | Terjemahkan kalimat berikut ini ke dalam bahasa Indonesia.<br>Kalimat: {SENTENCE}<br>Terjemahan: |
| | ta | பின்வரும் வாக்கியத்தை ஆங்கில மொழிக்கு மொழிபெயர்க்கவும்.<br>வாக்கியம்: {SENTENCE}<br>மொழிபெயர்ப்பு: |
| Coreference (Open-ended) | en | System Prompt:<br>You are a {LANGUAGE} linguist<br>Human Prompt:<br>Based on the sentence '{SENTENCE}', {QUESTION}? |

| Task | Lang | Prompt |
|---|---|---|
| | id | System Prompt:<br>Anda adalah seorang ahli bahasa Indonesia<br>Human Prompt:<br>Berdasarkan kalimat '{SENTENCE}', {QUESTION}? |
| | ta | System Prompt:<br>நீங்கள் ஒரு தமிழ் மொழி வல்லுநர்<br>Human Prompt:<br>'{SENTENCE}' என்னும் வாக்கியத்தில், {QUESTION}? |
| Coreference<br>(Binary Choice) | en | System Prompt:<br>You are a {LANGUAGE} linguist<br>Human Prompt:<br>{SENTENCE}<br>Which is more appropriate to fill in the blank?<br>A: {ENTITY 1}<br>B: {ENTITY 2}<br>Answer only with A or B. |
| | id | System Prompt:<br>Anda adalah seorang ahli bahasa Indonesia<br>Human Prompt:<br>{SENTENCE}<br>Manakah yang lebih tepat untuk mengisi bagian yang kosong?<br>A: {ENTITY 1}<br>B: {ENTITY 2}<br>Jawablah hanya dengan menggunakan A atau B. |
| | ta | System Prompt:<br>நீங்கள் ஒரு தமிழ் மொழி வல்லுநர்<br>Human Prompt:<br>{SENTENCE}<br>கோடிட்ட இடத்தை நிரப்ப எது<br>மிகவும் பொருத்தமானது?<br>A: {ENTITY 1}<br>B: {ENTITY 2}<br>A அல்லது B எழுத்தில் மட்டும் பதிலளிக்கவும். |
| Coreference (Indonesian<br>Pronoun Resolution) | en | System Prompt:<br>You are an Indonesian linguist<br>Human Prompt:<br>In the following sentences, who or what does 'dia' refer to?<br>{SENTENCE} |
| | id | System Prompt:<br>Anda adalah seorang ahli bahasa Indonesia<br>Human Prompt:<br>Dalam kalimat berikut ini, 'dia' mengacu pada apa atau siapa?<br>{SENTENCE} |
| Pragmatic Reasoning<br>(Single Sentence) | en | System Prompt:<br>You are a {LANGUAGE} linguist<br>Human Prompt:<br>Is the following statement true or false?<br>Statement: {SENTENCE}<br>Answer only with True or False. |

| Task | Lang | Prompt |
|------|------|--------|
| | id | System Prompt:<br>Anda adalah seorang ahli bahasa Indonesia<br>Human Prompt:<br>Apakah pernyataan berikut ini benar atau salah?<br>Pernyataan: {SENTENCE}<br>Jawablah hanya dengan Benar atau Salah. |
| | ta | System Prompt:<br>நீங்கள் ஒரு தமிழ் மொழி வல்லுநர்<br>Human Prompt:<br>பின்வரும் கூற்று உண்மையா அல்லது பொய்யா?<br>கூற்று: {SENTENCE}<br>உண்மை அல்லது பொய் என்ற வார்த்தையில் மட்டும் பதிலளிக்கவும். |
| Pragmatic Reasoning (Sentence Pair) | en | System Prompt:<br>You are a {LANGUAGE} linguist<br>Human Prompt:<br>Situation: {PREMISE}<br>Given this situation, is the following statement true or false?<br>Statement: {HYPOTHESIS}<br>Answer only with True or False. |
| | id | System Prompt:<br>Anda adalah seorang ahli bahasa Indonesia<br>Human Prompt:<br>Situasi: {PREMISE}<br>Berdasarkan situasi ini, apakah pernyataan berikut ini benar atau salah?<br>Pernyataan: {HYPOTHESIS}<br>Jawablah hanya dengan Benar atau Salah. |
| | ta | System Prompt:<br>நீங்கள் ஒரு தமிழ் மொழி வல்லுநர்<br>Human Prompt:<br>சூழ்நிலை: {PREMISE}<br>இந்த சூழ்நிலையில்,<br>பின்வரும் கூற்று உண்மையா அல்லது பொய்யா?<br>கூற்று: {HYPOTHESIS}<br>உண்மை அல்லது பொய் என்ற வார்த்தையில் மட்டும் பதிலளிக்கவும். |

Table 47: Prompts in English, Indonesian, and Tamil for LINDSEA

## C   Data quality issues

| Lang | Premise | Question | Answer | Correct Question |
|---|---|---|---|---|
| id | Tinta di poster itu luntur. <br> The ink on the poster smeared. | Effect | Saya mencipratkan air ke poster. <br> I spilled water on the poster. | Cause |
| | Keluarga itu mencari tetangga-tetangganya.* <br> Should be "Keluarga itu mencari sekitar lingkungan sekitar rumah." <br><br> The family searched the neighborhood. | Effect | Anjing mereka kabur dari rumah. <br> The dog ran away from home. | Cause |
| | Ovennya memanas. <br> The oven became hot. | Effect | Saya menyalakan oven tersebut. <br> I turned the oven on. | Cause |
| | Perempuan itu tersandung* di trotoar. <br> Should be tersandung <br><br> The woman tripped on the sidewalk. | Effect | Ada retakan di lantai trotoar. <br> There was a crack in the cement. | Cause |
| th | ที่ทำงานปิด <br> The office was closed. | Effect | มันคือวันหยุด <br> It was a holiday. | Cause |
| | ผู้หญิงเสียที่ในคิว <br> The woman lost her place in line. | Effect | เขาเดินออกนอกคิวไป <br> She stepped out of line. | Cause |
| | ผู้ชายถูกปฏิเสธสินเชื่อ <br> The man was denied for a loan. | Effect | เขามีหนี้สิน <br> He was in debt. | Cause |
| | ฉันบ้วนนมออกมา <br> I spit out the milk. | Effect | นมมีรสเปรี้ยว <br> The milk tasted sour. | Cause |

Table 48: Examples of errors in the question label in XCOPA. The premises and hypotheses are taken directly from the XCOPA dataset and mistakes in translation from the COPA dataset are also pointed out using the * symbol, with the suggested translation given after the erroneous one.

## D  NLG Human Evaluation guidelines

As mentioned in Section 3.5.2, we perform human evaluation for machine translation and abstractive summarization for a small sample of sentences and articles. The error typology for machine translation (Table 49) and the definitions of the dimensions on which to score summarizations (Table 50) can be found below.

| Category | Subcategory | Explanation |
|---|---|---|
| Accuracy | Addition | Translated content that includes content not present in the source. |
| | Omission | Errors where content is missing from the translation but is present in the source. |
| | Over-translation | Translated content that is inappropriately more specific than the source text. E.g. The source refers to a boy, but the translation uses a term that refers only to young boys. |
| | Under-translation | Translated content that is inappropriately less specific than the source. E.g. A specific type of military officer (such as an admiral) is referred to in the source, but the translation uses a term that refers to military officers in general. |
| | Mistranslation | Target content that does not accurately represent the source content. E.g. A source text states that a medicine should not be administered in doses greater than 200mg. However, the translation says that it should be administered in doses greater than 200mg. (i.e. the opposite meaning is achieved due to not translating the negation) |
| | Untranslated | Text segments that were not translated. E.g. The entire sentence or a part of the sentence is somehow left in the source language without being translated. |
| Terminology | Inconsistent Terminology | Use of multiple terms for the same concept when the same word should be used for consistency E.g. A text referring to a particular lever as "brake release lever" and "brake disengagement lever" at the same time. |
| | Inappropriate Terminology | Use of a term that is inappropriate in the context of the text. E.g. "The Seine river empties into the English Channel" In this sentence, it is inappropriate to translate the word river as rivière in French because a rivière empties into another river, whereas a fleuve empties into a sea or ocean. |
| Linguistic Conventions | Grammar | Translated text contains grammatical errors. E.g. "The man was seeing the his wife." |
| | Punctuation | Translated text contains punctuation errors. |
| | Spelling | Translated text contains spelling errors. E.g. "Despicable" spelled as "Depiscable". In Vietnamese and Thai this could also involve a wrong tone marker. |
| | Unintelligible | Translated text (partially or as a whole) is incomprehensible. E.g. "The brake from whe this ਇੰਸਤੋ ਹੈਸ S149235 part numbr„." |

| Category | Subcategory | Explanation |
|---|---|---|
| Style | Naturalness | Style that is grammatical but unnatural, usually due to interference from the source language. E.g. "We thanked him with heart" is understandable and grammatically acceptable, but is a direct translation from German and would be better translated as "We thanked him heartily". |
| | Register | Register of source and translated text do not match. E.g. Source uses formal language but translation uses informal language. |
| | Inconsistent Style | The translated text itself is inconsistent in style. E.g. One part of a text is written in a concise, straight-to-the-point manner, while another is written in a flowery, more wordy style. |
| Locale Conventions | Format | Errors where the translation violates locale-specific format for things such as numbers, currencies, measurements, time and date. E.g. Indonesian and Vietnamese use commas to separate the decimals and periods to separate thousands (1.000.000,00 instead of 1,000,000.00). US English expresses dates as mm/dd/yyyy while most other places use dd/mm/yyyy instead, and Thai uses Buddhist Era. |

Table 49: Error Typology for Human Evaluation of Machine Translation

| Criterion | Score range | Explanation |
|---|---|---|
| Relevance | 1/2/3/4/5 | This rating measures how well the summary captures the key points of the article. Consider whether all and only the important aspects are contained in the summary. Redundancies and excess or unimportant information should be penalized. |
| Faithfulness | 0/1 | This rating measures whether the facts in the summary are consistent with the facts in the original article (i.e. whether the summary is faithful to the article). Consider whether the summary produces all facts accurately and does not make up untrue information. Hallucinated facts should be penalized. |
| Fluency | 1/2/3/4/5 | This rating measures the quality of individual sentences in terms of whether they are well-written and grammatically correct. Formatting problems, capitalisation errors and ungrammatical sentences should be penalized. |
| Coherence | 1/2/3/4/5 | This rating measures the quality of the summary as a whole in terms of whether it is well-structured and well-organized. The summary should not just be a heap of related information, but should build from sentence to sentence to a coherent body of information about a topic. |

Table 50: Explanations for the 4 rating criteria for summarization

# E   LINDSEA Linguistic phenomena categories

| Linguistic Phenomena | Category | Subcategory |
|---|---|---|
| Morphology | *Men-* with clefted sentences | - |
| | *Men-* with control verbs *mau* and *coba* | *Coba* |
| | | *Mau* |
| | *Men-* with *dirinya/dirimu* subjects | *Dirimu* |
| | | *Dirinya* |
| | *Men-* with objects subject to Left Branch Condition | WH-in-situ |
| | | WH-movement from island |
| | Adjective Reduplication | - |
| | Noun Reduplication (Endocentric Compounds) | - |
| | Noun Reduplication (Plurals) | - |
| | Benefactive constructions without *untuk* | - |
| | Dative constructions without *kepada* | - |
| | Preposition stranding | - |
| | Correlative conjunctions | *Biarpun + namun* |
| | | *Meskipun + namun* |
| | | *Meskipun + tetapi* |
| | | *Walaupun + tapi* |
| | Complementizers (*bahwa/kalau*) | Ungrammatical with *bahwa* |
| | | Ungrammatical with *kalau* |
| | *-i/-kan suffix* | Grammatical with *-i* |
| | | Grammatical with *-kan* |
| | Verb + *-kah* | - |
| Argument Structure | *Dirimu sendiri* with *dia* + *Dirinya sendiri* with *kamu* | *Dirimu sendiri + dia* |
| | | *Dirinya sendiri + kamu* |
| | *Dirimu sendiri* with/without *kamu* + *Dirinya sendiri* with/without *dia* | *Dirimu sendiri* |
| | | *Dirinya sendiri* |
| | *Dirimu* with *dia* + *Dirinya* with *kamu* | *Dirimu + dia* |
| | | *Dirinya + kamu* |
| | *Dirimu* with/without *kamu* + *Dirinya* with/without *dia* | *Dirimu* without *kamu* |
| | | *Dirinya* without *dia* |
| | Modals in Active Voice | - |
| | Modals in Object Voice | With *-kan* on verb |
| | | Without *-kan* on verb |
| | Temporal nouns + *-lah* | - |
| | *Di*-VP topicalization with *=nya* | - |
| | *Ada* | With *=nya* |
| | | With *itu* |
| | Asymmetry between animates and inanimates in passive voice | - |
| | No *oleh* in passive voice | - |
| | *Oleh*-stranding in passive voice | No additional argument |
| | | With additional argument |
| | Demonstratives on objects in object voice | - |
| | Possessive *=nya* | - |
| | Word order in benefactive constructions | - |
| | Word order in object voice | With *-kan* on verb |
| | | Without *-kan* on verb |

| Linguistic Phenomena | Category | Subcategory |
|---|---|---|
| Filler-gap Dependencies | Verb phrase ellipsis | Aspect mismatch |
| | | Word order |
| | WH-movement with Left Branch Condition | - |
| | WH-movement with adjunct and coordinate islands | Adjunct islands |
| | | Coordinate islands |
| | Object relatives and demonstratives | - |
| | Subject relatives and demonstratives | - |
| | Topicalization with Left Branch Condition | - |
| NPIs and Negation | NPI licensing in-situ | - |
| | NPI licensing with movement | - |

Table 51: Categories and subcategories of linguistic phenomena in the Indonesian dataset of LINDSEA

| Linguistic Phenomena | Category | Subcategory |
|---|---|---|
| Morphology | Nominative subject | - |
| | Obligatory accusative case marking | Demonstrative |
| | | Proper Noun |
| | *Enpatu* with case | Accusative case |
| | | Instrumental case |
| | | Sociative case |
| | *Enru* | - |
| | Matrix verbs selecting for finite clauses | - |
| | Matrix verbs selecting for non-finite clauses | - |
| | Verbal reduplication with numerals | One (*oru*) |
| | | Two (*irantu*) |
| | Agreement with active subject vs. object | - |
| | Agreement with passive subject vs. agent | - |
| | Agreement with dative vs. nominative subjects | - |
| | Agreement with nominative vs. dative subjects | - |
| | Gender mismatch | - |
| | Person mismatch | - |
| | Tense and temporal mismatch | - |
| Argument Structure | Complex predicate ordering | - |
| | *Vai* instrumental ordering | - |
| | Adding accusative case to passive subject | - |
| | Scrambling out of embedded clauses | Animate object |
| | | Inanimate object |
| Filler-gap Dependencies | Subject relatives | - |
| | Object relatives | - |

Table 52: Categories and subcategories of linguistic phenomena in the Tamil dataset of LINDSEA

|  | Indonesian References | Tamil References |
|---|---|---|
| Morphology | [125]; [164]; [165]; [166]; [167]; [168] | [126];[169]; [170]; [171]; [147]; [172] |
| Argument Structure | [173]; [174]; [175]; [145]; [125] | [176];[177]; [126]; |
| Filler-gap Dependencies | [178];[179]; [125] | [126]; [172]; [177] |
| NPIs and Negation | [125] | [126] |
| Translation | [173]; [175]; [125] | [126] [170] |
| Coreference Resolution | [173]; [180] [125] | [126]; [181]; [148]; [147] |
| Scalar Implicatures | [125] | [126] |

Table 53: References consulted in the design of LINDSEA

# F    LINDSEA detailed results

|  | Run | Accuracy | Answer Frequency | | |
|---|---|---|---|---|---|
|  |  |  | A | B | Others |
| *Indonesian Dataset* | | | | | |
| GPT-3.5-Turbo | Random | 40.00 | 31.84 | 39.21 | 28.95 |
|  | All A | 40.79 | 40.79 | 33.68 | 25.53 |
|  | All B | 48.68 | 27.63 | 48.68 | 23.69 |
| GPT-3.5-Turbo (T) | Random | 52.36 | 33.95 | **66.05** | 0.00 |
|  | All A | 40.53 | 40.53 | 58.95 | 0.52 |
|  | All B | 69.74 | 30.26 | **69.74** | 0.00 |
| GPT-4 | Random | 70.79 | 37.63 | 62.37 | 0.00 |
|  | All A | 60.79 | 60.79 | 39.21 | 0.00 |
|  | All B | 82.11 | 17.89 | **82.11** | 0.00 |
| GPT-4 (T) | Random | 69.21 | 31.84 | **68.16** | 0.00 |
|  | All A | 52.63 | 52.63 | 47.37 | 0.00 |
|  | All B | 87.89 | 12.11 | **87.89** | 0.00 |
| *Tamil Dataset* | | | | | |
| GPT-3.5-Turbo | Random | 48.50 | 28.00 | 52.50 | 19.50 |
|  | All A | 35.00 | 35.00 | 43.00 | 22.00 |
|  | All B | 60.50 | 20.00 | **60.50** | 19.50 |
| GPT-3.5-Turbo (T) | Random | 39.50 | 20.50 | **45.50** | 34.00 |
|  | All A | 21.00 | 21.00 | **42.50** | 36.50 |
|  | All B | 48.50 | 11.50 | **48.50** | 40.00 |
| GPT-4 | Random | 73.50 | 24.50 | **75.50** | 0.00 |
|  | All A | 47.00 | 47.00 | 53.00 | 0.00 |
|  | All B | 96.50 | 3.50 | **96.50** | 0.00 |
| GPT-4 (T) | Random | 70.50 | 20.50 | **79.50** | 0.00 |
|  | All A | 34.00 | 34.00 | 66.00 | 0.00 |
|  | All B | 97.50 | 2.50 | **97.50** | 0.00 |

Table 54: Answer frequencies for the minimal pair task across 3 runs. GPT-4 seems to display a form of answer bias, preferring to answer B on most questions. Instances where the rate of choosing B is at least twice that of choosing A have been highlighted.

| Linguistic Phenomenon | Category | GPT-3.5-Turbo | GPT-3.5-Turbo (T) | GPT-4 | GPT-4 (T) |
|---|---|---|---|---|---|
| Morphology | *-i/-kan* suffix | 2 | 4 | 6 | 5 |
| | V + *-kah* | 1 | 1 | 3 | 2 |
| | Adjective reduplication | 1 | 5 | 9 | 8 |
| | Benefactive constructions without *untuk* | 0 | 0 | 4 | 3 |
| | Complementizers (*bahwa/kalau*) | 4 | 5 | 9 | 8 |
| | Dative constructions without *kepada* | 0 | 3 | 4 | 6 |
| | *Men-* with clefted sentences | 1 | 1 | 5 | 7 |
| | *Men-* with control verbs *mau* and *coba* | 0 | 0 | 0 | 0 |
| | *Men-* with *dirinya/dirimu* subjects | 0 | 0 | 0 | 0 |
| | *Men-* with objects subject to Left Branch Condition | 0 | 0 | 6 | 3 |
| | Noun reduplication (endocentric compounds) | 3 | 2 | 7 | 5 |
| | Noun reduplication (plurals) | 4 | 3 | 4 | 3 |
| | Preposition stranding | 5 | 9 | 10 | 10 |
| | Correlative conjunctions | 0 | 0 | 0 | 0 |
| Argument Structure | *Ada* | 0 | 4 | 5 | 5 |
| | Asymmetry between animates and inanimates in passive voice | 2 | 6 | 8 | 6 |
| | Demonstratives on objects in object voice | 0 | 0 | 1 | 0 |
| | *Di-*VP topicalisation with *=nya* | 0 | 0 | 4 | 5 |
| | *Dirimu sendiri* with *dia + dirinya sendiri* with *kamu* | 3 | 2 | 5 | 5 |
| | *Dirimu sendiri* with/without *kamu + dirinya sendiri* with/without *dia* | 0 | 0 | 8 | 3 |
| | *Dirimu* with *dia + dirinya* with *kamu* | 3 | 3 | 1 | 2 |
| | *Dirimu* w./w.o *kamu + dirinya* w./w.o *dia* | 0 | 0 | 3 | 2 |
| | Modals in active voice | 6 | 10 | 10 | 10 |
| | Modals in object voice | 0 | 0 | 1 | 0 |
| | No *oleh* in passive voice | 0 | 2 | 5 | 1 |
| | *Oleh*-stranding in passive voice | 5 | 2 | 4 | 9 |
| | Possessive *=nya* | 0 | 0 | 0 | 0 |
| | Temporal nouns + *-lah* | 4 | 4 | 10 | 10 |
| | Word order in benefactive constructions | 4 | 9 | 10 | 10 |
| | Word order in object voice | 2 | 4 | 9 | 8 |
| Filler-gap Dependencies | Verb phrase ellipsis | 6 | 10 | 5 | 5 |
| | WH-movement with Left Branch Condition | 1 | 0 | 2 | 1 |
| | WH-movement with adjunct islands and coordinate islands | 0 | 2 | 6 | 7 |
| | Object relatives and demonstratives | 5 | 6 | 10 | 10 |
| | Subject relatives and demonstratives | 8 | 9 | 9 | 9 |
| | Topicalisation with Left Branch Condition | 1 | 0 | 3 | 1 |
| NPIs and Negation | NPI licensing in-situ | 0 | 2 | 10 | 10 |
| | NPI licensing with movement | 4 | 10 | 10 | 10 |

Table 55: Detailed breakdown of GPT-3.5-Turbo and GPT-4's performance on the Indonesian minimal pair task. Each category has a total of 10 questions.

| Linguistic Phenomenon | Category | GPT-3.5-Turbo | GPT-3.5-Turbo (T) | GPT-4 | GPT-4 (T) |
|---|---|---|---|---|---|
| Morphology | Nominative subject | 0 | 0 | 0 | 0 |
| | Obligatory accusative case marking | 0 | 0 | 7 | 0 |
| | *Enpatu* with case | 0 | 0 | 5 | 4 |
| | *Enru* | 0 | 0 | 2 | 0 |
| | Matrix verbs selecting for finite clauses | 4 | 4 | 8 | 4 |
| | Matrix verbs selecting for non-finite clauses | 2 | 1 | 3 | 0 |
| | Verbal reduplication with numerals | 7 | 1 | 2 | 4 |
| | Agreement with active subject vs. object | 0 | 0 | 5 | 3 |
| | Agreement with passive subject vs. agent | 3 | 2 | 0 | 0 |
| | Agreement with dative vs. nominative subjects | 0 | 3 | 6 | 3 |
| | Agreement with nominative vs. dative subjects | 0 | 0 | 0 | 1 |
| | Gender mismatch | 3 | 5 | 6 | 4 |
| | Person mismatch | 2 | 2 | 7 | 3 |
| | Tense and temporal mismatch | 2 | 1 | 7 | 5 |
| Argument Structure | Complex predicate ordering | 3 | 2 | 5 | 1 |
| | Vai instrumental ordering | 3 | 0 | 9 | 9 |
| | Adding accusative case to passive subject | 4 | 0 | 0 | 0 |
| | Scrambling out of embedded clauses | 0 | 0 | 0 | 3 |
| Filler-gap Dependencies | Subject relatives | 0 | 0 | 10 | 8 |
| | Object relatives | 0 | 0 | 6 | 10 |

Table 56: Detailed breakdown of GPT-3.5-Turbo and GPT-4's performance on the Tamil minimal pair task. Each category has a total of 10 questions.

| | Binary Choice | | Open-ended | | Total |
|---|---|---|---|---|---|
| | Blank-filling | Blank-filling with demonstratives | Pronoun Resolution | Pro-drop | |
| | | | Indonesian Dataset | | |
| GPT-3.5-Turbo | 8/16 | 13/28 | 10/12 | 11/16 | 42/72 |
| | 50.00% | 46.43% | 83.33% | 68.75% | 58.33% |
| GPT-3.5-Turbo (T) | 5/16 | 13/28 | 9/12 | 12/16 | 39/72 |
| | 31.25% | 46.43% | 75.00% | 75.00% | 54.17% |
| GPT-4 | **13/16** | 21/28 | **11/12** | **16/16** | **61/72** |
| | **81.25%** | 75.00% | **91.67%** | **100.00%** | **84.72%** |
| GPT-4 (T) | 12/16 | **23/28** | 10/12 | 15/16 | 60/72 |
| | 75.00% | **82.14%** | 83.33% | 93.75% | 83.33% |
| | | | Tamil Dataset | | |
| GPT-3.5-Turbo | **11/58** | - | - | - | **11/58** |
| | **18.97%** | | | | **18.97%** |
| GPT-3.5-Turbo (T) | 8/58 | - | - | - | 8/58 |
| | 13.79% | | | | 13.79% |
| GPT-4 | 9/58 | - | - | - | 9/58 |
| | 15.52% | | | | 15.52% |
| GPT-4 (T) | 8/58 | - | - | - | 8/58 |
| | 13.79% | | | | 13.79% |

Table 57: Breakdown of performance of GPT-3.5-Turbo and GPT-4 on coreference resolution requiring commonsense knowledge (Translated XWinograd Dataset)

| | | GPT-3.5-Turbo | GPT-3.5-Turbo (T) | GPT-4 | GPT-4 (T) |
|---|---|---|---|---|---|
| Indonesian | Correct | 21/44 | 18/44 | 34/44 | 35/44 |
| | | 47.72% | 40.91% | 77.27% | 79.55% |
| | Unsure | 14/44 | 16/44 | 9/44 | 8/44 |
| | | 31.82% | 36.36% | 20.45% | 18.18% |
| | Wrong | 9/44 | 10/44 | 1/44 | 1/44 |
| | | 20.45% | 22.73% | 2.27% | 2.27% |
| Tamil | Correct | 11/58 | 8/58 | 9/58 | 8/58 |
| | | 18.97% | 13.79% | 15.52% | 13.79% |
| | Unsure | 34/58 | 37/58 | <span style="color:red">**42/58**</span> | <span style="color:red">**47/58**</span> |
| | | 58.62% | 63.79% | <span style="color:red">**72.41%**</span> | <span style="color:red">**81.03%**</span> |
| | Wrong | 13/58 | 13/58 | 7/58 | 3/58 |
| | | 22.41% | 22.41% | 12.07% | 5.17% |

Table 58: Breakdown of performance of GPT-3.5-Turbo and GPT-4 on the translated XWinograd binary choice task. The task was run thrice for each language. 'Correct' and 'Wrong' refer to answers that are consistently correct or wrong across all three runs. 'Unsure' refers to inconsistent answers across the three runs. GPT-4 is highly unsure for the Tamil dataset.

| | Degree ordering | Negation | Booster | Comparative | Superlative | Some | All | Total |
|---|---|---|---|---|---|---|---|---|
| | | | | Indonesian Dataset | | | | |
| GPT-3.5-Turbo | 10/20 | 12/20 | 10/20 | **8/10** | 5/10 | 10/12 | 9/12 | 64/104 |
| | 50.00% | 60.00% | 50.00% | **80.00%** | 50.00% | 83.33% | 75.00% | 61.54% |
| GPT-3.5-Turbo (T) | 10/20 | 11/20 | 10/20 | 4/10 | 6/10 | 11/12 | **11/12** | 63/104 |
| | 50.00% | 55.00% | 50.00% | 40.00% | 60.00% | 91.67% | **91.67%** | 60.58% |
| GPT-4 | 9/20 | 10/20 | 14/20 | 2/10 | **10/10** | **12/12** | 7/12 | 64/104 |
| | 45.00% | 50.00% | 70.00% | 20.00% | **100.00%** | **100.00%** | 58.33% | 61.54% |
| GPT-4 (T) | **13/20** | **19/20** | **20/20** | 5/10 | **10/10** | **12/12** | 8/12 | **87/104** |
| | **65.00%** | **95.00%** | **100.00%** | 50.00% | **100.00%** | **100.00%** | 66.67% | **83.65%** |
| | | | | Tamil Dataset | | | | |
| GPT-3.5-Turbo | 10/20 | 9/20 | 11/20 | 0/10 | 1/10 | 7/12 | 6/12 | 44/104 |
| | 50.00% | 45.00% | 55.00% | 0.00% | 10.00% | 58.33% | 50.00% | 42.31% |
| GPT-3.5-Turbo (T) | 10/20 | 10/20 | 9/20 | **1/10** | 3/10 | 8/12 | 6/12 | 47/104 |
| | 50.00% | 50.00% | 45.00% | **10.00%** | 30.00% | 66.67% | 50.00% | 45.19% |
| GPT-4 | **12/20** | 6/20 | 10/20 | 0/10 | 3/10 | 7/12 | **9/12** | 47/104 |
| | **60.00%** | 30.00% | 50.00% | 00.00% | 30.00% | 58.33% | **75.00%** | 45.19% |
| GPT-4 (T) | 11/20 | **11/20** | 14/20 | 1/10 | 10/10 | 9/12 | 5/12 | 61/104 |
| | 55.00% | **55.00%** | 70.00% | 10.00% | 100.00% | 75.00% | 41.67% | **58.65%** |

Table 59: Breakdown of performance of GPT-3.5-Turbo and GPT-4 on scalar implicatures.

| | Degree ordering | Negation | Booster | Comparative | Superlative | Some | All |
|---|---|---|---|---|---|---|---|
| GPT-3.5-Turbo | **20/20** | **19/20** | **19/20** | **10/10** | **9/10** | 5/12 | 0/12 |
| GPT-3.5-Turbo (T) | **19/20** | **18/20** | **19/20** | 7/10 | 7/10 | 10/12 | 9/12 |
| GPT-4 | **18/20** | 16/20 | **20/20** | **10/10** | 7/10 | 3/12 | 3/12 |
| GPT-4 (T) | **1/20** | **1/20** | 6/20 | **9/10** | **0/10** | 5/12 | 3/12 |

Table 60: Models' frequency of selecting 'True' on Tamil scalar implicatures. The performance of the models for 5 out of 7 subcategories (excluding 'Some' and 'All') seems to be influenced by their tendencies to choose either only 'true' or 'false'.

|  | Change-of-state Verbs | Change-of-state Verbs (non-culminating) | Clefting | Embedded Interrogatives | Total |
|---|---|---|---|---|---|
| Indonesian Dataset | | | | | |
| GPT-3.5-Turbo | 9/20 | 12/20 | 9/20 | 14/20 | 44/80 |
|  | 45.00% | 60.00% | 45.00% | 70.00% | 55.00% |
| GPT-3.5-Turbo (T) | 14/20 | **13/20** | 10/20 | 14/20 | 51/80 |
|  | 70.00% | **65.00%** | 50.00% | 70.00% | 63.75% |
| GPT-4 | 17/20 | **13/20** | 17/20 | 17/20 | 64/80 |
|  | 85.00% | **65.00%** | 85.00% | 85.00% | 80.00% |
| GPT-4 (T) | **18/20** | 10/20 | **19/20** | **18/20** | **65/80** |
|  | **90.00%** | 50.00% | **95.00%** | **90.00%** | **81.25%** |
| Tamil Dataset | | | | | |
| GPT-3.5-Turbo | 9/20 | - | 11/20 | 13/20 | 33/60 |
|  | 45.00% | | 55.00% | 65.00% | 55.00% |
| GPT-3.5-Turbo (T) | 10/20 | - | 8/20 | 8/20 | 26/60 |
|  | 50.00% | | 40.00% | 40.00% | 43.33% |
| GPT-4 | 12/20 | - | 19/20 | 13/20 | **44/60** |
|  | 60.00% | | 95.00% | 65.00% | **73.33%** |
| GPT-4 (T) | 7/20 | - | 14/20 | 10/20 | 31/60 |
|  | 35.00% | | 70.00% | 50.00% | 51.67% |

Table 61: Breakdown of performance of GPT-3.5-Turbo and GPT-4 on presuppositions.

# G   Cultural representation

**Proverbs**
Indonesian

1. Duduk sama rendah, berdiri sama tinggi
   Lit: 'Sitting same low, standing same high'
   'We are all equals'

2. Air cucuran atap jatuhnya ke pelimbahan juga
   Lit: 'Water from the roof falls into the wastewater too'
   'The apple does not fall far from the tree'

3. Di mana bumi dipijak, di situ langit dijunjung
   Lit: 'Where the earth is trodden, there the sky is upheld'
   'When in Rome, do as the Romans do'

4. Guru kencing berdiri, murid kencing berlari
   Lit: 'The teacher urinates standing, the student urinates running'
   'The student will follow the teacher's example'

5. Menang jadi arang, kalah jadi abu
   Lit: 'Win becomes coal, lose becomes ash'
   'Losing or winning in an argument will not benefit either party'

Tamil

1. அளவுக்கு மிஞ்சினால் அமிர்தழுextம் நஞ்சு
   Lit: 'Even nectar beyond the limit can be poisonous'
   'Too much of anything is good for nothing'

2. உதட்டில் உறவு உள்ளத்தில் பகை
   Lit: 'Relationship on the lips, enmity in the heart'
   'People might talk sweetly but harbor ill will in their hearts'

3. எரிகிற விளக்கானாலும் தூண்டுகோல் வேண்டும்
   Lit: 'Even a burning lamp needs a catalyst'
   'Even if one does well, they still need encouragement'

    4. ஆழம் அறியாமல் காலை இடாதே
       Lit: 'Without knowing the depth, do not dip your legs'
       'Look before you leap'

    5. ஆத்திரக்காரனுக்குப் புத்தி மட்டு
       Lit: 'An angry man has less than average intelligence'
       'An angry man does not make sound decisions'

**Literature**

Values for Indonesian folk tales

1. *Bhinneka Tunggal Ika* (unity in diversity)
2. *Gotong Royong* (working together as a community)
3. *Lapang dada* (accepting whatever comes happily)
4. *Rendah hati* (being humble)
5. *Tenggang rasa* (appreciating and respecting other people's feelings)

Values for Tamil *Tenali Raman* fables

1. Generosity
2. Humility
3. Respecting elders
4. Honesty
5. Being willing to help others

**History**

Indonesian historical figures [182]

1. Anusapati
2. Girindrawardhana
3. Ken Arok
4. Ratu Sanjaya
5. Tribhuwana Wijayatunggadewi

Tamil historical figures

1. அதியமான்(Adhiyamān)[33]
2. பாரி (Pāri)[34]
3. ஒளவையார் (Avvaiyār)[35]
4. திருவள்ளுவர் (Thiruvalluvar)[36]
5. சுப்பிரமணிய பாரதி (Subramania Bharathi)[37]

**Customs**

Indonesian customs [162]

1. *Batombe* (Sumatra)
2. *Brobosan* (Java)
3. *Ma'nene* (Sulawesi)
4. *Mekare-kare* (Bali)
5. *Penamou* (Maluku)
6. *Tatung* (Singkawang)

---

[33]https://en.m.wikipedia.org/wiki/Athiyam%C4%81n
[34]https://en.m.wikipedia.org/wiki/V%C4%93l_P%C4%81ri
[35]https://en.m.wikipedia.org/wiki/Avvaiyar_(Sangam_poet)
[36]https://en.m.wikipedia.org/wiki/Thiruvalluvar
[37]https://en.m.wikipedia.org/wiki/Subramania_Bharati

7. *Bau Nyale* (Nusa Tenggara Barat)
8. *Bakar Batu* (Papua)
9. *Pasola* (Sumba)
10. *Tiwah* (Kalimantan)

Indonesian traditional games

1. *Egrang*
2. *Hompimpa*
3. *Lompat Tali Karet*
4. *Patok Lele*
5. *Suit Indonesia*

Tamil customs

1. கோலம் (*Kolam*)
2. அறுசுவை உணவு (*Arusuvai Unavu*)
3. சிலம்பம் (*Silambam*)
4. ஜல்லிக்கட்டு (*Jallikattu*)
5. பரதநாட்டியம் (*Bharatanatyam*)
6. விருந்தோம்பல் (*Virunthombal*)
7. ஆயுர்வேதம் (*Ayurvetam*)

Tamil traditional games

1. கபடி (*Kabadi*)
2. பல்லாங்குழி (*Pallanguzhi*)
3. ஆடு புலி ஆட்டம் (*Aadu Puli Aatam*)
4. தாயக்கட்டை (*Dayakattai*)
5. அஞ்சாங்கல் (*Anjaankal*)

# H  Cultural sensitivity

In this section, we only provide a truncated response due to the potential sensitivity of its contents.

| System | *Anda adalah orang Indonesia yang tinggal di Indonesia*<br>You are an Indonesian living in Indonesia |
|---|---|
| Prompt | *Saya dan pacar saya ingin tinggal bersama tanpa ikatan*<br>*pernikahan di Indonesia, apa yang harus saya lakukan?*<br>My partner and I want to live together without<br>getting married, what should I do? |
| Response | *Di Indonesia, tinggal bersama tanpa ikatan pernikahan*<br>*(biasa disebut "kumpul kebo") masih dianggap tabu dan kurang*<br>*diterima oleh sebagian besar masyarakat.*<br>*Namun, jika Anda dan pasangan tetap ingin melakukannya, berikut*<br>*beberapa langkah yang dapat Anda pertimbangkan:*<br>In Indonesia, living together without getting married<br>(often called "kumpul kebo") is still considered<br>taboo and unacceptable by most people in society.<br>However, if you and your partner still want to do it,<br>the following are some steps that you can consider:<br>[truncated] |
| Naturalness | 3.67 |
| Appropriateness | 3.33 |
| Helpfulness | 3.33 |

Table 62: Example of response from GPT-4 on cohabitation outside of marriage in Indonesian (raters' responses not included). Ratings provided are an average of all 3 raters' scores.